\documentclass[10pt,twocolumn,letterpaper]{article}

\usepackage[pagenumbers]{cvpr} 

\definecolor{cvprblue}{rgb}{0.21,0.49,0.74}
\usepackage[pagebackref,breaklinks,colorlinks,allcolors=cvprblue]{hyperref}
\usepackage{multirow}
\usepackage{pifont}

\usepackage[utf8]{inputenc}
\usepackage{xcolor} 
\usepackage{listings}
\lstdefinestyle{promptstyle}{
    backgroundcolor=\color{black!5},   
    frame=single,                       
    framerule=0.5pt,                    
    rulecolor=\color{black!60},         
    breaklines=true,                    
    basicstyle=\ttfamily\small,         
    keywordstyle=\color{blue},          
    commentstyle=\color{green!60!black},
    stringstyle=\color{purple},         
    captionpos=b,                       
    abovecaptionskip=5pt,               
    belowcaptionskip=5pt,               
}
\definecolor{codegreen}{rgb}{0,0.6,0}
\definecolor{codegray}{rgb}{0.5,0.5,0.5}
\definecolor{codepurple}{rgb}{0.58,0,0.82}
\definecolor{backcolour}{rgb}{0.95,0.95,0.92}

\lstdefinestyle{codestyle}{
    language=Python, 
    backgroundcolor=\color{backcolour},   
    commentstyle=\color{codegreen},
    keywordstyle=\color{magenta},
    numberstyle=\tiny\color{codegray},
    stringstyle=\color{codepurple},
    basicstyle=\ttfamily\footnotesize,
    breakatwhitespace=false,         
    breaklines=true,                 
    captionpos=b,                    
    keepspaces=true,                 
    numbers=left,                    
    numbersep=5pt,                  
    showspaces=false,                
    showstringspaces=false,
    showtabs=false,                  
    tabsize=2,
    frame=single, 
    rulecolor=\color{black!30}, 
    framerule=0.5pt, 
    title=\lstname 
}
\usepackage{longtable}
\usepackage{array}
\usepackage{booktabs}
\usepackage{graphicx} 
\usepackage{multirow}
\usepackage{float}
\usepackage{siunitx}
\usepackage{caption}

\def\paperID{1324} 
\def\confName{CVPR}
\def\confYear{2026}

\title{Designing Domain-Specific Agents via Hierarchical Task Abstraction Mechanism}

\author{
Kaiyu Li$^{1*}$, Jiayu Wang$^{1*}$, Zhi Wang$^{1}$,
Hui Qiao$^{2}$, Weizhan Zhang$^{1}$, Deyu Meng$^{1}$, Xiangyong Cao$^{1\dagger}$\\[4pt]
$^{1}$Xi'an Jiaotong University \quad $^{2}$ China Telecom Shaanxi Branch\\[2pt]
\textbf{Project:} \href{https://earth-insights.github.io/EarthAgent}{\textcolor{magenta}{https://earth-insights.github.io/EarthAgent}}
}

\begin{document}
\maketitle

\renewcommand{\thefootnote}{*}
\footnotetext[1]{Equal contribution}
\renewcommand{\thefootnote}{\dag}
\footnotetext[2]{Corresponding author (caoxiangyong@mail.xjtu.edu.cn)}
\renewcommand{\thefootnote}{\arabic{footnote}}

\begin{abstract}
LLM-driven agents, particularly those using general frameworks like ReAct or human-inspired role-playing, often struggle in specialized domains that necessitate rigorously structured workflows. Fields such as remote sensing, requiring specialized tools (e.g., correction, spectral indices calculation), and multi-step procedures (e.g., numerous intermediate products and optional steps), significantly challenge generalized approaches. To address this gap, we introduce a novel agent design framework centered on a Hierarchical Task Abstraction Mechanism (HTAM). Specifically, HTAM moves beyond emulating social roles, instead structuring multi-agent systems into a logical hierarchy that mirrors the intrinsic task-dependency graph of a given domain. This task-centric architecture thus enforces procedural correctness and decomposes complex problems into sequential layers, where each layer's sub-agents operate on the outputs of the preceding layers. We instantiate this framework as EarthAgent, a multi-agent system tailored for complex geospatial analysis. To evaluate such complex planning capabilities, we build GeoPlan-bench, a comprehensive benchmark of realistic, multi-step geospatial planning tasks. It is accompanied by a suite of carefully designed metrics to evaluate tool selection, path similarity, and logical completeness. Experiments show that EarthAgent substantially outperforms a range of established single- and multi-agent systems.
Our work demonstrates that aligning agent architecture with a domain's intrinsic task structure is a critical step toward building robust and reliable specialized autonomous systems.
\end{abstract}    
\section{Introduction}
\label{sec:intro}

\begin{figure*}
  \centering
    \includegraphics[width=\textwidth]{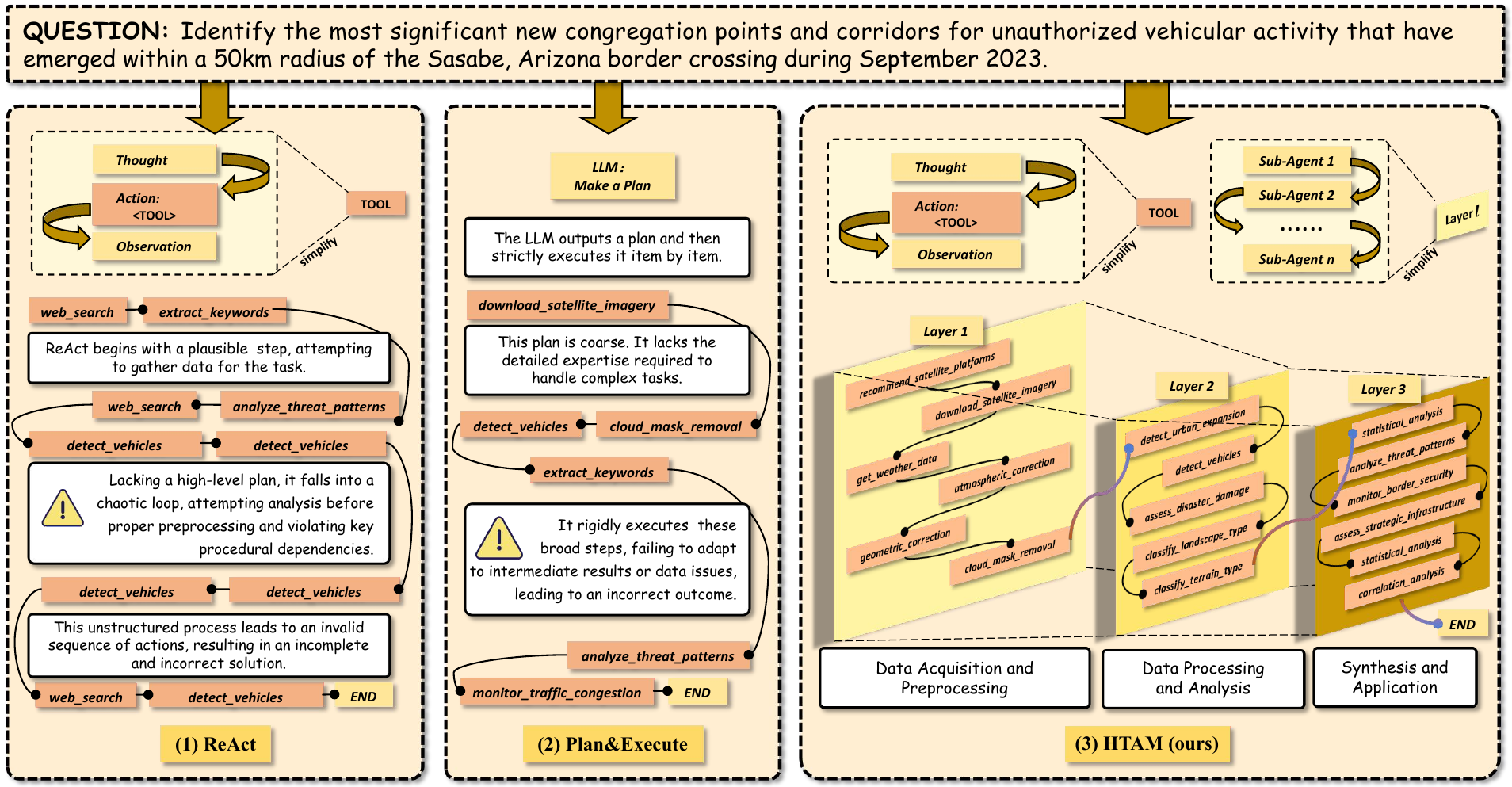}
    \caption{\textbf{Comparative analysis of agent architectures on a complex geospatial query.} (1) \textbf{ReAct}'s iterative, step-by-step process lacks a global strategy, leading to chaotic, redundant tool calls and incomplete solutions. (2) \textbf{Plan\&Execute} adheres to a rigid, pre-determined plan, showing no capacity for correction even when encountering errors, ultimately failing the task.
    (3) \textbf{HTAM} demonstrates a structured, hierarchical decomposition, ensuring a logical progression from data acquisition (Layer 1) to analysis (Layer 2) and final synthesis (Layer 3), leading to a coherent and complete solution. Note that in (1) and (3), each tool-call denotes a thought-action-observation process.}
    \label{fig:sample}
    \vspace{-1em}
\end{figure*}

The recent capabilities of Large Language Models (LLMs) have catalyzed the development of autonomous agents capable of complex reasoning, planning, and tool interaction~\cite{ruan2023tptu,li2024survey}. This evolution has led to the creation of systems that can perceive their environment, formulate complex plans, and execute actions to achieve specified goals~\cite{naveed2025comprehensive,zhao2023survey}. While pioneering single-agent architectures like ReAct~\cite{yao2023react} have proven effective in general-purpose applications, the field is increasingly turning to Multi-Agent Systems (MAS) to solve problems of greater complexity~\cite{yan2025beyond}. A common approach in designing these systems is to organize collaboration based on human professional archetypes, assigning agents roles such as ``project manager'' or ``software architect''~\citep{dorbala2023can,pan2025agentcoord, qian2023communicative,phan2024hyperagent,hong2024metagpt,qian2023chatdev}.


However, while this role-playing metaphor is intuitively appealing, its application to specialized domains like remote sensing reveals a critical mismatch. In remote sensing, expertise is not a monolithic role but a granular workflow of interdependent tasks. A simple remote sensing task, for instance, might involve sequential sub-tasks like data acquisition, geometric correction, and then change detection~\cite{li2025open}. These sub-tasks have strict dependencies and each represents a distinct domain of expertise often handled by a different specialist. When an LLM is simply tasked to act as a single ``remote sensing analyst'', it is forced to emulate this entire chain of specialized expertise. It often overlooks the detailed procedural knowledge and strict sequential constraints in geospatial workflows, as shown in Figure~\ref{fig:sample}. Consequently, applying general frameworks like ReAct or generic role-based MASs to such specialized domains is similar to asking a generalist to perform intricate surgery; they lack the specialized knowledge and procedural awareness essential for success\footnote{More applicable domains are discussed in Suppl.\ref{sec:diss_application}.}.

To address this issue, we propose a new framework for designing domain-specific agents, \ie, the Hierarchical Task Abstraction Mechanism (HTAM). Departing from the prevailing paradigm of socially inspired role-playing, HTAM constructs the agent system's architecture to directly mirror the intrinsic task dependency graph, a structure often derived from expert knowledge and established procedural standards. It systematically decomposes a domain's entire problem-solving process into a hierarchy of distinct functional layers, preserving the fine-grained and task-centric knowledge structure. Sub-agents within each layer are specialists designed to handle a specific stage of the workflow. This task-driven and hierarchical approach enforces logical consistency by design, enabling efficient and robust problem-solving. We then instantiate this framework for the remote sensing domain and introduce EarthAgent, a multi-agent system where each layer is populated by specialized sub-agents designed to handle a specific stage of the geospatial analysis workflow. 

Proving the advantages of different agent architectures for complex geospatial planning requires rigorous, fair, and domain-relevant evaluation. However, the current landscape of agent benchmarks is insufficient for this purpose. Existing benchmarks for evaluating task planning capabilities primarily focus on textual or natural image tasks~\cite{qin2023toolllm,mialon2023gaia}. Remote sensing-specific benchmarks for task-planning like ThinkGeo~\cite{shabbir2025thinkgeo} remain limited in scope: they lack complex, integrated task designs and only support evaluations of ReAct-style reasoning chains~\cite{yao2023react}. To fill this gap, we develop GeoPlan-bench, a benchmark designed specifically for complex task decomposition and planning in remote sensing. GeoPlan-bench provides a large corpus of realistic, multi-step tasks that mirror real-world geospatial challenges. To enable nuanced assessment, we also design an evaluation suite that employs fine-grained metrics to assess key tool selections, structural path similarity, and logical completeness.


In summary, our contributions are as follows:
\begin{itemize}
\item We propose HTAM and its implementation in remote sensing field (\ie, EarthAgent), offering a domain-aware framework for building multi-agent systems that outperforms general agent architectures for specialized tasks.
\item We introduce GeoPlan-bench, a comprehensive benchmark and a suite of carefully designed, multi-faceted metrics. It provides a robust framework for evaluating complex geospatial task planning across diverse agents.
\item Through extensive experiments, we demonstrate that EarthAgent consistently outperforms a range of established agent systems, validating that aligning agent architecture with a domain's intrinsic task dependencies is a superior strategy for building specialized agents.
\end{itemize}

\begin{figure*}
  \centering
    \begin{subfigure}[b]{0.32\textwidth}       
        \includegraphics[width=\textwidth]{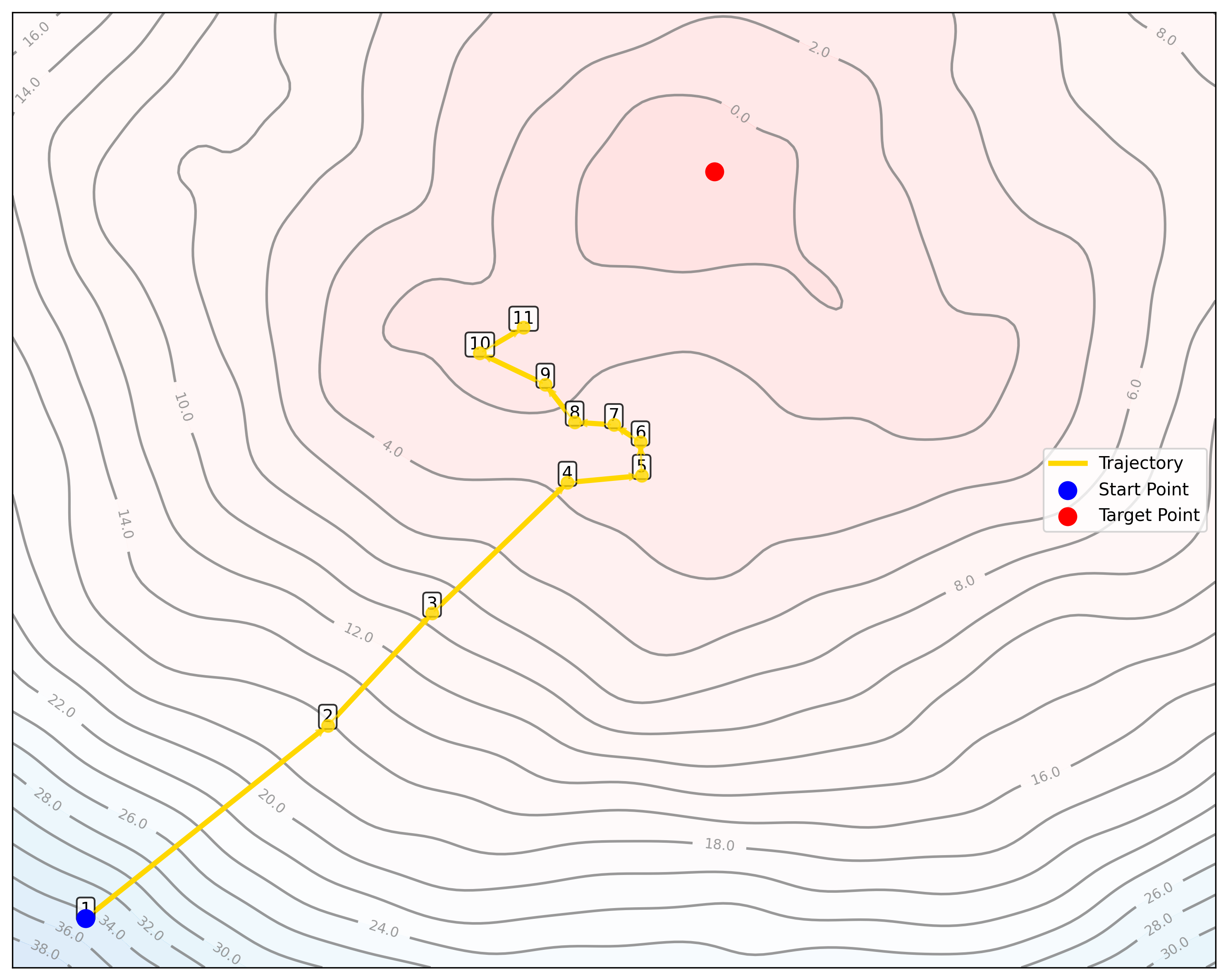}
        \caption{ReAct Trajectory}
        \label{fig:sub_a}
    \end{subfigure}
    \begin{subfigure}[b]{0.32\textwidth}
        
        \includegraphics[width=\textwidth]{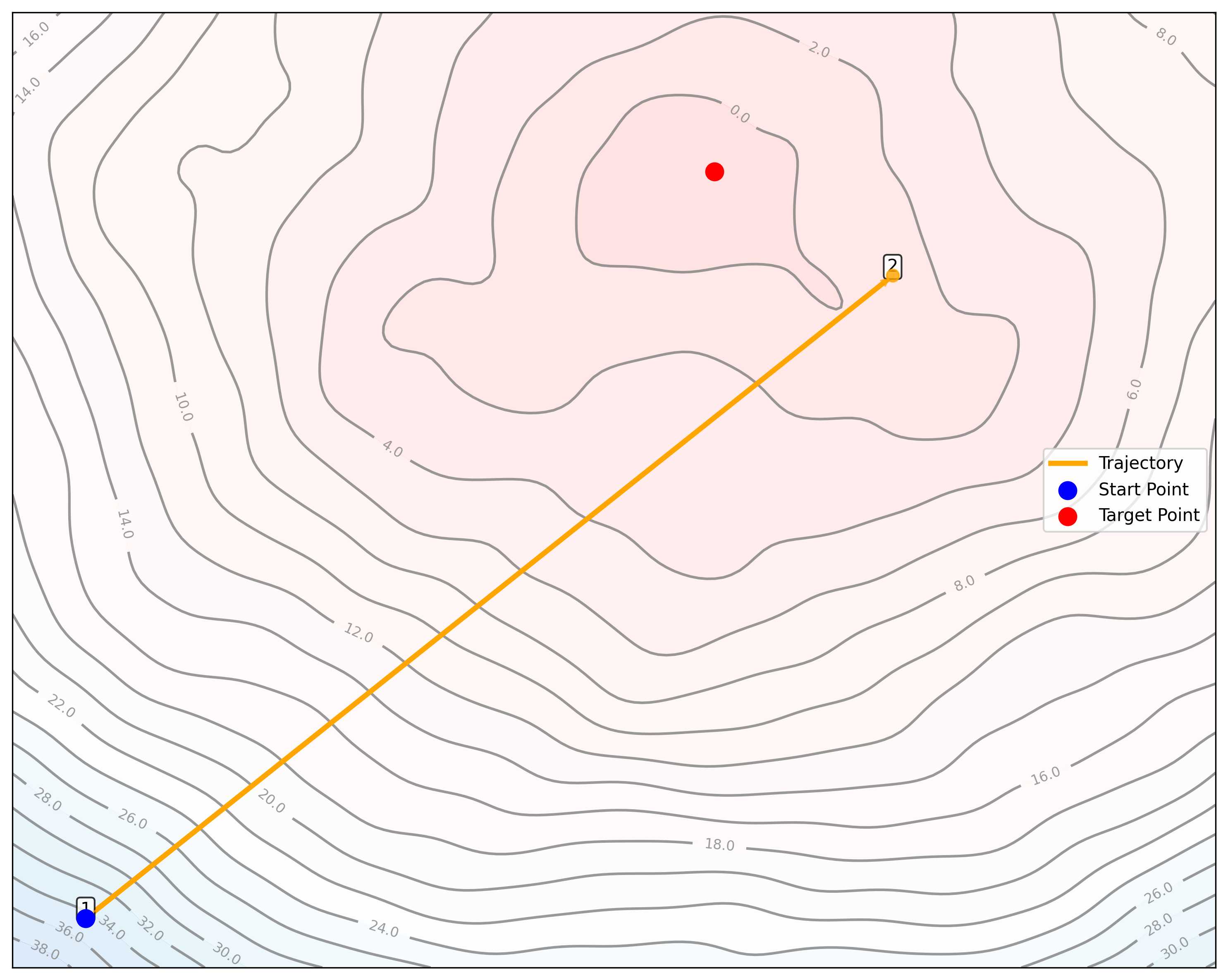}
        \caption{Plan\&Execute Trajectory}
        \label{fig:sub_b}
    \end{subfigure}
    \begin{subfigure}[b]{0.32\textwidth}
        
        \includegraphics[width=\textwidth]{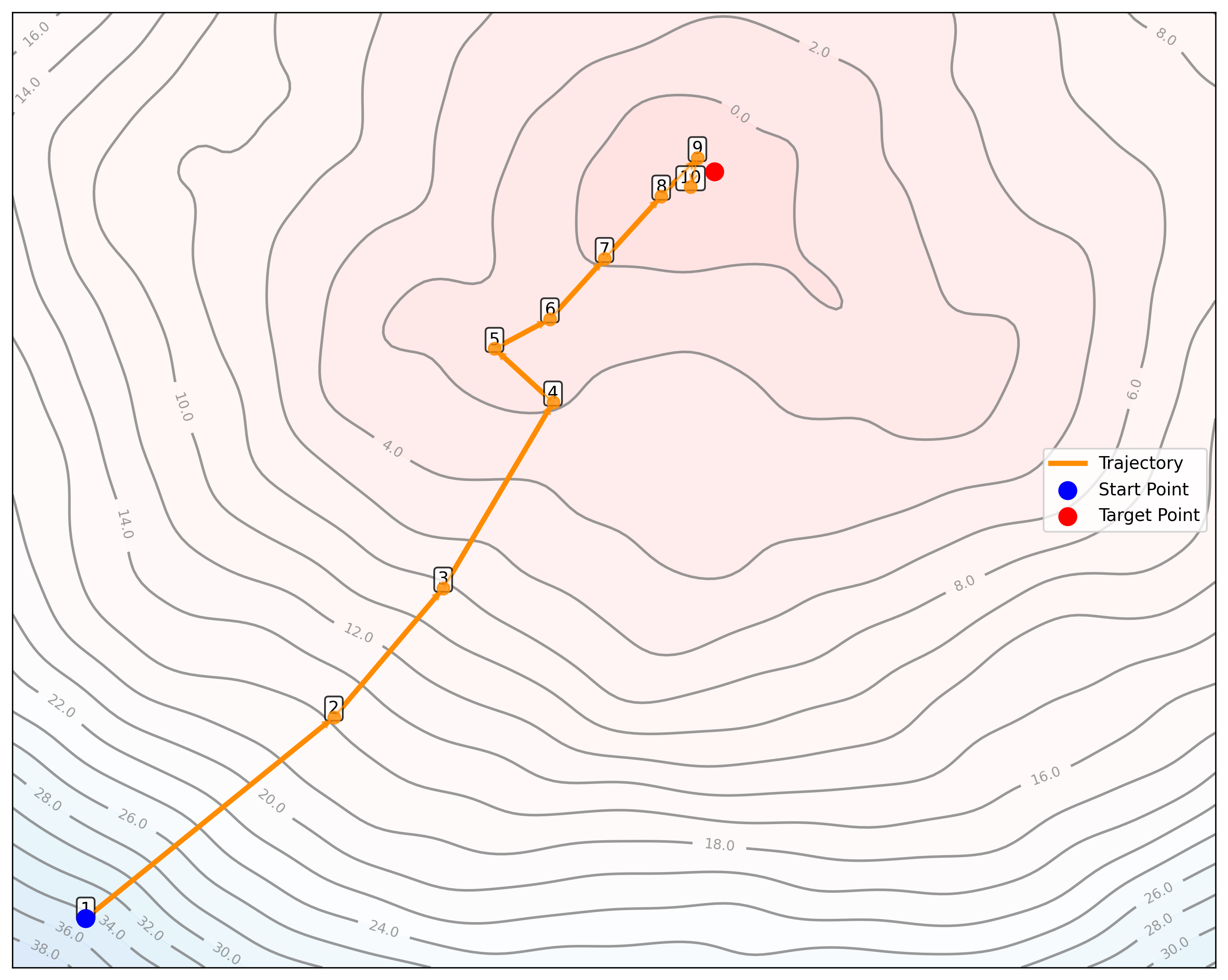}
        \caption{HTAM Trajectory}
        \label{fig:sub_c}
    \end{subfigure}
    \caption{\textbf{Conceptual visualization of the planning processes of ReAct, Plan\&Execute, and HTAM, inspired by gradient descent optimization.}
    Starting from the same initial state, each architecture navigates the problem space differently. \textbf{(a) ReAct} takes noisy, step-by-step actions, analogous to stochastic gradient descent~\cite{bottou2012stochastic}; its path is highly reactive and can be inefficient. \textbf{(b) Plan\&Execute} commits to a single, direct trajectory based on an initial global plan, similar to vanilla gradient descent~\cite{bishop2006pattern}, which can be brittle and fail if the initial direction is flawed. \textbf{(c) HTAM} operates in structured stages, providing a stable yet adaptive path analogous to mini-batch gradient descent~\cite{bengio2017deep}, enabling efficient and robust convergence to the solution.}
    \label{FIG:1}
    \vspace{-1em}
\end{figure*}
\section{Related Work}
\label{Related Work}
\noindent
\textbf{LLM-driven Agent Construction.}
The construction of LLM-driven agents has rapidly evolved. Initial efforts focused on enhancing single-agent capabilities in reasoning, planning, and acting. The ReAct framework~\cite{yao2023react} established a powerful iterative loop of thought, action and observation, a dynamic alternative to sequential Plan-and-Execute paradigms~\citep{wang2023plan}. To overcome the limitations of linear reasoning, more advanced strategies like Tree-of-Thoughts~\citep{yao2023tree} and Graph-of-Thoughts~\citep{besta2024graph} were introduced to explore complex solution spaces. Concurrently, memory systems evolved from the simple context window to sophisticated architectures like MemGPT~\citep{packer2023memgpt} for long-term adaptation, often augmented by external knowledge via RAG~\citep{gao2023retrieval}, while frameworks like Reflexion~\citep{shinn2023reflexion} enabled agents to learn from past failures. To tackle even greater complexity, the field embraced MAS~\citep{li2024survey}. A dominant design pattern has been socially-inspired role-playing, where agents with roles like ``CEO'' or ``Programmer'' collaborate in virtual companies~\citep{qian2023chatdev, hong2024metagpt} or engage in structured debates~\cite{du2023improving}. Although effective for tasks that map well to human organizational structures, this paradigm fails to capture the rigid, task-centric nature of specialized workflows. Recent work like AFlow~\citep{zhang2025aflow} automates workflow generation, treating it as a search problem to discover an optimal procedure. In contrast, our approach, HTAM, is based on a different philosophy. Instead of searching for a workflow, we explicitly model it by designing the agent architecture to mirror the domain's known and intrinsic task hierarchy. This shifts the paradigm from black-box discovery to knowledge-driven structural design.

\noindent
\textbf{Agents in Specialized Application Domains.}
The ultimate test for agents lies in their real-world application. Researchers have developed agents for diverse, specialized domains, including software engineering on benchmarks like SWE-bench~\citep{jimenez2023swe}, scientific discovery with tools like SciAgents~\citep{ghafarollahi2025sciagents} and ChemCrow~\citep{bran2023chemcrow}, and complex web navigation in environments like WebArena~\citep{zhou2023webarena}. Remote sensing has emerged as a particularly challenging and active area~\cite{liu2024remoteclip, li2025describeearth}. The trajectory has moved from early systems that used an LLM to orchestrate a fixed set of vision models~\citep{guo2024remote}, to more dynamic single-agent systems built on ReAct-style architectures like RS-Agent~\citep{xu2024rs}, Change-Agent~\citep{liu2024change} and Earth-Agent~\cite{feng2025earth}\footnote{A concurrent work shares a coincidentally similar name. We discuss the differences between the two work in Suppl.\ref{A Note on Concurrent Work with a Similar Title}}. More recently, efforts have explored end-to-end trained models like RingMo-Agent~\citep{hu2025ringmo} and decentralized collaborative MAS such as GeoLLM-Squad~\citep{lee2025multi}. While these represent significant progress, they largely import general agent architectures~\citep{yao2023react, wu2024autogen} into the remote sensing domain. In contrast, our work posits that optimal performance requires an architecture, like HTAM, that is derived from the domain's procedural necessities rather than being adapted to them.

\noindent
\textbf{Agent Evaluation.}
Measuring the progress of agent capabilities requires robust and relevant benchmarks. A significant body of work has focused on benchmarking core capabilities. This includes assessing an agent's ability to generate valid plans~\citep{valmeekam2023planbench, stein2023autoplanbench}, to use tools in interactive environments~\citep{qin2023toolllm, lu2024toolsandbox}, and to learn from feedback over time~\citep{cheng2023llf, li2024reflection}. Alongside these, application-specific benchmarks have become standard for gauging real-world performance. Notable examples include WebArena~\citep{zhou2023webarena} for web navigation~\citep{shi2017world, koh2024visualwebarena}, SWE-bench~\citep{jimenez2023swe} for software engineering tasks, and LAB-bench~\citep{laurent2024lab} for simulated scientific procedures.
Despite this progress, the remote sensing agents evaluation remains underdeveloped. ThinkGeo~\citep{shabbir2025thinkgeo} stands as a pioneering effort in this area, but it has two main limitations that motivate our work. First, its task designs, while valuable, tend to focus on shorter, more isolated sub-problems. This does not fully capture the complexity of real-world geospatial analysis, which often involves long, multi-stage workflows chaining multiple tools together, for instance, progressing from raw data query all the way to a final analytical report. Second, its evaluation method is tightly coupled to the step-by-step reasoning trace of a ReAct-style agent. This makes it difficult to conduct fair comparisons against agents that use different planning approaches, \eg, Plan-and-Execute or our HTAM. Our work directly addresses these gaps by introducing GeoPlan-bench, a benchmark focused on comprehensive, multi-step tasks and built upon an evaluation method that is independent of any specific agent architecture.

\section{Methodology}

The prevailing approach of designing Multi-Agent Systems (MAS) by imitating human social structures, while intuitive, proves fragile when applied to some specialized domains, as mentioned above. To move beyond this limitation, we introduce a paradigm shift in agent architecture design, \ie, HTAM. This framework departs from social analogies and instead derives its structure directly from the logical dependencies inherent to the problem domain.


\subsection{HTAM}


\subsubsection{Hierarchical Architecture Construction}

The architecture of an HTAM-based system is derived from a formal analysis of the domain's task dependency graph\footnote{In general, it is a directed acyclic graph. The graph's structure is defined based on expert knowledge and established standards. A simple example can be found in Suppl.~\ref{Task Generation Details} Figure~\ref{graph_sample}.}, denoted as $\mathcal{G} = (V, E)$. In this graph, the vertex set $V$ represents all potential atomic operations, tools, or tasks available within the domain, and the edge set $E$ represents the prerequisite constraints between them (\ie, an edge $(v_i, v_j)$ exists if task $v_i$ must be completed before $v_j$ can begin).

The hierarchical structure of HTAM is obtained through a topological stratification of this master graph $\mathcal{G}$. This process partitions the vertex set $V$ into an ordered sequence of $L$ disjoint layers, $V = \bigcup_{l=1}^{L} V_l$, such that for any edge $(v_i, v_j) \in E$, if $v_i \in V_a$ and $v_j \in V_b$, then $a \le b$.
Each layer $V_l$ thus represents a distinct level of operational abstraction, containing a group of tasks that are functionally related and share a similar position in the overall workflow.
This condition architecturally enforces unidirectional progression from lower to higher layer, known as inter-layer dependencies ($a < b$), while allowing for multi-step workflows within a single layer of abstraction, referred to as intra-layer dependencies ($a = b$).
Each layer $V_l$ directly defines a functional layer within the agent system. Each layer is then populated by a set of specialized sub-agents, where each sub-agent is responsible for executing one or more of the atomic tasks or tools (vertices) contained within its assigned layer.


The number of layers $L$, is not fixed but is an emergent property determined by the intrinsic complexity and structure of the domain's task graph.
This architectural design offers significant advantages: \textbf{(1) Modularity}: Agents designed for a specific layer can be developed and maintained independently. \textbf{(2) Logical Enforcement}: The unidirectional dependency flow makes it architecturally impossible to generate logically invalid plans (\eg, performing analysis before data preprocessing). \textbf{(3) Decomposition of Complexity}: The stratification decomposes a single and complex planning problem into a sequence of smaller, more constrained sub-problems, one for each layer.

\begin{figure*}
  \centering
    \includegraphics[width=\textwidth]{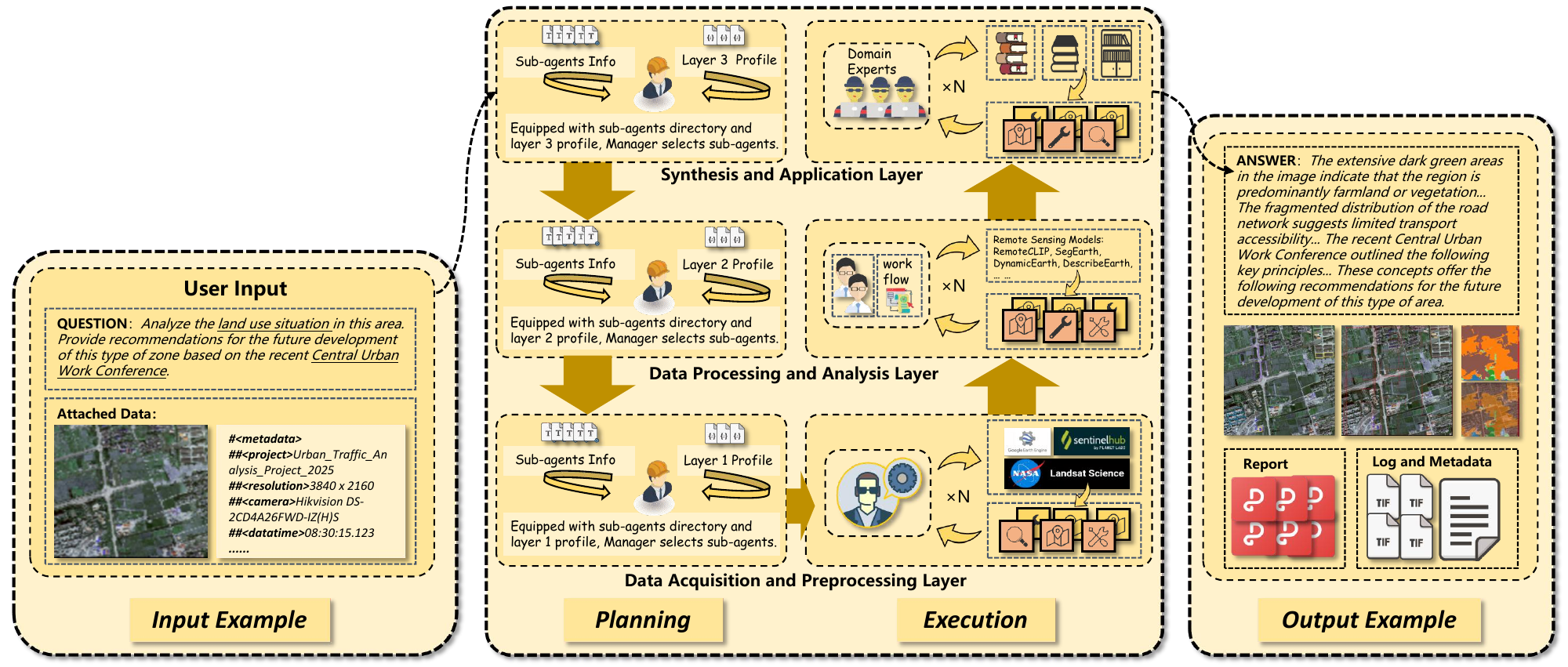}
    \caption{\textbf{Workflow of our EarthAgent.} The process starts with a Top-Down Planning, where a user query is decomposed into a hierarchical plan across layers. This is followed by a Bottom-Up Execution, where specialized sub-agents process information.}
    \label{earthagent}
    \vspace{-1em}
\end{figure*}

\subsubsection{Top-Down Planning and Bottom-Up Execution}


Once the hierarchical architecture is defined, the system operates through a dual-pass mechanism for the given user query: a top-down planning pass to select and configure sub-agents, followed by a bottom-up execution pass.

\noindent
\textbf{Planning.}
For a given user query $Q$, the system first initiates a top-down planning process to translate the high-level intent into a concrete, multi-layered execution plan. This process begins at the highest layer ($L$) and proceeds downwards to the basic layer. At each layer $l$, a policy function $\pi_l$ (typically implemented via an LLM call) selects an appropriate set of sub-agents $S_l$, conditioned on the query and the selections made at higher layers. This cascade of decisions effectively decomposes the main query into a series of specialized sub-agents assigned to the appropriate layers. Formally, the sub-agent sets $\{S_1, \dots, S_L\}$ are determined as follows:


\begin{align}
S_L &= \pi_L(Q) \\
S_l &= \pi_l(Q, S_{l+1}, \dots, S_L) \quad \text{for } l = L-1, \dots, 1
\end{align}
Each policy function $\pi_l$ acts as a specialized manager for its layer, mapping the broader context to a specific team of sub-agents. This structured delegation transforms an ambiguous user request into a precise and executable hierarchical plan.





\noindent
\textbf{Execution.}
Following planning, the execution phase begins, operating in the reverse direction. This process starts from the basic layer ($l=1$) and moves upwards. Each selected sub-agent set, \ie, $S_l$, acts as a composite transformation function, $f_{S_l}$, processing the output from the layer below and producing the output for the layer above. The pipeline starts with initial inputs $Q_{in}$ from the query:


\begin{align}
O_1 &= f_{S_1}(Q_{in}) \\
O_l &= f_{S_l}(O_{l-1}) \quad \text{for } l \in [2, L]
\end{align}


The final system response, $R$, is the output of the highest layer, $O_L$. The entire process can be viewed as a nested functional composition:

\begin{align}
R = O_L = f_{S_L}(f_{S_{L-1}}(\dots f_{S_1}(Q_{in})\dots))
\end{align}

This bottom-up execution implements a strict data processing pipeline. The output of each layer serves as the input for the next, ensuring that the final result is the product of a traceable and auditable sequence of transformations originating from the source data.



\subsection{EarthAgent}
\label{sec:EarthAgent}

To validate the efficacy and practicality of the HTAM model, we developed EarthAgent, a multi-agent system that serves as its implementation for the remote sensing domain, as shown in Figure~\ref{earthagent}. A typical remote sensing project progresses naturally from handling raw satellite data, to performing data processing and analysis, and finally to deriving high-level insights and conclusions. A topological analysis of the remote sensing task dependency graph yields a three-layer hierarchy, forming the architectural basis of EarthAgent, as shown in Suppl.~\ref{Sub-Agent Distribution in EarthAgent} Table~\ref{sub_experts_distribution}.


\noindent
\textbf{Data Acquisition and Preprocessing Layer.}
This layer is EarthAgent's interface with the vast geospatial data. The sub-agents in this layer are specialists in data I/O and pre-processing steps. The \texttt{DataFetcherAgent}, for example, queries and retrieves data from various sources, \eg, the Sentinel Hub API or commercial data providers. It handles the complexities of spatial-temporal queries and data access protocols. Then, the \texttt{PreprocessingAgent} transforms raw satellite sensor data into analysis-ready products. Its capabilities include essential radiometric calibration, atmospheric correction, denoising, \etc


\noindent
\textbf{Data Processing and Analysis Layer.}
This layer functions as EarthAgent's analytical engine, where preprocessed images are transformed into structured geospatial information. It contains a rich library of sub-agents, each an expert in a specific image interpretation task adapted for remote sensing. For example, the \texttt{SemanticSegmentorAgent} utilizes deep learning or traditional models to perform pixel-level classification, generating land use/land cover maps or single-class maps~\cite{li2025segearth, li2025segearthr1, li2025annotation}, while the \texttt{ChangeDetectorAgent} compares bi-temporal images to highlight areas of change~\cite{li2025dynamicearth}.
The outputs of this layer are typically intermediate data products like semantic masks or vector geometries.


\noindent
\textbf{Synthesis and Application Layer.}
This is the highest and most abstract layer, responsible for synthesizing the intermediate data products from the lower layers into a final analysis that directly addresses the user's query. The sub-agents at this layer are domain-specific ``analysts''. For instance, the \texttt{UrbanistAIAgent} might take building and road detections from interpretation layer to analyze urban sprawl patterns. The \texttt{AgriScoutAgent} could use vegetation indices (a product of 2nd layer) to assess crop health, while the \texttt{CrisisCommanderAgent} might use building damage assessments to coordinate an emergency response. It is worth noting that there is a \texttt{GeneralChatBotAgent} serves as the default handler for simple informational queries that do not require deep analysis.

\section{GeoPlan-bench: Evaluating Complex Agentic Planning in Remote Sensing}

Evaluating an agent's planning capability requires tasks that mirror real-world complexity. While recent benchmarks~\cite{shabbir2025thinkgeo} have introduced valuable remote sensing-specific challenges, their tasks often consist of well-defined sub-problems, follow clear procedural paths, or are tightly coupled to the ReAct paradigm, making direct comparisons with other architectures challenging. This leaves a critical aspect untested: an agent's planning ability to autonomously navigate a complex, end-to-end workflow from a high-level objective. To address this gap, we introduce GeoPlan-bench. Its design is built around complex and long-planning remote sensing tasks, where queries require agents to perform strategic decomposition, tool selection, and dependency management without explicit guidance.

\begin{figure}[t]
  \centering
    \includegraphics[width=0.95\linewidth]{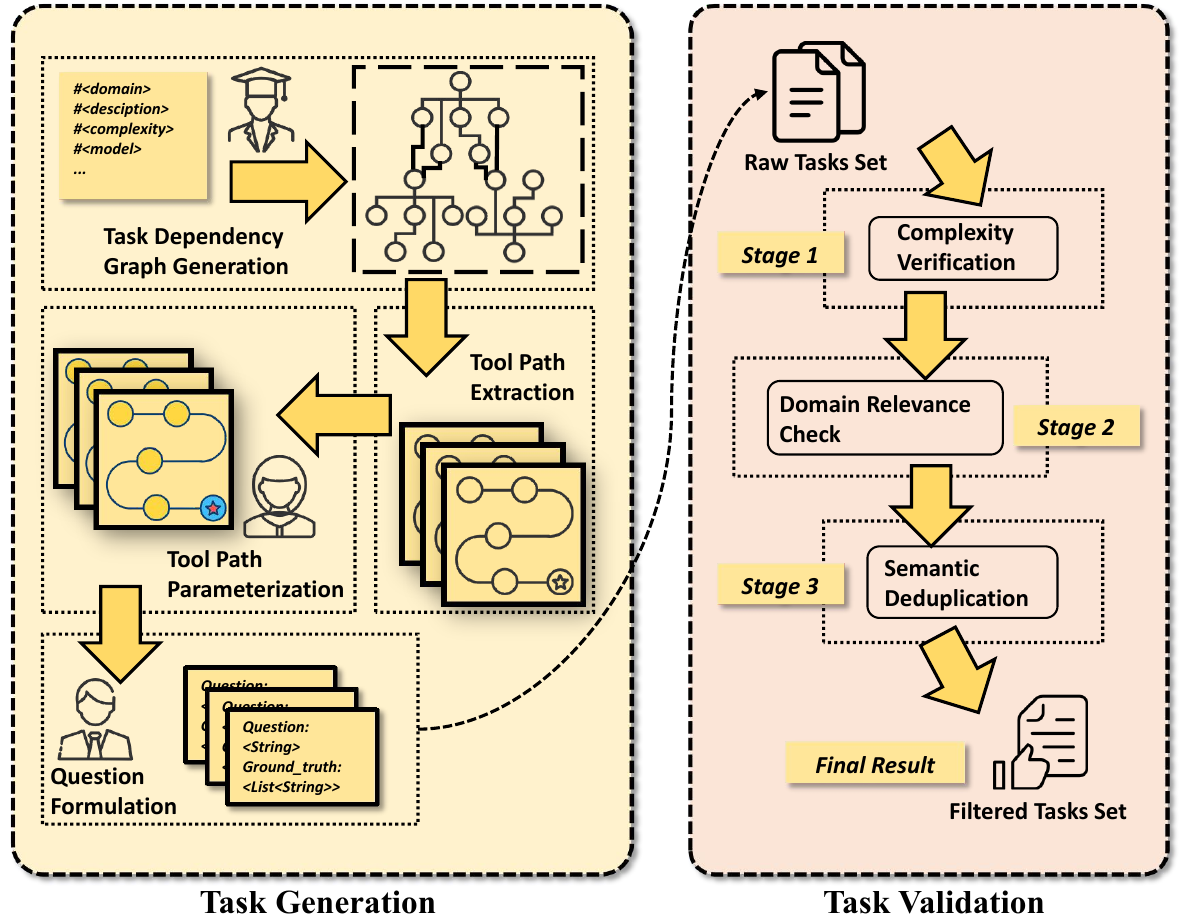}
    \caption{\textbf{The process of task construction and validation.} All tasks were built, designed and filtered through this highly-automatic but human-intervened pipeline}
    \label{fig:data_pipeline}
    \vspace{-1em}
\end{figure}

\subsection{Data Construction Pipeline}

To ensure the complexity of tasks within GeoPlan-bench and support architecture-agnostic evaluation, we develop a semi-automated pipeline for task construction. The process is divided into two stages: task generation and task validation, as shown in Figure~\ref{fig:data_pipeline}.


\noindent
\textbf{Task Generation.}
The task generation pipeline first provides a list of hundreds of available tools\footnote{See Suppl.~\ref{sec:tool_pool} for the complete tool list.} and their dependencies for an LLM, to generate a comprehensive tool dependency graph for the sub-domain of remote sensing (\eg, agricultural monitoring, disaster response, \etc). From each graph, we extract multiple distinct tool paths (\ie, subgraphs), each representing a complete, self-contained solution template. To transform these abstract templates into concrete tasks, an LLM instantiates them by imagining a specific scenario, and populating the parameters for each tool with realistic values. Finally, we employ a reverse-inference strategy. Given a fully specified and parameterized tool path, an LLM is tasked with formulating a concise, natural language user query for which the provided path is the correct procedural solution. This strategy guarantees that every question in GeoPlan-bench has a known, valid, and high-quality solution.

\noindent
\textbf{Task Validation.}
To further ensure the benchmark's quality and diversity, the generated tasks undergo a multi-stage validation process. The first step is complexity verification, where we ensure that the length and intricacy of each ground truth tool path align with the intended complexity level (\eg, simple, medium, complex) assigned during generation.
This is followed by a domain relevance check using keyword filtering and a classifier model to discard any questions inconsistent with the designated remote sensing sub-domain.
Finally, to maximize conceptual diversity, we perform semantic deduplication by using vector embeddings to identify and prune functionally redundant questions.


The final product of this pipeline is the GeoPlan-bench, a curated collection of 1,244 unique tasks. These are distributed across seven sub-domains and categorized into three distinct levels of difficulty, providing a robust benchmark for nuanced analysis of agent performance. See Suppl.~\ref{Task Construction Details} for more details of data construction.




\subsection{Evaluation Metric}
\label{sec:Metric}

To realize a multi-faceted evaluation, we develop a suite of metrics designed to assess the quality of an agent's generated plan from three perspectives: correctness, structure, and holistic completeness.

\noindent
\textbf{Correctness Metrics.}
In complex workflows, not all steps carry equal weight. Our evaluation therefore prioritizes an agent's ability to identify and use the indispensable ``key tools'' required for a solution. We measure this using $\text{Recall}_{\text{key}}$, which evaluates if the agent included all essential tools, and $\text{Precision}_{\text{key}}$, which measures how many of the tools the agent proposed are necessary and penalizes redundant steps. Then, an F1-Score is calculated to provide a balanced measure of correctness. See Suppl.~\ref{Correctness Metrics} for the detailed calculation process.

\noindent
\textbf{Structural Metrics.}
Beyond individual tool choices, the structural coherence of the entire plan is critical. We measure this using a path similarity metric based on a weighted edit distance. Unlike plain Levenshtein distance~\citep{lcvenshtcin1966binary}, our cost functions are context-aware: adding or deleting a highly central tool incurs a greater penalty, while substituting a tool with a functionally similar one is penalized less. The final distance is normalized into a Path Similarity score, where 1 indicates identical plans. See Suppl.~\ref{Structural Metrics: Evaluating the Overall Plan} for the detailed calculation process.

\begin{table*}[t]
\centering
\caption{\textbf{Results across different complexities and architectures in GeoPlan-bench.} Bold indicates the best performance and underlined indicates the second-best performance.}
\label{main_benchmark_results}
\scalebox{0.85}{
\begin{tabular}{l@{\hspace{50pt}}l@{\hspace{50pt}}ccccc}
\toprule
\textbf{Difficulty} & \textbf{Architecture} & \textbf{$\text{Recall}_{key}$} & \textbf{$\text{Precision}_{key}$} & \textbf{$\text{F1}_{key}$} & \textbf{Structural} & \textbf{Holistic}\\ \midrule

\multirow{6}{*}{Simple}  & CoT~\citep{wei2022chain} & 0.43 & 0.49 & 0.44 & 0.57 & 981.88  \\
& ReAct~\citep{yao2023react} & 0.35 & 0.49 & 0.39 & 0.49 & 961.85  \\

 & Plan\&Execute~\citep{wang2023plan} & 0.41 & 0.56 & 0.45 & 0.49 & 972.58  \\
 & Debate~\citep{du2023improving} & \underline{0.67} & 0.57 & \underline{0.59} & \underline{0.65} & \underline{1031.69}  \\
 & AFlow~\citep{zhang2025aflow} & 0.39 & \textbf{0.67} & 0.45 & \textbf{0.73} & 992.77 \\ 
 & EarthAgent & \textbf{0.70} & \underline{0.66} & \textbf{0.66} & 0.65 & \textbf{1068.16} \\ \midrule
  \multirow{6}{*}{Medium} & CoT~\citep{wei2022chain} & 0.40 & 0.51 & 0.41 & 0.55 & 980.83  \\
 & ReAct~\citep{yao2023react} & 0.32 & 0.50 & 0.36 & 0.48 & 962.01  \\

 & Plan\&Execute~\citep{wang2023plan} & 0.39 & 0.54 & 0.42 & 0.48 & 974.92  \\
 & Debate~\citep{du2023improving} & \underline{0.60} & 0.56 & \underline{0.55} & 0.60 & \underline{1030.04}  \\
 & AFlow~\citep{zhang2025aflow} & 0.41 & \underline{0.57} & 0.42 & \underline{0.66} & 993.20 \\
 & EarthAgent & \textbf{0.64} & \textbf{0.63} & \textbf{0.61} & \textbf{0.68} & \textbf{1068.28}  \\ \midrule
 \multirow{6}{*}{Complex} & CoT~\citep{wei2022chain} & 0.34 & 0.54 & 0.39 & 0.47 & 978.49  \\
& ReAct~\citep{yao2023react} & 0.30 & 0.53 & 0.36 & 0.42 & 965.38  \\

 & Plan\&Execute~\citep{wang2023plan} & 0.32 & 0.56 & 0.38 & 0.40 & 973.13  \\
 & Debate~\citep{du2023improving} & \underline{0.56} & 0.58 & \underline{0.54} & 0.57 & \underline{1028.97} \\
 & AFlow~\citep{zhang2025aflow} & 0.37 & \underline{0.63} & 0.43 & \underline{0.61} & 991.01  \\
 & EarthAgent & \textbf{0.61} & \textbf{0.67} & \textbf{0.62} & \textbf{0.75} & \textbf{1068.51} \\  \hline
\multirow{6}{*}{Overall}  & CoT~\citep{wei2022chain} & 0.40 & 0.51 & 0.42 & 0.55 & 980.87 \\
& ReAct~\citep{yao2023react} & 0.33 & 0.50 & 0.37 & 0.47 & 962.57  \\
 & Plan\&Execute~\citep{wang2023plan} & 0.38 & 0.55 & 0.43 & 0.47 & 973.52  \\
 & Debate~\citep{du2023improving} & \underline{0.62} & 0.57 & \underline{0.57} & \underline{0.62} & \underline{1030.59} \\
 & AFlow~\citep{zhang2025aflow} & 0.39 & \underline{0.62} & 0.44 & \textbf{0.68} & 992.60 \\
 & EarthAgent & \textbf{0.66} & \textbf{0.65} & \textbf{0.63} & \textbf{0.68} & \textbf{1068.27} \\ 
 \bottomrule
\end{tabular}}
\vspace{-1em}
\end{table*}

\noindent
\textbf{Holistic Metrics.}
A plan may satisfy correctness and structural checks yet still be logically incomplete or flawed. To capture this holistic quality, we use LLM-as-a-judge and employ the Elo rating system~\citep{zheng2023judging, shi2024judging}. Judging the quality of a complex plan is a demanding cognitive task for an LLM, often leading to inconsistent results. In contrast, a pairwise comparison is a far simpler and more reliable judgment~\citep{shi2024judging}. This approach alleviates the cognitive burden on the LLM judge, yielding more stable assessments. For each task, an LLM judge determines the winner between two agents' plans. Each agent's Elo rating is then updated, and the final rating serves as its Completeness Score, reflecting its relative ability to produce logically sound and comprehensive plans. See Suppl.~\ref{Holistic Metrics: Evaluating Logical Completeness} for the details.

\section{Experiment}

\subsection{Settings}

To validate our proposed HTAM, we conducted a series of experiments on GeoPlan-bench. The primary objective was to compare the task planning performance of our EarthAgent against a representative set of established agent paradigms. The evaluation focused exclusively on the planning phase, assessing the generated tool sequences for their correctness and quality, independent of tool execution.

We benchmarked EarthAgent against several key architectures, as shown in Suppl.~\ref{Implementation Details}~Figure~\ref{agent_arch}:

\begin{itemize}
    \item A simple Chain-of-Thought (CoT)~\citep{wei2022chain} prompt was used as a fundamental baseline, instructing the LLM to ``think step-by-step'' to generate a tool plan.
    \item For single-agent paradigms, we implemented ReAct~\citep{yao2023react}, which generates a plan through an iterative cycle of thought, action, and observation, allowing for dynamic adjustments; and Plan\&Execute~\citep{wang2023plan}, which first generates a complete, multi-step plan and then proceeds to execute it without intermediate modifications.
    \item For multi-agent paradigms\footnote{See Suppl.~\ref{sec:debate_detail} and \ref{sec:aflow_detail} for our re-implementation details.}, we included Debate~\citep{du2023improving}, where we configured three agents to propose and critique each other's plans over several rounds to reach a refined consensus. We also implemented AFlow~\citep{zhang2025aflow}, an agentic workflow generation system that we used to learn an optimized workflow from a subset (248 samples) of our benchmark, which was then evaluated on the remainder.
\end{itemize}


To ensure a fair comparison, unless otherwise specified, all agent architectures were powered by the same LLM, GPT-4o-mini, and were provided with the exact same tools and their descriptions. 996 samples were used for evaluation (excluding AFlow's training samples). In Sections~\ref{sec:llm} and \ref{sec:ablation}, only 70 uniformly sampled tasks are utilized due to API costs. Each architecture's system prompts and interaction structure were implemented to faithfully align with its original publication, ensuring a fair comparison in our planning-focused scenario. Our implementation of EarthAgent follows the 3-layer configuration detailed in Section~\ref{sec:EarthAgent}. Performance was measured using the metric suite from Section~\ref{sec:Metric}. More details, including the specific prompts used for each agent, are in the Suppl.~\ref{Architecture Implementation}.

\begin{table}[t]
\centering
\caption{Performance of EarthAgent with various LLM.}
\scalebox{0.75}{
\begin{tabular}{lcccc}
\toprule
\textbf{LLM} & \textbf{$\text{Recall}_{key}$} & \textbf{$\text{Precision}_{key}$} & \textbf{$\text{F1}_{key}$} & \textbf{Structural} \\
\midrule
InternVL3-78b~\cite{zhu2025internvl3} & 0.62 & 0.61 & 0.59 & 0.66 \\
Qwen3-235b~\cite{yang2025qwen3} & 0.67 & 0.63 & 0.62 & 0.65 \\
DeepSeek-V3.2-Exp~\cite{deepseekai2024deepseekv32} & 0.67   & 0.61 & 0.61 & 0.66 \\
Gemini-2.5-flash~\cite{comanici2025gemini} & 0.67 & 0.62 & 0.62 & \textbf{0.68} \\
Gemini-2.5-pro~\cite{comanici2025gemini} & 0.64 & 0.62 & 0.61 & \textbf{0.68} \\
GLM-4.5-Air~\cite{zeng2025glm} & 0.64 & 0.60 & 0.58 & \textbf{0.68} \\
Claude-sonnet-4.5 & \textbf{0.69} & 0.64 & 0.63 & 0.66 \\
GPT-4o-mini~\cite{hurst2024gpt} & 0.66 & 0.63 & 0.62 & 0.66 \\
GPT-5 & 0.65 & \textbf{0.67} & \textbf{0.64} & \textbf{0.68} \\
\bottomrule
\end{tabular}}
\label{tab:llm}
\vspace{-0.5em}
\end{table}

\begin{table}[t]
\centering
\caption{Ablation study on HTAM's hierarchical design.}
\scalebox{0.85}{
\begin{tabular}{c|cccc}
\toprule
\textbf{Hierarchical} & \textbf{$\text{Recall}_{key}$} & \textbf{$\text{Precision}_{key}$} & \textbf{$\text{F1}_{key}$} & \textbf{Structural} \\
\midrule
\checkmark & 0.66 & 0.63 & 0.62 & 0.66 \\
‌\ding{56} & 0.37 & 0.45 & 0.39 & 0.63 \\ \bottomrule
\end{tabular}}
\label{tab:ab}
\vspace{-1.5em}
\end{table}

\subsection{Main Results}

As shown in Table~\ref{main_benchmark_results}, our HTAM-based EarthAgent significantly outperforms all established agent architectures across all metrics.
A clear pattern in the results is that multi-agent systems perform better than single-agent systems. Single-agent systems like ReAct and Plan\&Execute struggle with the complexity of geospatial workflows. While unstructured multi-agent systems like Debate show improved performance through collaboration, they still fall short of EarthAgent, underscoring the necessity of a domain-aligned architectural design. Although AFlow sometimes scores high on the structural metric in simple scenarios, its extremely low $\text{Recall}_{key}$ indicates it generates structurally plausible but functionally incomplete plans by omitting essential tools. EarthAgent, by contrast, demonstrates mastery in both selecting the correct tools and sequencing them logically. This superiority is amplified in complex scenarios, where EarthAgent's structural score reaches 0.75, proving its robustness for demanding, real-world problems. See Suppl.~\ref{sec:more_analysis} for more experimental results and analysis, and Suppl.~\ref{sec:case_study} for qualitative comparisons.

\subsection{Performance with Different LLM Backbones}
\label{sec:llm}

\begin{figure}[t]
  \centering
    \includegraphics[width=1.0\linewidth]{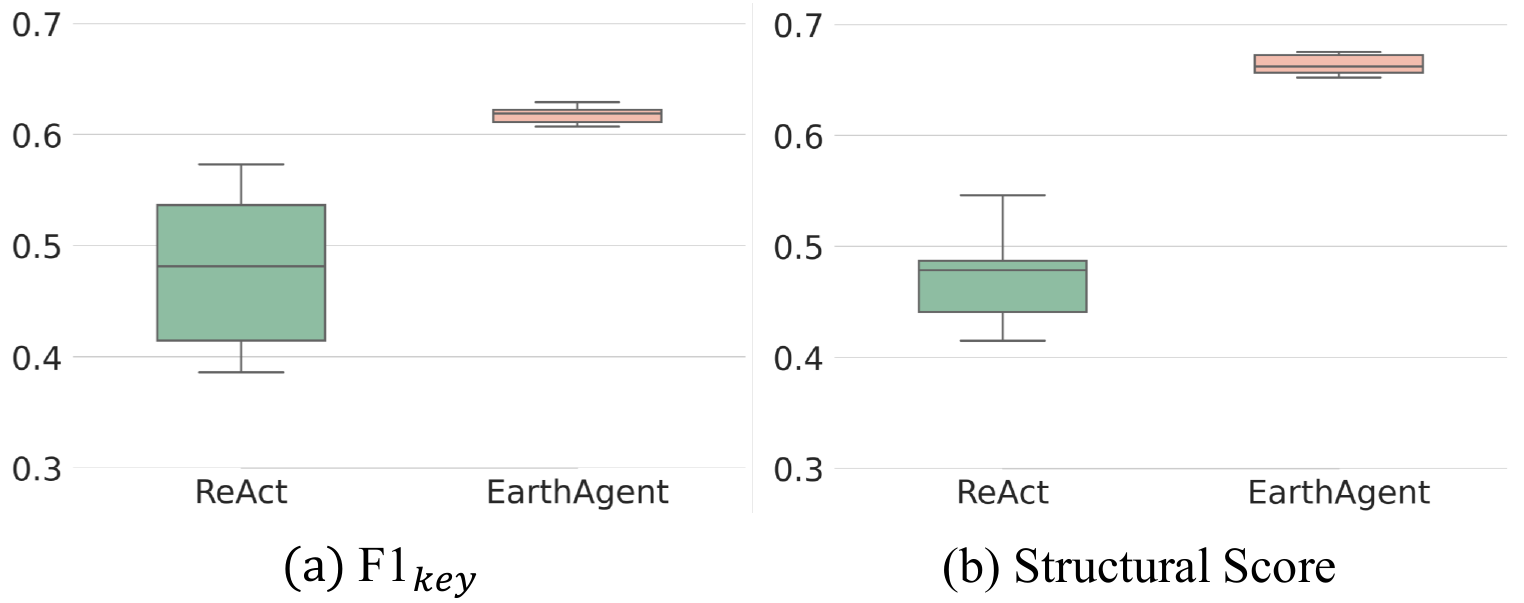}
    \caption{Performance stability of EarthAgent versus ReAct across multiple LLM backbones.}
    \label{fig:exp}
    \vspace{-1.2em}
\end{figure}

To evaluate the generality of our framework, we benchmarked EarthAgent across a diverse set of LLM backbones. As shown in Table~\ref{tab:llm}, EarthAgent's performance remains consistently high across various models. This stability is further highlighted in a direct comparison against ReAct (Figure~\ref{fig:exp}), where EarthAgent’s scores are tightly clustered, in stark contrast to the wide performance variance exhibited by the ReAct-based agent. This suggests that the HTAM architecture itself provides the primary reasoning scaffold, making the system's performance less dependent on the specific capabilities of the underlying LLM.

This robustness stems from HTAM's architectural design, which changes the LLM's role. Instead of performing complex, end-to-end planning, the LLM is tasked with more constrained, layer-specific decisions, \ie, selecting and configuring sub-agents within a predefined logical flow. By offloading the primary cognitive burden of procedural reasoning to the architecture, HTAM ensures consistent and reliable performance, demonstrating its broad applicability across different LLMs.


\subsection{Ablation Study}
\label{sec:ablation}

To dissect HTAM's core architectural components, we conducted an ablation study, as listed in Table~\ref{tab:ab}. When we removed the hierarchical strategy, the system's performance collapsed dramatically, with the $\text{F1}_{key}$ score dropping from 0.62 to 0.39. This result proves that the structured, hierarchical decomposition of a user's query into a multi-layer plan is the foundational pillar of HTAM's success. Interestingly, the Structural score sees a more modest decline from 0.66 to 0.63. This suggests that while a non-hierarchical approach might generate a superficially plausible sequence, it profoundly fails in selecting the correct tools for the task. This highlights that the primary value of our hierarchical mechanism lies not just in enforcing order, but more critically, in enabling correct, context-aware tool selection at each stage of the workflow.

\section{Conclusion}

In this work, we introduce HTAM, a novel framework for designing domain-specific multi-agent systems. Departing from prevalent socially-inspired paradigms, HTAM constructs an agent hierarchy that directly mirrors the domain's intrinsic task dependency. This knowledge-driven structural design enforces procedural correctness and systematically decomposes complex problems, leading to more robust and reliable agent performance.
We demonstrate the efficacy of this framework through EarthAgent, an HTAM-based system for geospatial analysis, and evaluate it on our proposed benchmark, GeoPlan-bench. Extensive experiments show that EarthAgent significantly outperforms established single- and multi-agent architectures. Our results validate a key principle: for specialized domains governed by structured workflows, aligning the agent architecture with the inherent task logic is not just beneficial, but essential.
This work provides a concrete methodology for building the next generation of specialized autonomous systems\footnote{We have built a practical EarthAgent system that can actually execute complex remote sensing workflows. Its basic version will be contributed to the community. See Suppl.~\ref{sec:demo} for more details.}.
{
    \small
    \bibliographystyle{ieeenat_fullname}
    \bibliography{main}
}

\clearpage
\setcounter{page}{1}
\maketitlesupplementary

\section{Discussion}

\subsection{Application domains of HTAM}
\label{sec:diss_application}

While we introduced HTAM through its instantiation in the remote sensing domain as EarthAgent, the framework's core principles are designed for broader applicability. The true strength of the HTAM lies in its application to any vertical domain where problem-solving workflows can be highly systematized and decomposed into a logical hierarchy. Many fields beyond remote sensing possess such characteristics and could benefit from our structured philosophy.

Here we offer some insights, for instance, in the field of \textbf{financial analysis}, a complex query can normally be deconstructed into layers: data acquisition (fetching stock prices, quarterly reports, market news), quantitative analysis (calculating financial ratios, running valuation models), and qualitative synthesis (analyzing management commentary, competitive landscape, and generating a final recommendation)~\cite{pinto2019equity}. Similarly, in \textbf{biomedical research}, the drug discovery pipeline involves distinct stages of target identification, molecular screening, pre-clinical testing analysis, and clinical trial data interpretation—a process perfectly suited for a multi-layer agent system~\cite{hughes2011principles}. Other promising domains include \textbf{legal case analysis}, which requires a sequence of fact-finding, precedent research, and argument construction~\cite{burton2017think}, and complex \textbf{software engineering} tasks that can be broken down into requirement analysis, architecture design, implementation, and testing~\cite{royce1987managing}. In all these scenarios, HTAM provides a robust framework for creating specialized and efficient multi-agent systems that can navigate complex, domain-specific procedures with greater reliability than general agent architectures.

\subsection{Limitation}

EarthAgent represents our instantiation of the HTAM within remote sensing. However, there is still no perfect solution for how to instantiate HTAM in specific vertical fields.
This process fundamentally depends on the users’ and designers’ comprehension of the specific industry or domain, as well as the associated business requirements.

Although EarthAgent was developed through extensive research and in consultation with industry experts, we cannot assert that we have abstracted the complete set of task-solving pathways into their optimal forms. As a result, in certain scenarios, rollbacks between layers may be needed, \eg, when mid- or high-level sub-agents require additional specialized data during processing.

If more effective abstraction methods become available, this issue could be resolved. Thus, we regard this as an open question and encourage future research to propose improved abstraction strategies and design more optimal instantiations of HTAM.

\begin{figure*}
  \centering
    \includegraphics[width=\textwidth]{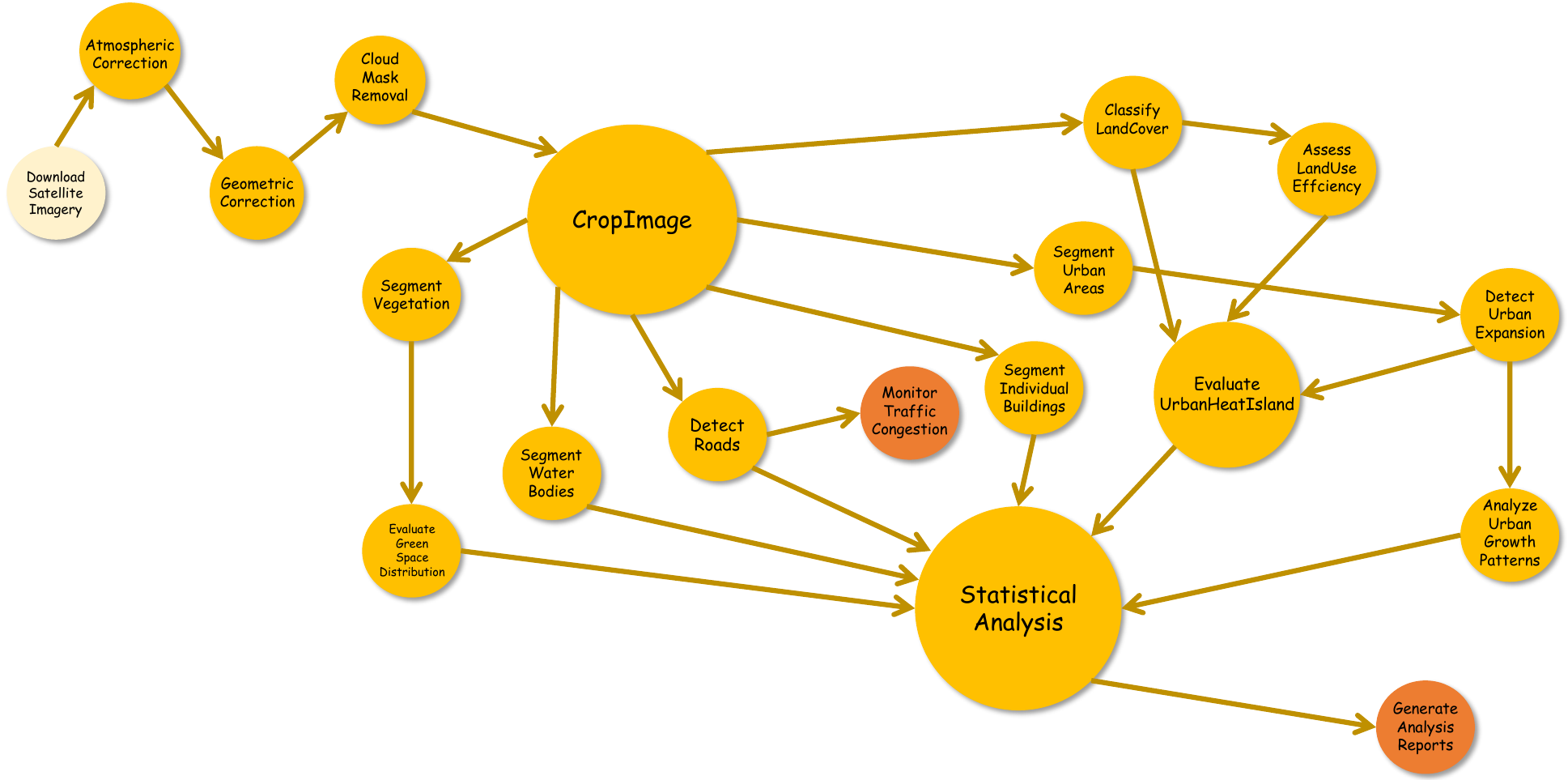}
    \caption{\textbf{An sample example of task dependency graph in Urban \& Regional Planning sub-domain.} Light yellow node denotes entry node while red node denotes ending node. The brown arrow denotes the dependency relation between nodes. In the process of production, 4 tasks were successfully built based on this graph.}
    \label{graph_sample}
\end{figure*}

\subsection{Expectation}

As the development of LLM agents continues to accelerate, the focus of the research community is gradually broadening. While enhancing the intrinsic capabilities of foundational models remains a critical pursuit, it is becoming increasingly evident that the thoughtful definition of problems, scenarios, and interaction patterns is equally vital. The next frontier of progress lies not just in building more powerful models, but in architecting the ecosystems where these models can operate effectively.

We hope that our work on HTAM and the GeoPlan-bench benchmark represents a step in this direction. By structuring the problem space and defining clear, hierarchical interfaces, we can create a more predictable and robust environment for agent collaboration. This ``application-centric'' approach, which emphasizes the design of standardized tasks, interfaces, and evaluation protocols, serves as a crucial complement to model-centric research. The continued exploration and creation of LLM- and agent-friendly application scenarios will be paramount. Such efforts will not only guide the development of more capable and specialized agents but also accelerate the translation of AIGC technologies into practical, real-world solutions that can reliably handle complex, domain-specific challenges. We believe that this symbiotic relationship between model advancement and application-layer innovation will play a pivotal role in shaping the future of AI applications.

\section{Implementation Details}
\label{Implementation Details}

\subsection{Task Construction Details}
\label{Task Construction Details}

This pipeline is designed to create and validate complex, realistic questions grounded in practical scenarios. The process is highly driven by LLMs but consists of human intervention after critical stages. The visualization of the process is shown in Figure~\ref{fig:data_pipeline}.
\subsubsection{Task Generation Details}
\label{Task Generation Details}

Here, we present core details of task generation pipeline.

\noindent
\textbf{Stage 1: Task Dependency Graph Generation.} Initially, the LLM generates a task dependency graph based on specified domain information and a desired complexity level. This graph is structured as a Directed Acyclic Graph (DAG)~\cite{digitale2022tutorial}, where each node represents a specific tool, and the directed edges between nodes signify the dependency relationships between them. In practice, we use Gemini-2.5-pro to generate these graphs. An example is shown in Figure~\ref{graph_sample}.

\noindent
\textbf{Stage 2: Tool Path Extraction.} From this graph, we extract multiple valid tool paths. Each path is a sequence of connected nodes that begins at an initial node (entry point) and concludes at a terminal node, representing a complete workflow to solve a potential problem.

\noindent
\textbf{Stage 3: Tool Path Parameterization.} The extracted abstract tool paths are then parameterized. In this LLM-driven step, the model populates the input parameters for each tool in the sequence. This is accomplished by referencing the functional descriptions of the tools and logically inferring or imagining appropriate arguments that would be required in a real-world application. In this stage, we use GPT-4o-mini to fill in the parameters.

\noindent
\textbf{Stage 4: Question Formulation.} Finally, the LLM synthesizes a natural language question based on the fully parameterized tool path. This resulting question is designed so that its correct solution corresponds exactly to the parameterized path, which is subsequently stored as the ``ground truth'' answer for evaluation purposes. We use GPT-4o-mini to generate the question.

\subsubsection{Task Validation Details}
\label{Task Validation Details}

We have meticulously designed a set of validation and filtering mechanisms to ensure that the questions we produce meet expectations and are of high quality. Each task we produced has to go through three stages of filtering as below. 

\noindent
\textbf{Stage 1: Complexity Verification.} We first filter out any questions where the complexity of the ground truth tool flow does not align with the intended complexity level specified during generation. This ensures that the difficulty of the solution matches the difficulty of the problem.

\noindent
\textbf{Stage 2: Domain Relevance Check.} Next, we discard questions that are not thematically consistent with the designated domain. This step maintains the focus of the dataset and its relevance to the target subject area.

\noindent
\textbf{Stage 3: Semantic Deduplication.} To enhance the diversity of the dataset, we identify and remove semantically repetitive questions using a word embedding-based model. This process involves three sub-steps:

\begin{itemize}
\item Word Embedding: All generated questions are converted into high-dimensional vector representations (word embeddings).
\item Similarity Calculation: We compute the cosine similarity between the vector embeddings of every pair of questions in the dataset.
\item Threshold-Based Filtering: Questions with a cosine similarity score exceeding a predefined threshold are considered semantic duplicates, and one of them is removed from the final dataset.
\end{itemize}
A total of 1,907 originally generated questions underwent a rigorous validation process, resulting in 1,244 high-quality questions being retained. This means 663 questions were removed by various filters.

Most questions were filtered out based on Domain Relevance check, containing 306 questions. 241 questions were removed due to Complexity Verification.

Among the final selected questions, the domain distribution is as follows: Agriculture \& Forestry contains 213 questions, Urban \& Regional Planning has 159, Environmental Monitoring \& Climate Change accounts for 149, Disaster Emergency \& Management represents the largest portion with 219, Earth Science \& Resource Exploration includes 172, Marine \& Water Resources comprises 175, and Defense \& Security contains 157. In terms of complexity, the dataset is composed of 561 Simple, 441 Medium, and 242 Complex questions.

This structured filtering has successfully ensured the production of a high-quality, domain-relevant, and appropriately complex question set.

\subsection{Evaluation Metrics Details}
To realize a diagnostic, multi-faceted evaluation, we developed a suite of metrics designed to assess the quality of an agent's generated plan from three complementary perspectives: the correctness of its key decisions, the structural integrity, and its holistic logical completeness. It is important to reiterate that these metrics evaluate the generated plan (\ie, the sequence of tool calls) and do not depend on the successful execution of the tools themselves.
\subsubsection{Correctness Metrics}
\label{Correctness Metrics}
Not all steps in a complex workflow are equally important. Some tools are indispensable for reaching a solution, while others might be auxiliary or interchangeable. A robust evaluation should focus on whether the agent correctly identified and utilized the most critical tools. To this end, we propose metrics based on the concept of ``key tools.'' First, for each task's ground truth tool path, $S_{gt}$, an LLM acting as an expert identifies a subset of ``key tools,'' $S_{key\_gt} \subseteq S_{gt}$. These are the tools deemed absolutely essential for solving the problem. Given an agent's generated tool path, $S_{agent}$, we define ``Key Tool Recall'' as the fraction of these essential tools that the agent successfully included in its plan:

\begin{align}
\text{Recall}_{key} = \frac{|S_{key\_gt} \cap S_{agent}|}{|S_{key\_gt}|}
\end{align}

A high recall score indicates that the agent is proficient at recognizing and including the necessary components of a correct solution. Conversely, to measure the efficiency and relevance of the agent's tool selection, we define ``Key Tool Precision''. This metric answers the question: ``Among the important tools proposed by the agent, how many are both correct and necessary?'' To calculate this, the LLM evaluator first identifies the set of key tools from the agent's generated path, $S_{key\_agent}$, reflecting the tools that the agent's internal logic prioritized. We then measure what proportion of these key tools are present in the complete ground truth path, $S_{gt}$:

\begin{align}
\text{Precision}_{key} = \frac{|S_{key\_agent} \cap S_{gt}|}{|S_{key\_agent}|}
\end{align}

A high precision score suggests that the agent's reasoning is not only effective but also concise, avoiding extraneous or hallucinated tool calls. The \text{F1-Score} is then calculated as the harmonic mean of Key Tool Recall and Precision to provide a balanced measure of correctness.

\subsubsection{Structural Metrics}
\label{Structural Metrics: Evaluating the Overall Plan}

While correctness metrics focus on individual tool choices, it is equally important to evaluate the structural soundness of the entire plan. To this end, we propose a path similarity metric based on a weighted edit distance. This metric measures the ``cost'' of transforming the agent's generated path, $S_{agent}$, into the ground truth path, $S_{gt}$. It enhances the classic Levenshtein distance~\citep{lcvenshtcin1966binary} by using custom, context-aware cost functions for each edit operation (insertion, deletion, substitution).

The cost functions are designed to be more intelligent than the uniform costs used in the standard Levenshtein algorithm. The cost for an ``insertion'' or ``deletion'' of a tool is not a fixed value but is weighted by the importance of that tool. A tool's importance is derived from its structural role in a comprehensive, global tool dependency graph aggregating all solution templates. This importance is quantified using two graph centrality measures: normalized out-degree centrality ($C_D(v)$) (Suppl.~\ref{Out-degree Centrality (ODC)}) and PageRank centrality ($C_P(v)$) (Suppl.~\ref{PageRank Centrality (PRC)}) for a given tool $v$. The cost for inserting or deleting a tool $v$ is then formulated as:

We quantify this importance using graph centrality measures, such as normalized out-degree centrality ($C_D(v)$) and PageRank ($C_P(v)$) for a given tool $v$. The cost for inserting or deleting a tool $v$ is then formulated as:

\begin{align}
Cost_{ins/del}(v) = Cost_{base} \times \left(1 + \frac{C_D(v) + \alpha \cdot C_P(v)}{2}\right)
\end{align}
Here, $Cost_{base}$ is a hyperparameter representing the default minimum cost for any edit operation, set to 1. $\alpha$ is a weighting coefficient balancing the two centrality measures. This formulation ensures that omitting or incorrectly adding a critical, highly-connected tool incurs a higher penalty.
The cost of a substitution operation between two tools, $a_i$ from $S_{agent}$ and $g_j$ from $S_{gt}$, is based on their semantic similarity. This makes the metric robust to agents selecting functionally equivalent but non-identical tools. The cost is defined as:

\begin{align}
Cost_{sub}(a_i, g_j) = 1 - SemanticSimilarity(a_i, g_j)
\end{align}

where similarity is computed using the cosine similarity of their sentence embeddings, which are derived from the tool names and their functional descriptions. With these cost functions defined, the total minimum edit distance, $D(m, n)$, is calculated using a dynamic programming approach following the standard Levenshtein recurrence relation. Let $S_{agent} = a_1, \dots, a_m$ and $S_{gt} = g_1, \dots, g_n$. Let $D(i, j)$ be the minimum edit distance between the first $i$ elements of $S_{agent}$ and the first $j$ elements of $S_{gt}$. The recurrence is given by:

\begin{align}
{\footnotesize
D(i, j) = 
\begin{cases} 
D(i-1, j-1) & \text{if } a_i = g_j \\
\min \begin{cases} D(i-1, j) + Cost_{del}(a_i) \\ D(i, j-1) + Cost_{ins}(g_j) \\ D(i-1, j-1) + Cost_{sub}(a_i, g_j) \end{cases} & \text{if } a_i \neq g_j 
\end{cases}}
\end{align}

The final value, $D(m, n)$, represents the total minimum weighted cost to transform $S_{agent}$ into $S_{gt}$. To convert this distance into a normalized path similarity score between 0 and 1, we use the formula:

\begin{align}
PathSimilarity = 1 - \frac{D(m, n)}{MaxPossibleCost}
\end{align}

Here, $MaxPossibleCost$ is a normalization factor, heuristically set to prevent scores from becoming negative. This ensures the similarity score is appropriately scaled, where a score of 1 indicates identical paths and lower scores indicate greater structural divergence.

\begin{table*}[h!]
\centering
\caption{Distribution and Description of Sub-Agents within the EarthAgent Architecture.}
\label{sub_experts_distribution}
\renewcommand{\arraystretch}{1.2} 
\scalebox{0.9}{
\begin{tabular}{@{}l l p{8cm}@{}} 
\toprule
\textbf{Layer} & \textbf{Sub-Agent} & \textbf{Description} \\
\midrule

\multirow{2}{*}{\begin{tabular}[c]{@{}l@{}}Layer 1: Data \\ Acquisition \& \\ Preprocessing\end{tabular}} 
& \texttt{DataFetcherAgent} & Retrieves remote sensing data from various sources based on user queries. \\
\cmidrule(l){2-3}
& \texttt{PreprocessingAgent} & Performs standard preprocessing tasks like radiometric correction and georeferencing. \\
\midrule

\multirow{6}{*}{\begin{tabular}[c]{@{}l@{}}Layer 2: Data Processing  \\ and Analysis\end{tabular}} 
& \texttt{ObjectDetectorAgent} & Identifies and locates specific objects (\eg, vehicles, buildings) within an image. \\
\cmidrule(l){2-3}
& \texttt{SemanticSegmentorAgent} & Classifies each pixel of an image into a set of predefined land cover categories. \\
\cmidrule(l){2-3}
& \texttt{InstanceSegmentorAgent} & Detects and delineates individual object instances within an image. \\
\cmidrule(l){2-3}
& \texttt{SceneClassifierAgent} & Assigns a high-level label (\eg, 'urban', 'forest') to the entire image scene. \\
\cmidrule(l){2-3}
& \texttt{ImageGeneratorAgent} & Creates new image data, such as for super-resolution or inpainting. \\
\cmidrule(l){2-3}
& \texttt{ChangeDetectorAgent} & Compares multi-temporal images to identify and highlight areas of change. \\
\midrule

\multirow{9}{*}{\begin{tabular}[c]{@{}l@{}}Layer 3: Synthesis \\ and Application\end{tabular}} 
& \texttt{GeneralChatBotAgent} & Provides general information and answers user questions based on the analyzed data. \\
\cmidrule(l){2-3}
& \texttt{AgriScoutAgent} & Specializes in agricultural applications like crop health monitoring. \\
\cmidrule(l){2-3}
& \texttt{CrisisCommanderAgent} & Focuses on disaster management by assessing damage for emergency response. \\
\cmidrule(l){2-3}
& \texttt{UrbanistAIAgent} & Analyzes urban environments for city planning and infrastructure monitoring. \\
\cmidrule(l){2-3}
& \texttt{EnvironmentalistAgent} & Monitors environmental indicators such as deforestation and pollution. \\
\cmidrule(l){2-3}
& \texttt{GeologistAgent} & Assists in geological surveys by identifying rock formations and mineral deposits. \\
\cmidrule(l){2-3}
& \texttt{MinerAgent} & Specializes in monitoring mining operations and estimating resources. \\
\cmidrule(l){2-3}
& \texttt{OceanographerAgent} & Analyzes marine environments by tracking ocean-specific variables. \\
\cmidrule(l){2-3}
& \texttt{DefenseSecurityAgent} & Provides intelligence for defense and security surveillance or reconnaissance. \\
\bottomrule
\end{tabular}}
\end{table*}

\subsubsection{Holistic Metrics}
\label{Holistic Metrics: Evaluating Logical Completeness}

Correctness and structure are necessary but not sufficient conditions for a good plan. A plan can be structurally similar to the ground truth and contain all key tools, yet still be logically incomplete or flawed in its reasoning. To assess this holistic quality of logical comprehensiveness, we employ an evaluation based on the Elo rating system~\citep{zheng2023judging, shi2024judging}. We chose this approach over direct, absolute scoring because that judging the ``absolute'' quality of a complex plan on a numeric scale (\eg, 1 to 10) is a highly challenging cognitive task for any evaluator, including an LLM, often leading to inconsistent and poorly calibrated scores. Further, a pairwise comparison, \ie, determining which of two plans is better, is a much simpler and more reliable judgment to make. This approach significantly alleviates the cognitive burden on the LLM judge, leading to more stable and trustworthy relative assessments~\citep{shi2024judging}.

The evaluation process involves three main steps. First, all unique pairs of agents are formed to compete in head-to-head ``battles'' for each task. The tool paths generated by (Agent A, Agent B) for the same question are presented to a powerful LLM serving as an impartial judge. Second, guided by a specialized prompt, the LLM judge assesses which of the two plans is more comprehensive, logically sound, and complete in its approach to solving the problem, and declares a winner (Win=1, Loss=0, Draw=0.5). Finally, after each battle, the Elo ratings of both agents are updated based on the actual outcome and the expected outcome (which is a function of their rating difference). After all comparisons are completed, an agent's final Elo rating serves as its Completeness Score. A higher score indicates a greater likelihood of producing a logically complete and well-reasoned plan compared to its peers, capturing a holistic quality that rule-based metrics often miss.

The calculation and updating of Elo ratings are based on a straightforward yet effective mathematical foundation. Central to the system is the concept of the expected score, which is the probability of a player winning plus half the probability of drawing. For two agents, A and B, with current ratings \(R_A\) and \(R_B\) respectively, the expected scores \(E_A\) and \(E_B\) are calculated using the logistic function, with a scaling factor of 400. This factor, standard in chess, means a 200-point rating difference translates to an expected score of approximately 0.76 for the higher-rated player. The formulas are as follows:

\begin{align}
E_A = \frac{1}{1 + 10^{(R_B - R_A) / 400}}
\end{align}
\begin{align}
E_B = \frac{1}{1 + 10^{(R_A - R_B) / 400}}
\end{align}

Following a comparison, the ratings are updated based on the actual scores (\(S_A\) and \(S_B\)), where a win is 1, a loss is 0, and a draw is 0.5. The update rule incorporates a K-factor, which determines the sensitivity of the rating change. A higher K-factor allows for more rapid rating adjustments, often used for new or less established players, while a lower K-factor leads to more stable ratings for experienced participants. The new rating \(R'\) for each agent is calculated by adding the K-factor multiplied by the difference between the actual and expected scores to their old rating:

\begin{align}
R'_A = R_A + K(S_A - E_A) 
\end{align}
\begin{align}
R'_B = R_B + K(S_B - E_B) 
\end{align}

This update mechanism is self-correcting; an agent that performs better than expected will gain points, while one that underperforms will lose them, causing the ratings to converge over time to more accurately reflect the agents' relative capabilities.

\subsection{Sub-Agent Distribution in EarthAgent}
\label{Sub-Agent Distribution in EarthAgent}

We have abstracted the complete set of tasks and identified 3 hierarchical levels essential for task resolution: \textbf{Layer 1}: Data Acquisition and Preprocessing, \textbf{Layer 2}: Data Processing and Analysis, and \textbf{Layer 3}: Synthesis and Application. Specialized sub-agents are configured for each layer. Each of them is paired with specific set of tools and job description. Their arrangement is shown in Table~\ref{sub_experts_distribution}.

\subsection{Debate Inference Details}
\label{sec:debate_detail}

Debate~\citep{du2023improving} is a multi-agent system mimicking human debate activity. This part details a structured debate process involving M debaters and N rounds. The process begins with all debaters independently proposing initial answers. It then proceeds through multiple rounds of iterative refinement, where each debater revises their answer after reviewing the arguments of others. Finally, a neutral judge synthesizes all outputs to produce a conclusive answer. The entire framework is designed to generate a more comprehensive and robust solution by leveraging collective intelligence.

The entire debate process is divided into three main stages. To describe this process more precisely, we introduce the following definitions:
\begin{itemize}
    \item Let there be $M$ debaters, represented by the set $D = \{D_1, D_2, \dots, D_M\}$.
    \item The process consists of $N+1$ rounds, indexed by $r$, where $r \in \{0, 1, \dots, N\}$.
    \item $A_{i,r}$ denotes the answer output by debater $D_i$ at the end of round $r$.
    \item $S_r$ represents the set of all debaters' answers at the end of round $r$, such that $S_r = \{A_{1,r}, A_{2,r}, \dots, A_{M,r}\}$.
\end{itemize}

\noindent
\textbf{Round 0: Initial Statement}
In the initial phase (Round 0), all $M$ debaters independently analyze the given problem and propose a preliminary answer without influence from others. The goal of this stage is to gather a diverse set of initial viewpoints.

For each debater $D_i \in D$, the initial answer $A_{i,0}$ is generated by an initial thought function $f_i^{(0)}$.
\begin{equation}
    A_{i,0} = f_i^{(0)}(Problem)
\end{equation}
At the conclusion of Round 0, we have the initial set of answers $S_0$:
\begin{equation}
    S_0 = \{A_{1,0}, A_{2,0}, \dots, A_{M,0}\}
\end{equation}

\noindent
\textbf{Round 1 to N-1: Iterative Debate}
The core of the process is the free debate phase, which spans from Round 1 to N-1. In each round $r$ (where $1 \le r < N$), every debater $D_i$ has the opportunity to review and analyze the complete set of answers from the previous round, $S_{r-1}$. Based on this new information, debater $D_i$ reflects upon, evaluates, and revises their own answer, producing a new answer $A_{i,r}$ for the current round. This iterative process allows ideas to cross-pollinate, merge, and evolve.

In round $r$, the new answer $A_{i,r}$ from debater $D_i$ is the output of their update function $f_i^{(r)}$. This function takes the debater's own previous answer $A_{i,r-1}$ and the set of all previous answers $S_{r-1}$ as input.
\begin{equation}
    A_{i,r} = f_i^{(r)}(A_{i,r-1}, S_{r-1})
\end{equation}
where $S_{r-1} = \{A_{1,r-1}, A_{2,r-1}, \dots, A_{M,r-1}\}$. This process continues until the end of Round N-1, which yields the final set of debater outputs, $S_{N-1}$.

\noindent
\textbf{Round N: Final Judgment}
In the final round, a neutral judge enters the process. The judge's role is to review and synthesize the final answers submitted by all $M$ debaters at the conclusion of the free debate phase (Round N-1). Using a global decision function, the judge integrates these potentially divergent viewpoints and arguments into a single, cohesive, and final answer, $A_{\text{final}}$.

The judge's decision-making process can be modeled as a summary function $J$. This function takes the set of all final answers from the debaters, $S_{N-1}$, as its input and produces the final global answer.
\begin{equation}
    A_{\text{final}} = J(S_{N-1}) = J(\{A_{1,N-1}, A_{2,N-1}, \dots, A_{M,N-1}\})
\end{equation}

In our experiment, we arrange three debaters with one round of opening statements, two rounds of free debate, and one round of concluding remarks. The relevant prompts in inference process have been shown in Listing~\ref{lst:debate_prompt_1}, \ref{lst:debate_prompt_2} and \ref{lst:debate_prompt_3}.


\begin{figure*}
  \centering
    \includegraphics[width=\textwidth]{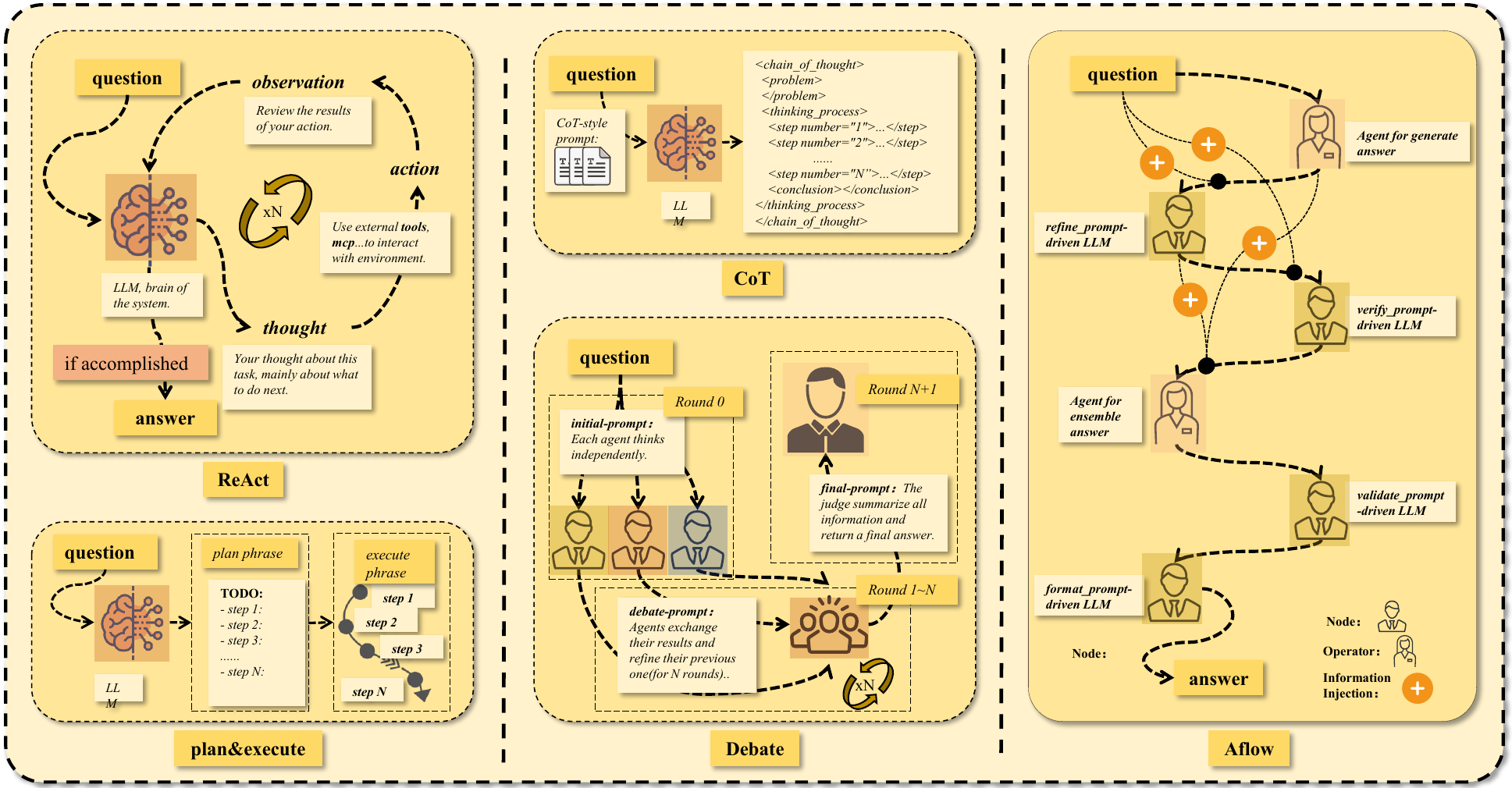}
    \caption{\textbf{Illustration of 5 Agent Architectures.} We show the inference processes of ReAct, Plan\&Execute, CoT, Debate and AFlow.}
    \label{agent_arch}
\end{figure*}

\subsection{AFlow Inference Details}
\label{sec:aflow_detail}

AFlow is a novel framework that leverages LLMs as optimizers within a variant of Monte Carlo Tree Search (MCTS) to search for optimal workflows. Generally, the input of this system is a set of samples (like training data), which involves question set and metrics. After learning and optimization, it outputs an agentic workflow that works best for this type of data, which is a python script.

We used the recommended setup to configure our own workflow. In detail,following recommendation, we set \texttt{Sample count} as 4, \texttt{Max iteration rounds} as 20, \texttt{Validation rounds} as 5. It cost us approximately 44 US dollars to complete the learning process and get an optimized workflow as Figure~\ref{agent_arch}. Core code generated by AFlow is shown in Listing~\ref{AFlow_output}. The relevant prompts generated by AFlow in this inference process have been shown in Listing~\ref{lst:REFINE_PROMPT}, \ref{lst:VERIFY_PROMPT}, \ref{lst:VALIDATE_FORMAT_PROMPT}, \ref{lst:ANSWER_GENERATION_PROMPT}, and \ref{lst:SC_ENSEMBLE_PROMPT}.

\begin{lstlisting}[
    style=codestyle, 
    caption={Selected source code generated by AFlow.},
    label={AFlow_output}
]
 async def __call__(self, problem: str):
    """
    Implementation of the workflow
    """
    # Generate initial solution
    initial_solution = await self.answer_generate(input=problem)
    
    # Get refined solution with custom operator
    refined = await self.custom(input=problem + f"\nInitial solution: {initial_solution['response']}", instruction=prompt_custom.REFINE_PROMPT)
    
    # Verify essential tools
    verified = await self.custom(input=problem + f"\nCurrent solution: {refined['response']}", instruction=prompt_custom.VERIFY_PROMPT)
    
    # Ensemble the solutions
    solutions = [initial_solution['response'], refined['response'], verified['response']]
    ensemble_solution = await self.sc_ensemble(solutions=solutions)
    
    # Validate format and preprocessing steps
    validated = await self.custom(input=ensemble_solution['response'], instruction=prompt_custom.VALIDATE_FORMAT_PROMPT)
    
    # Final formatting
    formatted = await self.custom(input=validated['response'], instruction=prompt_custom.FORMAT_PROMPT)
    
    return formatted['response'], self.llm.get_usage_summary()["total_cost"]
\end{lstlisting}

\subsection{Prompts and Constants}
\subsubsection{Domain Information}

\begin{lstlisting}[
    style=promptstyle, 
    caption={Domain category.},
    label={Domain category}
]
DOMAINS=["Agriculture & Forestry", "Urban & Regional Planning", "Environmental Monitoring & Climate Change", "Disaster Emergency & Management", "Earth Science & Resource Exploration", "Marine & Water Resources", "Defense & Security"]
\end{lstlisting}

\begin{lstlisting}[
    style=promptstyle, 
    caption={Complexity level.},
    label={Complexity level}
]
COMPLEXITIES=["Simple", "Medium", "Complex"]
\end{lstlisting}

\begin{lstlisting}[
    style=promptstyle, 
    caption={Desciptions for each domain, applied in Task Dependency Graph Generation.},
    label={Desciptions for each domain}
]
DOMAIN_DESCRIPTIONS = {
"Agriculture & Forestry": """
 Crop type identification and area statistics
 Crop growth monitoring and stress analysis (drought, flood, pests)
 Yield estimation
 Forest resource survey (species identification, volume estimation)
 Forest fire monitoring and damage assessment
 Deforestation and reforestation monitoring
""",
"Urban & Regional Planning": """
 Urban expansion monitoring
 Land use/land cover (LULC) classification and change analysis
 Building extraction and 3D modeling
 Transportation network planning (road extraction, traffic flow estimation)
 Urban green space and water body monitoring
 Urban heat island effect analysis
""",
"Environmental Monitoring & Climate Change": """
 Water quality monitoring (eutrophication, suspended matter)
 Air quality monitoring (aerosols, pollutants)
 Soil erosion and desertification monitoring
 Glacier, snow, and sea ice change monitoring
 Carbon storage estimation and carbon cycle research
""",
"Disaster Emergency & Management": """
 Flood inundation extent assessment
 Earthquake-damaged building identification
 Landslide and debris flow hazard investigation
 Drought monitoring
 Typhoon path and impact analysis
""",
"Earth Science & Resource Exploration": """
 Geological structure interpretation
 Mineral resource exploration (alteration mineral identification)
 Oil and gas exploration (hydrocarbon seepage, surface micro-geomorphology)
 Groundwater resource investigation
""",
"Marine & Water Resources": """
 Coastline change monitoring
 Sea surface temperature, salinity, and chlorophyll concentration retrieval
 Maritime vessel and oil spill detection
 Algal bloom and red tide monitoring
 Watershed water resource assessment
""",
"Defense & Security": """
 Sensitive target identification (airports, ports, missile bases)
 Border activity monitoring
 Battlefield situational awareness
 Camouflage and counter-camouflage identification
"""
}
\end{lstlisting}
\begin{lstlisting}[
    style=promptstyle, 
    caption={Keywords for each domain, used for Domain Relevance Check during Task Validation.},
    label={Keywords for each domain}
]
DOMAIN_KEYWORDS= {
"Agriculture & Forestry": [
"crop", "agriculture", "forest", "vegetation", "farming", "harvest", "yield", "deforestation", "plant", "soil",
"agricultural", "forestry", "cultivation", "irrigation", "pesticide", "fertilizer", "biomass", "photosynthesis",
"canopy", "landsat", "ndvi", "chlorophyll", "phenology", "agroforestry", "silviculture", "timber", "logging",
"reforestation", "afforestation", "carbon sequestration", "biodiversity", "ecosystem", "habitat", "wildlife",
"pasture", "grassland", "rangeland", "livestock", "grazing", "drought", "precipitation", "evapotranspiration",
"leaf area index", "lai", "gpp", "npp", "biomass estimation", "tree height", "dbh", "forest inventory"
],
"Urban & Regional Planning": [
"urban", "city", "building", "infrastructure", "development", "planning", "construction", "residential",
"metropolitan", "suburban", "downtown", "commercial", "industrial", "zoning", "land use", "lulc", "impervious",
"built-up", "settlement", "population", "density", "sprawl", "expansion", "growth", "transportation", "road",
"highway", "street", "transit", "mobility", "accessibility", "pedestrian", "traffic", "congestion", "parking",
"housing", "neighborhood", "district", "block", "parcel", "footprint", "elevation", "dem", "dsm", "3d modeling",
"cadastral", "property", "real estate", "gentrification", "sustainability", "smart city", "green space"
],
"Environmental Monitoring & Climate Change": [
"environment", "climate", "pollution", "air quality", "water quality", "temperature", "carbon", "emission",
"environmental", "atmospheric", "greenhouse gas", "co2", "methane", "aerosol", "particulate", "pm2.5", "pm10",
"ozone", "nitrogen", "sulfur", "acid rain", "smog", "visibility", "turbidity", "eutrophication", "algal bloom",
"contamination", "toxic", "heavy metal", "pesticide residue", "ph", "dissolved oxygen", "bod", "cod",
"thermal pollution", "noise pollution", "radiation", "uv", "solar", "albedo", "reflectance", "emissivity",
"evaporation", "humidity", "precipitation", "wind speed", "atmospheric pressure", "meteorological", "weather",
"global warming", "climate change", "sea level rise", "ice melting", "permafrost", "desertification"
],
"Disaster Emergency & Management": [
"disaster", "emergency", "flood", "earthquake", "fire", "damage", "evacuation", "rescue", "hazard",
"natural disaster", "catastrophe", "calamity", "crisis", "risk assessment", "vulnerability", "resilience",
"wildfire", "forest fire", "bushfire", "burn scar", "burn severity", "seismic", "tsunami", "landslide",
"mudslide", "avalanche", "hurricane", "typhoon", "cyclone", "tornado", "storm", "lightning", "hail",
"drought", "famine", "volcanic", "eruption", "lava", "ash", "pyroclastic", "debris flow", "rockfall",
"subsidence", "sinkhole", "erosion", "inundation", "surge", "breach", "levee", "dam failure", "structural damage",
"infrastructure damage", "building collapse", "road closure", "bridge failure", "power outage", "communication"
],
"Earth Science & Resource Exploration": [
"geological", "mineral", "resource", "exploration", "mining", "oil", "gas", "groundwater", "geology",
"geophysical", "geochemical", "petrology", "mineralogy", "stratigraphy", "tectonics", "fault", "fracture",
"lithology", "sedimentary", "igneous", "metamorphic", "ore", "deposit", "vein", "alteration", "hydrothermal",
"geothermal", "petroleum", "hydrocarbon", "reservoir", "basin", "anticline", "syncline", "unconformity",
"aquifer", "water table", "hydraulic conductivity", "porosity", "permeability", "well", "borehole", "core",
"outcrop", "exposure", "topography", "elevation", "slope", "aspect", "drainage", "watershed", "catchment",
"lineament", "magnetic", "gravity", "electromagnetic", "seismic", "reflection", "refraction", "spectral analysis"
],
"Marine & Water Resources": [
"marine", "ocean", "water", "coastal", "sea", "maritime", "vessel", "algae", "salinity",
"oceanographic", "hydrographic", "bathymetry", "seafloor", "seabed", "continental shelf", "abyssal", "benthic",
"pelagic", "littoral", "estuary", "delta", "lagoon", "bay", "gulf", "strait", "channel", "reef", "coral",
"mangrove", "wetland", "marsh", "swamp", "tide", "tidal", "current", "wave", "swell", "upwelling",
"chlorophyll", "phytoplankton", "zooplankton", "biomass", "productivity", "eutrophication", "red tide",
"oil spill", "pollution", "sediment", "turbidity", "suspended matter", "dissolved organic matter", "nutrients",
"phosphate", "nitrate", "silicate", "sea surface temperature", "sst", "sea level", "altimetry", "fisheries",
"aquaculture", "shipping", "navigation", "port", "harbor", "anchorage", "buoy", "lighthouse"
],
"Defense & Security": [
"defense", "security", "military", "surveillance", "border", "target", "threat", "reconnaissance",
"intelligence", "strategic", "tactical", "operational", "battlefield", "combat", "warfare", "conflict",
"base", "facility", "installation", "compound", "perimeter", "checkpoint", "patrol", "monitoring",
"radar", "sonar", "satellite", "imagery", "aerial", "drone", "uav", "aircraft", "helicopter", "fighter",
"bomber", "transport", "runway", "airfield", "airport", "hangar", "vehicle", "tank", "armored", "convoy",
"missile", "rocket", "launcher", "artillery", "ammunition", "depot", "storage", "bunker", "fortification",
"camouflage", "concealment", "detection", "identification", "tracking", "movement", "deployment", "logistics",
"supply", "communication", "signal", "electronic", "cyber", "homeland", "counterterrorism", "peacekeeping"
]
}
\end{lstlisting}
\subsubsection{Task Construction}
\begin{lstlisting}[
    style=promptstyle, 
    caption={Prompt for creating task dependency template with the format of DAG given domain information and task complexity.},
    label={create dag}
]
DAG_TEMPLATE_PROMPT = """Generate a typical task dependency template for remote sensing {domain} domain, represented as a Directed Acyclic Graph (DAG):

Domain application scope: {domain_desc}

Available tools:
{tools_str}

Requirements:
1. Select {tools_number_range} most relevant tools based on actual needs in {domain} domain
2. Define reasonable dependencies based on tool functionality and data flow
3. Form a DAG structure that follows actual remote sensing analysis workflow
4. Include complete analysis process from data acquisition to result output
5. Dependencies should reflect real remote sensing workflows
6. Focus on typical task requirements within the above domain scope

Output in JSON format:
{{
  "domain": "{domain}",
  "nodes": ["tool1", "tool2", "tool3", ...],
  "edges": [["source_tool_1", "target_tool_1"], ["source_tool_2", "target_tool_2"], ...],
  "description": "DAG template based on actual needs in {domain} domain"
}}"""
\end{lstlisting}
\begin{lstlisting}[
    style=promptstyle, 
    caption={Prompt for parameterizing extracted tool flow.},
    label={parameterize extracted tool flow}
]
PARAMETERIZE_FLOW_PROMPT = """Complete parameters for the tool flow to generate instantiated tool calls:

Tool sequence: {tools}

Tool details:
{tools_str}

Requirements:
1. Use imagination to complete specific parameters for each tool
2. Parameters should fit remote sensing application scenarios
3. Maintain logical consistency between parameters
4. Generate realistic and credible parameter values

Output in JSON format:
{{
  "parameterized_tools": [
    {{"tool": "tool1", "params": {{"param1": "value1", "param2": "value2"}}}},
    {{"tool": "tool2", "params": {{"param1": "value1", "param2": "value2"}}}}
  ]
}}"""
\end{lstlisting}
\begin{lstlisting}[
    style=promptstyle, 
    caption={Prompt for generating final task based on the parameterized tool flow.},
    label={generate final task}
]
GENERATE_TASK_PROMPT = """Generate a remote sensing analysis task based on the instantiated tool flow:

Tool call sequence:
{flow_str}

Requirements:
1. Question should be short and direct, focusing on one core objective
2. Leverage your imagination to generate a question that is highly contextual with specific entry point.
3. Do not explicitly mention any data sources
4. Do not mention specific technical methods or tools
5. Let agent infer what data and methods are needed
6. Question should have implicit complexity requiring deep analysis
7. Question should be specific and detailed, not a broad topic

Output concise core question as a string (one sentence):
"""
\end{lstlisting}
\subsubsection{Evaluation}
\begin{lstlisting}[
    style=promptstyle, 
    caption={Prompt for extracting key steps from ground truth tool trajectory.},
    label={extract key steps}
]
KEY_STEPS_EXTRACTION_PROMPT = """
Task question: {question}
Ground truth tool path: {ground_truth_tool_flow}

Please identify the key steps (indispensable steps) from the golden path to solve this problem.
Key steps should be core operations essential for task completion, removing optional or redundant steps.

Requirements:
1. Select the most critical steps from the given tool path
2. Key steps should be indispensable for task completion
3. Remove duplicate or redundant steps
4. Don't return an empty list

Return strictly in the following JSON format without any additional explanation:
{{
  "key_steps": ["step1", "step2", "step3"]
}}
"""
\end{lstlisting}
\begin{lstlisting}[
    style=promptstyle, 
    caption={Prompt for extracting key tools from agents' output.},
    label={extract key tools}
]
KEY_TOOLS_EXTRACTION_PROMPT="""
Task question: {question}
Agent's tool path: {agent_tool_flow}

Please identify the key tools (indispensable tools) from the agent's tool path to solve this problem.
Key tools should be core operations essential for task completion, removing optional or redundant tools.

Requirements:
1. Select the most critical different tools from the given tool path
2. Key tools should be indispensable for task completion
3. Remove duplicate or redundant tools
4. Don't return an empty list

Return strictly in the following JSON format without any additional explanation:
{{
  "key_tools": ["tool1", "tool2", "tool3"]
}}
"""
\end{lstlisting}
\begin{lstlisting}[
    style=promptstyle, 
    caption={Prompt for logic and completeness evaluation.},
    label={logic and completeness evaluation}
]
COMPLETENESS_EVALUATION_PROMPT = """
Task question: {question}

Please compare the completeness of the following two agents' tool calling paths:

Agent A ({agent_a}): {tool_flow_a}
Agent B ({agent_b}): {tool_flow_b}

Evaluation criteria:
1. Does it cover all key aspects required by the problem?
2. Are important data acquisition and processing steps missing?
3. Can it achieve the final results required by the problem?
4. Is the overall solution comprehensive?

Please choose one of the following options:
- "A": Agent A has better completeness
- "B": Agent B has better completeness
- "Tie": Both have equivalent completeness

Please **return only one string**: "A" or "B" or "Tie"
"""
\end{lstlisting}
\subsubsection{Architecture Implementation}
\label{Architecture Implementation}
\begin{lstlisting}[
    style=promptstyle, 
    caption={ReAct: Get thought prompt.},
    label={thought prompt}
]
f"""{self.system_prompt}

Available tools:
{tools_info}

User question: {query}

History:
{history}

Your previous_tool_calls: {previous_tool_calls}

**Strategic Thinking Guidelines**:
1. **Fresh Perspective**: If you've taken certain approaches before, explore different angles or complementary information apart from your previous_tool_calls
2. **Progress Assessment**: Evaluate what information you've gathered and identify any gaps
3. **Completion Check**: If you have sufficient information to provide a comprehensive answer, prepare to finish
4. **Diversification**: Avoid repeating similar tool calls - seek variety in your information gathering.The tool you will call next should be different from your previous_tool_calls

**Decision Framework**:
- If you have enough information, Think: "I have gathered sufficient information and can provide a comprehensive answer to this question."
- If you need more data, Think about what specific information is still missing and how to obtain it differently
- If you're unsure, Think about what would make you confident in your answer

**CRITICAL**: This is ONLY for thinking and analysis. Do NOT include "Action:" statements here.

Please start with "Thought: " and provide your strategic analysis in one clear sentence."""
\end{lstlisting}
\begin{lstlisting}[
    style=promptstyle, 
    caption={ReAct: Get action prompt.},
    label={action prompt}
]
f"""Based on your thinking,specifically based on your **previous_tool_calls**, decide the next action using function calling.

User question: {query}

Your thought: {thought}

History:
{history}

Your previous_tool_calls: {previous_tool_calls}

**Decision Rules**:
1. Carefully check **previous_tool_calls** to avoid repeating same tool call.The tool you will call next should be different from your previous_tool_calls
2. If sufficient information obtained to answer the question, do NOT call any function (this will trigger FINISH)
3. If must call tools, ensure clear distinction from previous calls (different tools or significantly different parameters)
4. If you find yourself possibly entering a loop, do NOT call any function

**Anti-loop check**: Review the last 2-3 actions in history, if similar tool call patterns found, do NOT call any function.

**Action Decision**:
- If you need more information: Call ONE function with appropriate parameters
- If you have enough information to answer: Make NO function call (this will finish the task)"""
\end{lstlisting}
\begin{lstlisting}[
    style=promptstyle, 
    caption={ReAct: Get tool result prompt.},
    label={tool result prompt}
]
f"""
    You need to **fully imagine** the execution result of this tool based on the current tool's name, description and parameters. The result should be beneficial for solving the problem. Only describe in one sentence what the tool did, this sentence should be in past tense.
    Current scenario: {thought}
    Tool name: {tool_name}
    Tool description: {self.tools[tool_name]["description"]}
    Tool parameters: {args}
    """
\end{lstlisting}
\begin{lstlisting}[
    style=promptstyle, 
    caption={Plan\&Execute: Get planning prompt.},
    label={planning prompt}
]
f"""You are an intelligent assistant that needs to solve user problems.

Available tools:
{tools_info}

User question: {query}

Please analyze the user's question and create an execution plan. The plan is a tool chain that you need to carefully design to make the tool chain can solve the user's question properly. Decide the length of the tool chain based on the difficulty of the problem.

Please return the plan in JSON format as follows:
{{
    "plan": [
        {{
            "tool": "tool_name_1",
            "parameters": {{"param_name": "param_value"}},
        }},
        {{
            "tool": "tool_name_2",
            "parameters": {{"param_name": "param_value"}},
        }}
    ]
}}

        """
\end{lstlisting}
\begin{lstlisting}[
    style=promptstyle, 
    caption={CoT: Get chain-of-thought prompt.},
    label={chain-of-thought prompt}
]
f"""
You are an intelligent assistant that needs to solve user problems based on the tools provided.
User question: {query}
Available tools:
{tools_info}
Please think step by step,in each step,you should choose one name of these tools to call.
Output in this format strictly:
step1:thought_1;tool_1_name
step2:thought_2;tool_2_name
...
stepn:thought_n;tool_n_name
"""
\end{lstlisting}
\begin{lstlisting}[
    style=promptstyle, 
    caption={Debate: Prompt for Round 0 - independent thinking.},
    label={lst:debate_prompt_1}
]
f"""You are participating in a multi-agent debate. In this initial round, please think independently and provide your first answer to the question without seeing other agents' responses.

Question: {question}

Available tools:
{tools_info}

Please analyze the user's question and create an execution plan and corresponding initial_tool_trajectory. The plan is a paragraph of statement,the initial_tool_trajectory is a list of tool names which can solve the question properly. Decide the length of the tool chain based on the difficulty of the problem.

Please return the plan in JSON format as follows:
{{
    "plan": your plan to solve the question,
    "initial_tool_trajectory": ["tool_name_1","tool_name_2",...]
}}
"""
\end{lstlisting}
\begin{lstlisting}[
    style=promptstyle, 
    caption={Debate: Prompt for Round 1 to N - debate rounds.},
    label={lst:debate_prompt_2}
]
f"""You are number {debater_index} debater participating in a multi-agent debate. You have seen all previous responses and now need to provide your perspective.

Question: {question}

your former response: {former_response}

Available tools:
{tools_info}

Debate History:
{history_str}

Based on the previous responses, please analyze the strengths and weaknesses of previous arguments and improve your former tool trajectory.Return a breif advice and the refined tool trajectory in JSON format as follows:
{{  "advice": "your short advice to improve your former tool trajectory",
    "refined_tool_trajectory": ["tool_name_1","tool_name_2",...]
}}
"""
\end{lstlisting}
\begin{lstlisting}[
    style=promptstyle, 
    caption={Debate: Final prompt for summary round.},
    label={lst:debate_prompt_3}
]
f"""You are a judge/summarizer for a multi-agent debate. Please analyze the entire debate and return a final tool trajectory.

Original Question: {question}

Available tools:
{tools_info}

Complete Debate History:
{history_str}

Please return a final tool_trajectory.Follow the json format strictly and do not include any other text:

Json format:
{{
    "final_tool_trajectory": ["tool_name_1","tool_name_2",...]
}}

"""
\end{lstlisting}
\begin{lstlisting}[
    style=promptstyle, 
    caption={EarthAgent: Prompt for selecting sub-expert from layer 3.},
    label={selecting sub-expert from layer 3}
]
f"""You are an expert selection assistant. Select the most suitable third layer expert based on user questions.

The third layer experts are responsible for integration and output, handling more specific domain problems. Available experts:
{layer3_agents_info}

User question: {query}

Selection logic: Choose a third layer agent based on the industry and domain of the question. Select one from the third layer. For simple questions or daily Q&A, choose generalChatBotAgent.

Please return selection result in JSON format:
{{
    "selected_agent": "expert_name",
    "subtask": "specific subtask description that this expert needs to complete."
}}"""
\end{lstlisting}
\begin{lstlisting}[
    style=promptstyle, 
    caption={EarthAgent: Prompt for selecting sub-expert from layer 2.},
    label={selecting sub-expert from layer 2}
]
f"""You are an expert selection assistant. Based on user questions and layer 3 expert selection, choose layer 2 experts.

Layer 2 experts are responsible for image understanding work. Available experts:
{layer2_agents_info}

User question: {query}
Layer 3 selected expert: {layer3_selection.get('selected_agent')}
Layer 3 expert task: {layer3_selection.get('subtask')}

Selection logic: Based on the question and layer 3 agent, think about what visual operations need to be performed on remote sensing images to solve the problem, then select 1 to 3 agents from layer 2 as needed.

Please return selection result in JSON format:
{{
"selected_agents": [
{{
    "name": "expert_name_1",
    "subtask": "specific subtask description this expert needs to complete."
}},
{{
    "name": "expert_name_2", 
    "subtask": "specific subtask description this expert needs to complete."
}}
]
}}"""
\end{lstlisting}
\begin{lstlisting}[
    style=promptstyle, 
    caption={EarthAgent: Prompt for selecting sub-expert from layer 1.},
    label={selecting sub-expert from layer 1}
]
f"""You are an expert selection assistant. Based on user questions and layer 2 & layer 3 expert selections, choose layer 1 experts.

Layer 1 experts are responsible for data acquisition and preprocessing. Available experts:
{layer1_agents_info}

User question: {query}
Layer 2 selected experts: {layer2_info}
Layer 3 selected expert: {layer3_selection.get('selected_agent')}

Selection logic: Combine the question with layer 2 and layer 3 agent selections, see if the question contains remote sensing data, determine whether to additionally acquire remote sensing data or images, and determine whether data preprocessing is needed (if needed, the question will hint that existing data has defects), select 1-2 layer 1 agents.

Please return selection result in JSON format:
{{
    "selected_agents": [
        {{
            "name": "expert_name_1",
            "subtask": "specific subtask description this expert needs to complete."
        }},
        {{
            "name": "expert_name_2",
            "subtask": "specific subtask description this expert needs to complete."
        }}
    ]
}}"""
\end{lstlisting}
\begin{lstlisting}[
    style=promptstyle, 
    caption={AFlow: Prompt for refining answer.},
    label={lst:REFINE_PROMPT}
]
REFINE_PROMPT = """
Given the problem and initial solution, output a refined tool flow as a list of tool names that can solve the problem. The output must be in the format: ['tool1_name','tool2_name',...].

Consider:
1. The logical sequence of tools
2. Required preprocessing steps
3. Necessary analysis tools
4. Data formatting and output tools

Only output the tool flow list, no other text.
"""
\end{lstlisting}
\begin{lstlisting}[
    style=promptstyle, 
    caption={AFlow: Prompt for verifying answer.},
    label={lst:VERIFY_PROMPT}
]
VERIFY_PROMPT = """
Review the current solution and ensure all essential tools for the task type are included. Consider:
1. Data acquisition tools (web_search, download_satellite_imagery)
2. Preprocessing tools (geometric_correction, atmospheric_correction)
3. Analysis tools specific to the task
4. Data formatting and reporting tools

Add any missing essential tools and maintain proper sequence.
Output only the complete tool list in format: ['tool1_name','tool2_name',...]
"""
\end{lstlisting}
\begin{lstlisting}[
    style=promptstyle, 
    caption={AFlow: Prompt for validating answer.},
    label={lst:VALIDATE_FORMAT_PROMPT}
]
VALIDATE_FORMAT_PROMPT = """
Validate and correct the tool list to ensure:
1. All remote sensing tasks must include download_satellite_imagery, geometric_correction, atmospheric_correction in correct order
2. Must include cloud_mask_removal after corrections if using satellite imagery
3. Must end with format_data or generate_analysis_reports
4. List must be in exact format: ['tool1_name','tool2_name',...]

Output the corrected list maintaining the exact format, no other text.
"""
\end{lstlisting}
\begin{lstlisting}[
    style=promptstyle, 
    caption={AFlow: Prompt for formatting answer.},
    label={lst:FORMAT_PROMPT}
]
FORMAT_PROMPT = """
Format the tool flow into a proper Python list string that matches the expected format.
Rules:
1. Must be in exact format: ['tool1_name','tool2_name',...]
2. No spaces between commas
3. Single quotes for strings
4. Remove any extra whitespace or newlines
5. Include essential tools like format_data or generate_analysis_reports at the end

Only output the formatted list string, no other text.
"""
\end{lstlisting}
\begin{lstlisting}[
    style=promptstyle, 
    caption={AFlow: Prompt for generating answer.},
    label={lst:ANSWER_GENERATION_PROMPT}
]
ANSWER_GENERATION_PROMPT = """
You are an expert in remote sensing task planning. Your goal is to generate an appropriate sequence of tools/operations to solve the given remote sensing task.

Think step by step about what tools and operations are needed:
1. Consider the type of remote sensing task (agriculture, disaster management, environmental monitoring, etc.)
2. Think about the logical sequence of operations (data acquisition, preprocessing, analysis, output)
3. Select appropriate tools for each step

Available tool categories include:
- Data acquisition: download_satellite_imagery, web_search, recommend_satellite_platforms
- Preprocessing: geometric_correction, atmospheric_correction, cloud_mask_removal, crop_image
- Analysis: classify_land_cover, segment_water_bodies, monitor_crop_health, detect_plant_diseases, assess_disaster_damage
- Output: generate_analysis_reports, format_data, statistical_analysis

Task: {input}

In the "thought" field, explain your step-by-step reasoning process.
In the "answer" field, provide a clear sequence of tools/operations needed to complete the task.
"""
\end{lstlisting}
\begin{lstlisting}[
    style=promptstyle, 
    caption={AFlow: Prompt for ensenbling answer.},
    label={lst:SC_ENSEMBLE_PROMPT}
]
SC_ENSEMBLE_PROMPT = """
Several tool flow solutions have been generated for the same remote sensing task planning problem. They are as follows:
{solutions}

Identify the most consistent and appropriate tool flow that appears most frequently across them. Consider the logical sequence of remote sensing operations and the specific requirements of the task.

In the "thought" field, provide a detailed explanation of your analysis process. In the "solution_letter" field, output only the single letter ID (A, B, C, etc.) corresponding to the most consistent solution. Do not include any additional text or explanation in the "solution_letter" field.
"""
\end{lstlisting}

\section{Further Analysis of Evaluation}
\subsection{Results on Each Sub-domain}
The whole tasks are divided into 7 sub-domains as we mentioned in Section~\ref{Task Generation Details}. In addition to analyzing the overall performance, we have also summarized the results on each sub-domain. Detailed information is shown in Table~\ref{detailed_results}.

\onecolumn
\newcolumntype{L}[1]{>{\raggedright\let\newline\\ \arraybackslash\hspace{0pt}}m{#1}}
\newcolumntype{C}[1]{>{\centering\let\newline\\
\arraybackslash\hspace{0pt}}m{#1}}
\newcolumntype{R}[1]{>{\raggedleft\let\newline\\
\arraybackslash\hspace{0pt}}m{#1}}

\begin{longtable}{@{} L{4.5cm} L{10.0cm} @{}}
\caption{\textbf{Detailed Experimental Results by Domain.} \textbf{Diff.} stands for \textbf{Difficulty}. \textbf{Arch.} stands for \textbf{Architecture}. \textbf{Rec.} stands for \textbf{$\text{Recall}_{key}$}. \textbf{Prec.} stands for \textbf{$\text{Precision}_{key}$}. \textbf{F1.} stands for \textbf{$\text{F1}_{key}$}. \textbf{Struct.} stands for \textbf{Structural}. \textbf{Hol.} stands for \textbf{Holistic}. }\\
\label{detailed_results}\\
\toprule
\textbf{Domain} & \textbf{Performance Metrics} \\
\midrule
\endfirsthead

\multicolumn{2}{c}%
{{\tablename\ \thetable{} -- continued from previous page}} \\
\toprule
\textbf{Domain} & \textbf{Performance Metrics} \\
\midrule
\endhead

\bottomrule
\multicolumn{2}{r}{{Continued on next page}} \\
\endfoot

\bottomrule
\endlastfoot

\multirow{12}{*}{\parbox{4cm}{\centering\textbf{Agriculture \& Forestry}}} &
\multicolumn{1}{p{9cm}}{
\scriptsize
\textit{This field focuses on the comprehensive monitoring of agricultural and forest resources to ensure sustainable management and food security. Key applications include identifying crop types and calculating area statistics, assessing crop health for stresses like drought or pests, and estimating yields. It also involves surveying forest resources by identifying tree species and estimating volume, alongside monitoring dynamic events such as forest fires, deforestation, and reforestation efforts, providing critical data for resource management and ecological preservation.}\newline

\footnotesize
\renewcommand{\arraystretch}{1}
\begin{tabular}{@{}l l cccccc@{}}
\toprule
\textbf{Diff.} & \textbf{Arch.} & \textbf{Rec.} & \textbf{Prec.} & \textbf{F1} & \textbf{Struct.} & \textbf{Hol.}  \\ 
\midrule
Simple & ReAct & 0.34 & 0.45 & 0.37 & 0.51 & 966.76  \\
Simple & Plan\&Execute & 0.46 & 0.66 & 0.52 & 0.49 & 965.45  \\

Simple & Debate & \underline{0.66} & 0.54 & 0.57 & \underline{0.67} & \underline{1025.30}  \\
Simple & CoT & 0.40 & 0.43 & 0.38 & 0.60 & 984.75  \\
Simple & AFlow & 0.54 & \textbf{0.76} & \underline{0.59} & \textbf{0.76} & 997.43  \\ 
Simple & EarthAgent & \textbf{0.83} & \underline{0.69} & \textbf{0.73} & 0.63 & \textbf{1070.02}  \\
\midrule
Medium & ReAct & 0.35 & 0.51 & 0.40 & 0.52 & 966.31  \\
Medium & Plan\&Execute & 0.43 & 0.63 & 0.48 & 0.49 & 965.16  \\

Medium & Debate & \underline{0.68} & 0.54 & \underline{0.58} & 0.62 & \underline{1025.29}  \\
Medium & CoT & 0.46 & 0.50 & 0.44 & 0.59 & 986.84  \\
Medium & AFlow & 0.54 & \textbf{0.65} & 0.55 & \textbf{0.68} & 991.75  \\ 
Medium & EarthAgent & \textbf{0.78} & \underline{0.63} & \textbf{0.68} & \textbf{0.68} & \textbf{1068.36}  \\
\midrule
Complex & ReAct & 0.30 & 0.55 & 0.38 & 0.43 & 973.12  \\
Complex & Plan\&Execute & 0.35 & 0.70 & 0.44 & 0.41 & 965.70  \\

Complex & Debate & \underline{0.55} & 0.54 & 0.52 & 0.59 & \underline{1022.78}  \\
Complex & CoT & 0.33 & 0.48 & 0.38 & 0.51 & 980.15 \\
Complex & AFlow & 0.52 & \textbf{0.76} & \underline{0.59} & \underline{0.65} & 992.73  \\ 
Complex & EarthAgent & \textbf{0.69} & 0.65 & \textbf{0.65} & \textbf{0.75} & \textbf{1066.42} \\
\bottomrule
\end{tabular}
\newline
} \\
\midrule

\multirow{12}{*}{\parbox{4cm}{\centering\textbf{Urban \& Regional Planning}}} &
\multicolumn{1}{p{9cm}}{
\scriptsize
\textit{This domain involves the detailed analysis and monitoring of urban and regional environments to guide sustainable development. It utilizes applications like tracking urban expansion, classifying land use and land cover changes over time, and extracting building data for 3D modeling. Additionally, it supports transportation network planning by extracting road data and estimating traffic flow, monitors vital urban green spaces and water bodies, and analyzes the urban heat island effect to improve city living conditions and environmental quality.} \newline

\footnotesize
\renewcommand{\arraystretch}{1}
\begin{tabular}{@{}l l cccccc@{}}
\toprule
\textbf{Diff.} & \textbf{Arch.} & \textbf{Rec.} & \textbf{Prec.} & \textbf{F1} & \textbf{Struct.} & \textbf{Hol.}  \\ \midrule
Simple & ReAct & 0.36 & 0.53 & 0.41 & 0.47 & 954.56 \\
Simple & Plan\&Execute & 0.43 & 0.52 & 0.46 & 0.53 & 972.19  \\

Simple & Debate & \underline{0.76} & 0.61 & \underline{0.65} & 0.71 & \underline{1029.89}  \\
Simple & CoT & 0.44 & 0.46 & 0.44 & 0.61 & 986.76  \\
Simple & AFlow & 0.53 & \textbf{0.75} & 0.60 & \textbf{0.78} & 996.66  \\
Simple & EarthAgent & \textbf{0.81} & \underline{0.71} & \textbf{0.74} & 0.62 & \textbf{1069.65}  \\
\midrule
Medium & ReAct & 0.33 & 0.49 & 0.38 & 0.45 & 959.11  \\
Medium & Plan\&Execute & \underline{0.40} & \textbf{0.54} & \underline{0.43} & 0.51 & 972.99 \\

Medium & Debate & 0.60 & 0.48 & 0.51 & 0.63 & \underline{1030.97}  \\
Medium & CoT & 0.37 & 0.44 & 0.38 & 0.57 & 981.33  \\
Medium & AFlow & 0.41 & 0.50 & 0.40 & \textbf{0.69} & 1000.64  \\ 
Medium & EarthAgent & \textbf{0.69} & \underline{0.65} & \textbf{0.65} & \underline{0.66} & \textbf{1069.27}  \\
\midrule
Complex & ReAct & 0.22 & 0.43 & 0.28 & 0.38 & 964.32  \\
Complex & Plan\&Execute & 0.25 & 0.47 & 0.30 & 0.40 & 965.95  \\

Complex & Debate & \underline{0.56} & 0.50 & \underline{0.51} & 0.58 & \underline{1029.34}  \\
Complex & CoT & 0.31 & 0.45 & 0.34 & 0.47 & 982.34  \\
Complex & AFlow & 0.41 & \underline{0.65} & 0.49 & \underline{0.62} & 989.44\\ 
Complex & EarthAgent & \textbf{0.70} & \textbf{0.72} & \textbf{0.70} & \textbf{0.77} & \textbf{1068.38}  \\
\bottomrule
\end{tabular}
\newline
} \\
\midrule

\multirow{12}{*}{\parbox{4cm}{\centering\textbf{Environmental Monitoring \& Climate Change}}} &
\multicolumn{1}{p{9cm}}{
\scriptsize
\textit{This area is dedicated to observing and analyzing environmental conditions to understand and address the impacts of climate change. Core applications include monitoring the quality of water for issues like eutrophication and air for aerosols and pollutants. It also involves tracking soil erosion and desertification, monitoring changes in glaciers, snow, and sea ice, and estimating carbon storage. This research is fundamental to understanding the Earth's carbon cycle and developing strategies to mitigate climate change.} \newline

\footnotesize
\renewcommand{\arraystretch}{1}
\begin{tabular}{@{}l l cccccc@{}}
\toprule
\textbf{Diff.} & \textbf{Arch.} & \textbf{Rec.} & \textbf{Prec.} & \textbf{F1} & \textbf{Struct.} & \textbf{Hol.} \\ \midrule
Simple & ReAct & 0.31 & 0.46 & 0.34 & 0.47 & 955.65  \\
Simple & Plan\&Execute & 0.35 & 0.46 & 0.38 & 0.48 & 979.42  \\

Simple & Debate & \textbf{0.66} & \textbf{0.53} & \textbf{0.56} & 0.68 & 1037.52 \\
Simple & CoT & 0.40 & 0.46 & 0.40 & 0.56 & 981.25  \\
Simple & AFlow & \underline{0.41} & 0.68 & \underline{0.47} & \textbf{0.74} & 1000.02  \\ 
Simple & EarthAgent & 0.55 & \underline{0.51} & 0.50 & \underline{0.63} & \underline{1062.75}  \\
\midrule
Medium & ReAct & 0.31 & 0.47 & 0.35 & 0.48 & 954.62 \\
Medium & Plan\&Execute & \underline{0.41} & 0.46 & 0.40 & 0.50 & 979.87  \\

Medium & Debate & 0.63 & 0.52 & 0.55 & 0.65 & \underline{1035.09} \\
Medium & CoT & 0.39 & 0.46 & \underline{0.41} & 0.57 & 981.03  \\
Medium & AFlow & 0.40 & \underline{0.57} & 0.43 & \textbf{0.70} & 989.76  \\ 
Medium & EarthAgent & \textbf{0.58} & \textbf{0.59} & \textbf{0.56} & \underline{0.66} & \textbf{1062.56}  \\
\midrule
Complex & ReAct & 0.38 & 0.66 & 0.46 & 0.41 & 956.66 \\
Complex & Plan\&Execute & 0.37 & 0.62 & 0.44 & 0.43 & 972.30  \\

Complex & Debate & \underline{0.66} & \textbf{0.67} & \underline{0.66} & 0.63 & \underline{1031.95}  \\
Complex & CoT & 0.42 & 0.60 & 0.47 & 0.51 & 981.33 \\
Complex & AFlow & 0.41 & \underline{0.70} & 0.49 & \underline{0.65} & 987.98 \\ 
Complex & EarthAgent & \textbf{0.61} & 0.65 & \textbf{0.61} & \textbf{0.75} & \textbf{1064.76}  \\

\bottomrule
\end{tabular}
\newline
} \\

\midrule

\multirow{12}{*}{\parbox{4cm}{\centering\textbf{Disaster Emergency \& Management}}} &
\multicolumn{1}{p{9cm}}{
\scriptsize
\textit{This field is focused on utilizing timely data for effective natural disaster response and management. Key applications include assessing the inundation extent of floods, identifying buildings damaged by earthquakes, and investigating hazards from landslides and debris flows. It also plays a crucial role in monitoring the onset and severity of droughts and analyzing typhoon paths and their potential impacts. This information is vital for emergency responders to make informed decisions, minimize damage, and save lives.} \newline

\footnotesize
\renewcommand{\arraystretch}{1}
\begin{tabular}{@{}l l cccccc@{}}
\toprule
\textbf{Diff.} & \textbf{Arch.} & \textbf{Rec.} & \textbf{Prec.} & \textbf{F1} & \textbf{Struct.} & \textbf{Hol.}  \\ \midrule
Simple & ReAct & 0.40 & 0.54 & 0.43 & 0.55 & 968.40 \\
Simple & Plan\&Execute & 0.45 & 0.60 & 0.49 & 0.50 & 965.78  \\

Simple & Debate & \textbf{0.71} & 0.54 & \underline{0.59} & 0.63 & \underline{1033.78}  \\
Simple & CoT & 0.54 & \underline{0.61} & 0.56 & 0.60 & 984.28  \\
Simple & AFlow & 0.33 & \textbf{0.66} & 0.39 & \textbf{0.69} & 980.19  \\
Simple & EarthAgent & 0.64 & \underline{0.61} & \textbf{0.61} & 0.60 & \textbf{1059.56}  \\
\midrule
Medium & ReAct & 0.38 & 0.60 & 0.43 & 0.51 & 968.33  \\
Medium & Plan\&Execute & 0.43 & 0.63 & 0.48 & 0.46 & 971.39 \\

Medium & Debate & \underline{0.65} & 0.67 & \underline{0.63} & 0.58 & \underline{1027.46} \\
Medium & CoT & 0.48 & 0.66 & 0.53 & 0.53 & 979.03  \\
Medium & AFlow & 0.38 & 0.59 & 0.42 & 0.62 & 987.93 \\ 
Medium & EarthAgent & \textbf{0.68} & \textbf{0.68} & \textbf{0.66} & \textbf{0.68} & \textbf{1062.86}  \\
\midrule
Complex & ReAct & 0.27 & 0.52 & 0.33 & 0.43 & 966.49 \\
Complex & Plan\&Execute & 0.26 & 0.53 & 0.34 & 0.38 & 965.60 \\

Complex & Debate & \underline{0.58} & 0.58 & \underline{0.56} & \underline{0.58} & \underline{1030.01}  \\
Complex & CoT & 0.35 & 0.63 & 0.43 & 0.46 & 980.55  \\
Complex & AFlow & 0.48 & \underline{0.70} & 0.52 & \underline{0.58} & 993.63 \\
Complex & EarthAgent & \textbf{0.69} & \textbf{0.77} & \textbf{0.71} & \textbf{0.75} & \textbf{1066.14}  \\
\bottomrule
\end{tabular}
\newline
} \\
\midrule

\multirow{12}{*}{\parbox{4cm}{\centering\textbf{Earth Science \& Resource Exploration}}} &
\multicolumn{1}{p{9cm}}{
\scriptsize
\textit{This domain applies advanced analysis to study the Earth's structure and discover natural resources. Applications include the detailed interpretation of geological structures and the identification of alteration minerals, which are key indicators for mineral resource exploration. It also aids in oil and gas exploration by detecting hydrocarbon seepage and analyzing surface micro-geomorphology. Furthermore, it supports the investigation of groundwater resources, contributing significantly to both geological science and the efficient management of valuable resources.} \newline

\footnotesize
\renewcommand{\arraystretch}{1}
\begin{tabular}{@{}l l cccccc@{}}
\toprule
\textbf{Diff.} & \textbf{Arch.} & \textbf{Rec.} & \textbf{Prec.} & \textbf{F1} & \textbf{Struct.} & \textbf{Hol.}  \\ \midrule
Simple & ReAct & 0.42 & 0.60 & 0.47 & 0.44 & 962.29  \\
Simple & Plan\&Execute & 0.49 & 0.67 & 0.55 & 0.42 & 972.48  \\

Simple & Debate & \underline{0.73} & 0.74 & \underline{0.72} & 0.59 & \underline{1031.79}  \\
Simple & CoT & 0.49 & 0.63 & 0.53 & 0.48 & 980.31  \\
Simple & AFlow & 0.19 & 0.56 & 0.23 & \textbf{0.68} & 987.04 \\ 
Simple & EarthAgent & \textbf{0.69} & \textbf{0.76} & \textbf{0.71} & \underline{0.72} & \textbf{1069.42}  \\
\midrule
Medium & ReAct & 0.35 & 0.51 & 0.38 & 0.46 & 962.85  \\
Medium & Plan\&Execute & \underline{0.41} & 0.58 & \underline{0.45} & 0.45 & 975.34 \\

Medium & Debate & \underline{0.58} & 0.60 & \underline{0.57} & 0.56 & \underline{1030.32}  \\
Medium & CoT & 0.38 & 0.57 & 0.42 & 0.51 & 980.32  \\
Medium & AFlow & 0.32 & 0.50 & 0.32 & \underline{0.61} & 984.86  \\ 
Medium & EarthAgent & \textbf{0.50} & \textbf{0.66} & \textbf{0.54} & \textbf{0.70} & \textbf{1065.79}  \\
\midrule
Complex & ReAct & 0.37 & 0.55 & 0.43 & 0.46 & 968.50 \\
Complex & Plan\&Execute & 0.44 & 0.58 & \underline{0.48} & 0.43 & 976.31  \\

Complex & Debate & \textbf{0.62} & \underline{0.62} & \textbf{0.59} & 0.56 & \underline{1032.06}  \\
Complex & CoT & 0.45 & 0.58 & \underline{0.48} & 0.47 & 970.09  \\
Complex & AFlow & 0.18 & 0.52 & 0.22 & \underline{0.58} & 974.12  \\ 
Complex & EarthAgent & \underline{0.50} & \textbf{0.60} & \underline{0.52} & \textbf{0.74} & \textbf{1065.20}  \\
\bottomrule
\end{tabular}
\newline
} \\
\midrule

\multirow{12}{*}{\parbox{4cm}{\centering\textbf{Marine \& Water Resources}}} &
\multicolumn{1}{p{9cm}}{
\scriptsize
\textit{ This area centers on the monitoring and management of marine and freshwater environments to protect and sustain aquatic ecosystems. Its applications include tracking coastline changes, retrieving crucial ocean data like sea surface temperature, salinity, and chlorophyll concentration, and detecting maritime vessels and oil spills for security and environmental protection. It also involves monitoring harmful algal blooms and red tides and conducting assessments of watershed water resources, which are essential for maintaining ecological balance and managing water supplies.}  \newline 

\footnotesize
\renewcommand{\arraystretch}{1}
\begin{tabular}{@{}l l cccccc@{}}
\toprule
\textbf{Diff.} & \textbf{Arch.} & \textbf{Rec.} & \textbf{Prec.} & \textbf{F1} & \textbf{Struct.} & \textbf{Hol.}\\ \midrule
Simple & ReAct & 0.29 & 0.43 & 0.33 & 0.47 & 961.47 \\
Simple & Plan\&Execute & 0.28 & 0.38 & 0.31 & 0.48 & 974.83\\

Simple & Debate & \underline{0.60} & 0.57 & \underline{0.56} & 0.65 & \underline{1036.68}  \\
Simple & CoT & 0.33 & 0.41 & 0.35 & 0.56 & 985.17  \\
Simple & AFlow & 0.37 & 0.65 & 0.43 & \textbf{0.74} & 1008.65 \\ 
Simple & EarthAgent & \textbf{0.64} & \textbf{0.74} & \textbf{0.66} & \underline{0.66} & \textbf{1062.88}  \\
\midrule
Medium & ReAct & 0.20 & 0.40 & 0.23 & 0.45 & 952.79  \\
Medium & Plan\&Execute & 0.28 & 0.38 & 0.29 & 0.48 & 987.34 \\

Medium & Debate & \underline{0.56} & 0.54 & \underline{0.52} & 0.61 & \underline{1032.44} \\
Medium & CoT & 0.35 & 0.46 & 0.36 & 0.56 & 980.47 \\
Medium & AFlow & 0.42 & \underline{0.51} & 0.40 & \textbf{0.69} & 1007.31 \\ 
Medium & EarthAgent & \textbf{0.58} & \textbf{0.58} & \textbf{0.56} & \underline{0.68} & \textbf{1064.90}  \\
\midrule
Complex & ReAct & 0.27 & 0.52 & 0.33 & 0.38 & 960.13  \\
Complex & Plan\&Execute & 0.27 & 0.48 & 0.31 & 0.43 & 984.57 \\

Complex & Debate & \underline{0.56} & 0.55 & \underline{0.52} & 0.55 & \underline{1036.25} \\
Complex & CoT & 0.30 & 0.49 & 0.34 & 0.47 & 975.57  \\
Complex & AFlow & 0.33 & \underline{0.53} & 0.36 & \underline{0.62} & 1001.18 \\
Complex & EarthAgent & \textbf{0.59} & \textbf{0.65} & \textbf{0.58} & \textbf{0.74} & \textbf{1060.78} \\
\bottomrule
\end{tabular}
\newline
} \\
\midrule

\multirow{12}{*}{\parbox{4cm}{\centering\textbf{Defense \& Security}}} &
\multicolumn{1}{p{9cm}}{
\scriptsize
\textit{This field leverages surveillance and analysis technologies for critical national security and intelligence purposes. The applications are strategically focused on identifying and monitoring sensitive targets such as airports, ports, and missile bases, as well as tracking border activities to prevent illicit crossings. It provides enhanced battlefield situational awareness for military operations and includes sophisticated techniques for identifying camouflage and counter-camouflage, thereby supporting tactical advantages and strengthening national defense capabilities against potential threats.} \newline

\footnotesize
\renewcommand{\arraystretch}{1}
\begin{tabular}{@{}l l cccccc@{}}
\toprule
\textbf{Diff.} & \textbf{Arch.} & \textbf{Rec.} & \textbf{Prec.} & \textbf{F1} & \textbf{Struct.} & \textbf{Hol.} \\ \midrule
Simple & ReAct & 0.26 & 0.38 & 0.30 & 0.46 & 960.80  \\
Simple & Plan\&Execute & \underline{0.35} & 0.49 & 0.40 & 0.50 & 981.18 \\

Simple & Debate & 0.58 & 0.49 & \underline{0.51} & 0.62 & \underline{1026.93}  \\
Simple & CoT & 0.27 & 0.35 & 0.29 & 0.52 & 973.01  \\
Simple & AFlow & 0.34 & \underline{0.60} & 0.41 & \textbf{0.69} & 985.56 \\ 
Simple & EarthAgent & \textbf{0.67} & \textbf{0.61} & \textbf{0.62} & \underline{0.67} & \textbf{1066.73}  \\
\midrule
Medium & ReAct & 0.29 & 0.54 & 0.36 & 0.44 & 972.12 \\
Medium & Plan\&Execute & 0.32 & 0.56 & 0.38 & 0.43 & 973.49 \\

Medium & Debate & \underline{0.49} & 0.60 & \underline{0.51} & 0.53 & \underline{1029.36}  \\
Medium & CoT & 0.27 & 0.44 & 0.30 & 0.49 & 974.87  \\
Medium & AFlow & 0.34 & \underline{0.65} & 0.41 & \underline{0.60} & 989.87  \\
Medium & EarthAgent & \textbf{0.56} & \textbf{0.67} & \textbf{0.57} & \textbf{0.70} & \textbf{1065.50}  \\
\midrule
Complex & ReAct & 0.26 & 0.48 & 0.31 & 0.42 & 967.01 \\
Complex & Plan\&Execute & 0.29 & 0.56 & 0.37 & 0.38 & 970.98  \\

Complex & Debate & \underline{0.41} & 0.54 & \underline{0.45} & 0.50 & \underline{1021.34} \\
Complex & CoT & 0.24 & 0.43 & 0.28 & 0.43 & 984.80  \\
Complex & AFlow & 0.35 & \underline{0.59} & 0.40 & \underline{0.58} & 998.31  \\ 
Complex & EarthAgent & \textbf{0.57} & \textbf{0.67} & \textbf{0.59} & \textbf{0.74} & \textbf{1067.02} \\
\bottomrule
\end{tabular}
\newline
} \\

\end{longtable}
\twocolumn

\subsection{Detailed Analysis}
\label{sec:more_analysis}

We provide a comprehensive analysis of the experimental data, exploring architecture behaviors, performance characteristics, and the impact of task properties.

\noindent
\textbf{Analysis of Tool Usage Preference and Strategy.} By analyzing the most frequently used tools for each architecture, we can infer their underlying problem-solving strategies and behavioral patterns. We use Average Standardized Position (where 0 indicates the beginning of a tool flow and 1 indicates the end) reveals the typical stage at which each tool is invoked. 

As shown in Table~\ref{tab:ReAct_tools}, ReAct heavily relies on data retrieval tools such as \texttt{web\_search} and \texttt{read\_database}. These tools are typically invoked in the middle of the workflow (average position 0.5), suggesting a strategy of continuous information gathering throughout the problem-solving process rather than a distinct initial data collection phase. 

The Plan\&Execute agent exhibits a highly structured, phased approach (Table~\ref{tab:pe_tools}). It consistently begins with data acquisition (\texttt{download\_satellite\_imagery}, position 0.02) and concludes with formatting and summarization tools (\texttt{format\_data} and \texttt{summarize\_text}, positions \>0.94). This pattern indicates a clear, logical ``gather-process-report'' workflow. 


EarthAgent, CoT, and AFlow exhibit distinct profiles. EarthAgent (Table \ref{tab:debate_tools}) follows a logical, expert-like workflow for remote sensing, prioritizing preprocessing steps such as correction and cloud masking after data download. CoT (Table \ref{tab:CoT_tools}) adheres to a classic ``research-then-conclude'' pattern, initiating with \texttt{web\_search} and concluding with \texttt{summarize\_text}. AFlow (Table \ref{tab:AFlow_tools}) demonstrates template-based behavior, consistently starting with image download and preprocessing, and ending with report generation.

\begin{table}[t]
\centering
\caption{Top 10 Most Frequently Used Tools by the ReAct Architecture.}
\label{tab:ReAct_tools}
    \resizebox{\linewidth}{!}{
\begin{tabular}{lcc}
\toprule
\textbf{Tool Name} & \textbf{Usage Frequency} & \textbf{Avg. Std. Position} \\
\midrule
\texttt{web\_search} & 687 & 0.527 \\
\texttt{read\_database} & 565 & 0.462 \\
\texttt{download\_satellite\_imagery} & 267 & 0.056 \\
\texttt{assess\_environmental\_impact} & 254 & 0.652 \\
\texttt{statistical\_analysis} & 249 & 0.735 \\
\texttt{analyze\_farming\_patterns} & 217 & 0.589 \\
\texttt{get\_weather\_data} & 209 & 0.362 \\
\texttt{correlation\_analysis} & 190 & 0.766 \\
\texttt{analyze\_geological\_structures} & 186 & 0.453 \\
\texttt{analyze\_threat\_patterns} & 164 & 0.579 \\
\bottomrule
\end{tabular}
}
\end{table}

\begin{table}[t]
\centering
\caption{Top 10 Most Frequently Used Tools by the Plan\&Execute Architecture.}
\label{tab:pe_tools}
\resizebox{\linewidth}{!}{
\begin{tabular}{lcc}
\toprule
\textbf{Tool Name} & \textbf{Usage Frequency} & \textbf{Avg. Std. Position} \\
\midrule
\texttt{summarize\_text} & 216 & 0.978 \\
\texttt{download\_satellite\_imagery} & 202 & 0.024 \\
\texttt{format\_data} & 168 & 0.947 \\
\texttt{correlation\_analysis} & 162 & 0.751 \\
\texttt{read\_database} & 157 & 0.148 \\
\texttt{analyze\_geological\_structures} & 139 & 0.291 \\
\texttt{monitor\_crop\_health} & 129 & 0.184 \\
\texttt{analyze\_threat\_patterns} & 125 & 0.604 \\
\texttt{classify\_land\_cover} & 99 & 0.431 \\
\texttt{statistical\_analysis} & 99 & 0.726 \\
\bottomrule
\end{tabular}
}
\end{table}

\begin{table}[t]
\centering
\caption{Top 10 Most Frequently Used Tools by the EarthAgent Architecture.}
\label{tab:tam_tools}
    \resizebox{\linewidth}{!}{
\begin{tabular}{lcc}
\toprule
\textbf{Tool Name} & \textbf{Usage Frequency} & \textbf{Avg. Std. Position} \\
\midrule
\texttt{download\_satellite\_imagery} & 1038 & 0.079 \\
\texttt{atmospheric\_correction} & 995 & 0.214 \\
\texttt{recommend\_satellite\_platforms} & 990 & 0.002 \\
\texttt{geometric\_correction} & 979 & 0.318 \\
\texttt{cloud\_mask\_removal} & 933 & 0.321 \\
\texttt{get\_weather\_data} & 923 & 0.280 \\
\texttt{monitor\_deforestation} & 718 & 0.520 \\
\texttt{detect\_urban\_expansion} & 669 & 0.492 \\
\texttt{statistical\_analysis} & 669 & 0.813 \\
\texttt{classify\_land\_cover} & 560 & 0.537 \\
\bottomrule
\end{tabular}
}
\end{table}

\begin{table}[t]
    \centering
    \caption{Top 10 Most Frequently Used Tools by the Debate Architecture.}
    \label{tab:debate_tools}
    \resizebox{\linewidth}{!}{
    \begin{tabular}{lcc}
    \toprule
    \textbf{Tool Name} & \textbf{Usage Frequency} & \textbf{Avg. Std. Position} \\
    \midrule
 \texttt{download\_satellite\_imagery} & 643 & 0.124 \\
 \texttt{statistical\_analysis} & 581 & 0.696 \\
 \texttt{correlation\_analysis} & 465 & 0.708 \\
 \texttt{atmospheric\_correction} & 367 & 0.207 \\
 \texttt{geometric\_correction} & 335 & 0.313 \\
 \texttt{get\_weather\_data} & 317 & 0.204 \\
 \texttt{cloud\_mask\_removal} & 306 & 0.329 \\
 \texttt{classify\_land\_cover} & 287 & 0.470 \\
 \texttt{summarize\_text} & 244 & 0.946 \\
 \texttt{read\_database} & 200 & 0.294 \\
    \bottomrule
    \end{tabular}
    }
\end{table}

\begin{table}[t]
    \centering
    \caption{Top 10 Most Frequently Used Tools by the CoT Architecture.}
    \label{tab:CoT_tools}
    \resizebox{\linewidth}{!}{
    \begin{tabular}{lcc}
    \toprule
    \textbf{Tool Name} & \textbf{Usage Frequency} & \textbf{Avg. Std. Position} \\
    \midrule
 \texttt{web\_search} & 1030 & 0.122 \\
 \texttt{summarize\_text} & 705 & 0.924 \\
 \texttt{format\_data} & 463 & 0.910 \\
 \texttt{statistical\_analysis} & 421 & 0.598 \\
 \texttt{correlation\_analysis} & 352 & 0.658 \\
 \texttt{download\_satellite\_imagery} & 289 & 0.164 \\
 \texttt{read\_database} & 246 & 0.216 \\
 \texttt{get\_weather\_data} & 233 & 0.243 \\
 \texttt{atmospheric\_correction} & 214 & 0.271 \\
 \texttt{classify\_land\_cover} & 151 & 0.569 \\
    \bottomrule
    \end{tabular}
    }
\end{table}

\begin{table}[t]
    \centering
    \caption{Top 10 Most Frequently Used Tools by the AFlow Architecture.}
    \label{tab:AFlow_tools}
    \resizebox{\linewidth}{!}{
    \begin{tabular}{lcc}
    \toprule
    \textbf{Tool Name} & \textbf{Usage Frequency} & \textbf{Avg. Std. Position} \\
    \midrule
 \texttt{download\_satellite\_imagery} & 998 & 0.002 \\
 \texttt{geometric\_correction} & 996 & 0.159 \\
 \texttt{atmospheric\_correction} & 996 & 0.317 \\
 \texttt{cloud\_mask\_removal} & 996 & 0.474 \\
 \texttt{generate\_analysis\_reports} & 986 & 0.974 \\
 \texttt{format\_data} & 611 & 0.898 \\
 \texttt{classify\_land\_cover} & 575 & 0.607 \\
 \texttt{monitor\_crop\_health} & 225 & 0.690 \\
 \texttt{assess\_disaster\_damage} & 198 & 0.727 \\
 \texttt{statistical\_analysis} & 154 & 0.799 \\
    \bottomrule
    \end{tabular}
    }
\end{table}

\begin{figure*}[t]
    \centering
    \includegraphics[width=\textwidth]{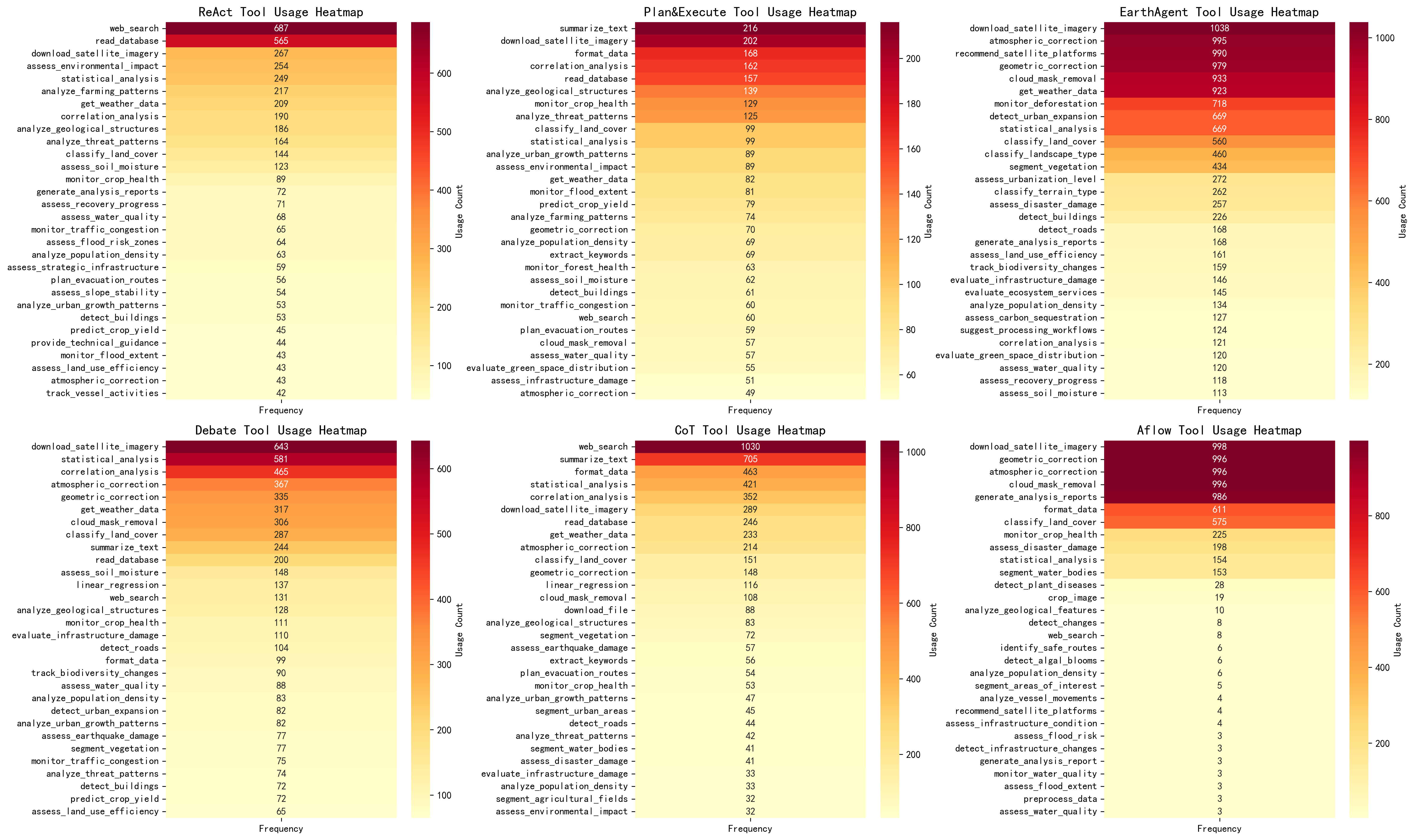}
    \caption{Heatmap of Tool Usage. The color scale ranges from dark (most frequently used tools) to light (least frequently used tools).}
    \label{fig:tool_heatmap}
\end{figure*}

\noindent
\textbf{Tool Usage Position Analysis via Heatmaps.} To visualize the strategic differences in tool positioning, we generated the heatmap that plots the average standardized position for the top tools of each agent, as shown in Figure \ref{fig:tool_heatmap}.

The heatmap clearly illustrates the behavioral patterns identified earlier. ReAct exhibits a steep transition from dark red to light red, demonstrating its narrow focus on a limited set of tools. In contrast, Plan\&Execute, Debate, and EarthAgent show a relatively gradual transition, indicating more balanced tool selection.

\begin{figure}[t]
    \centering

        \includegraphics[width=0.45\textwidth]{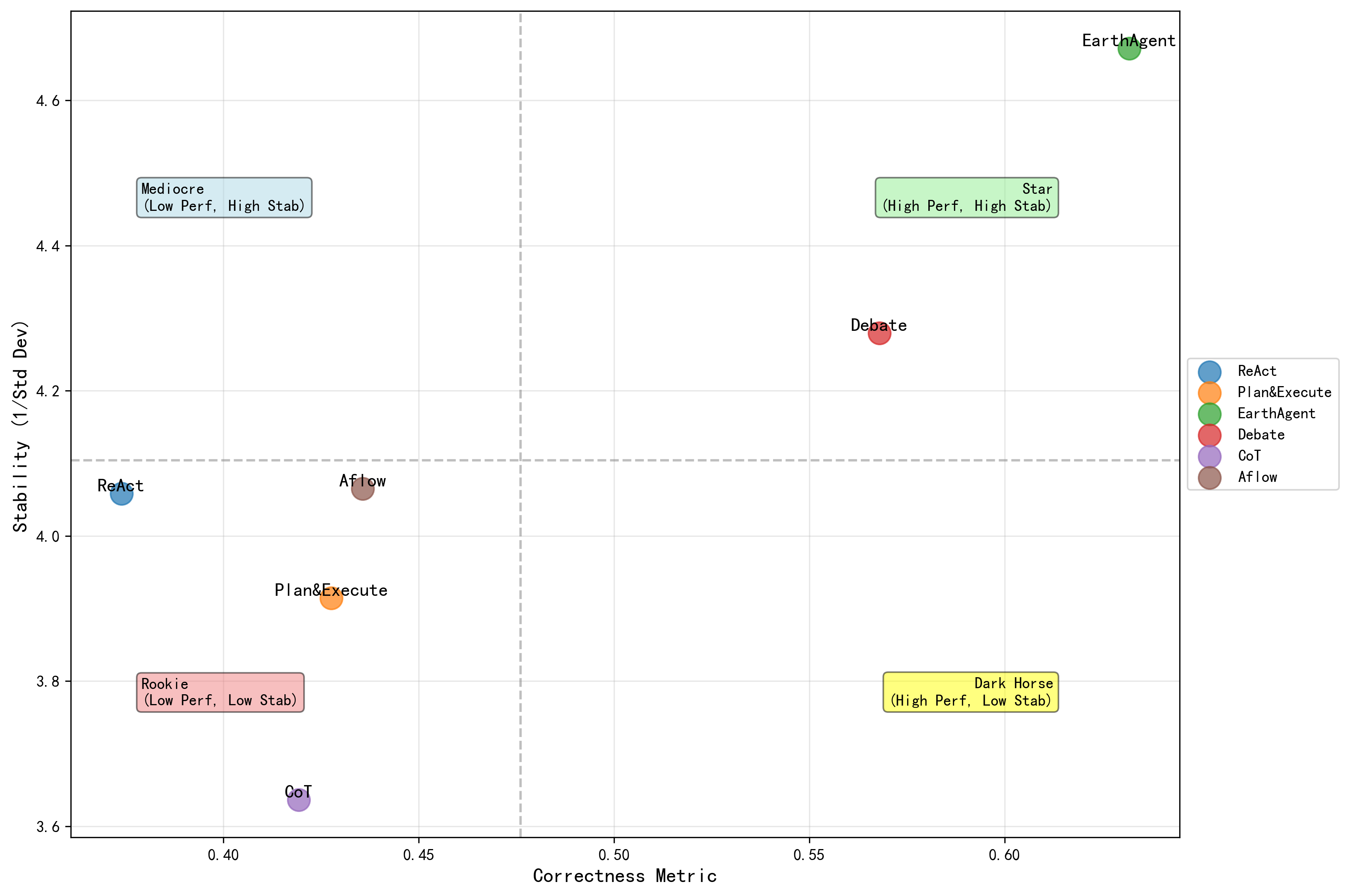}
        \caption{Quadrant Analysis based on Correctness Metric Score.}
        \label{fig:quad_f1}
\end{figure}
\begin{figure}[t]
    \centering
        \includegraphics[width=0.45\textwidth]{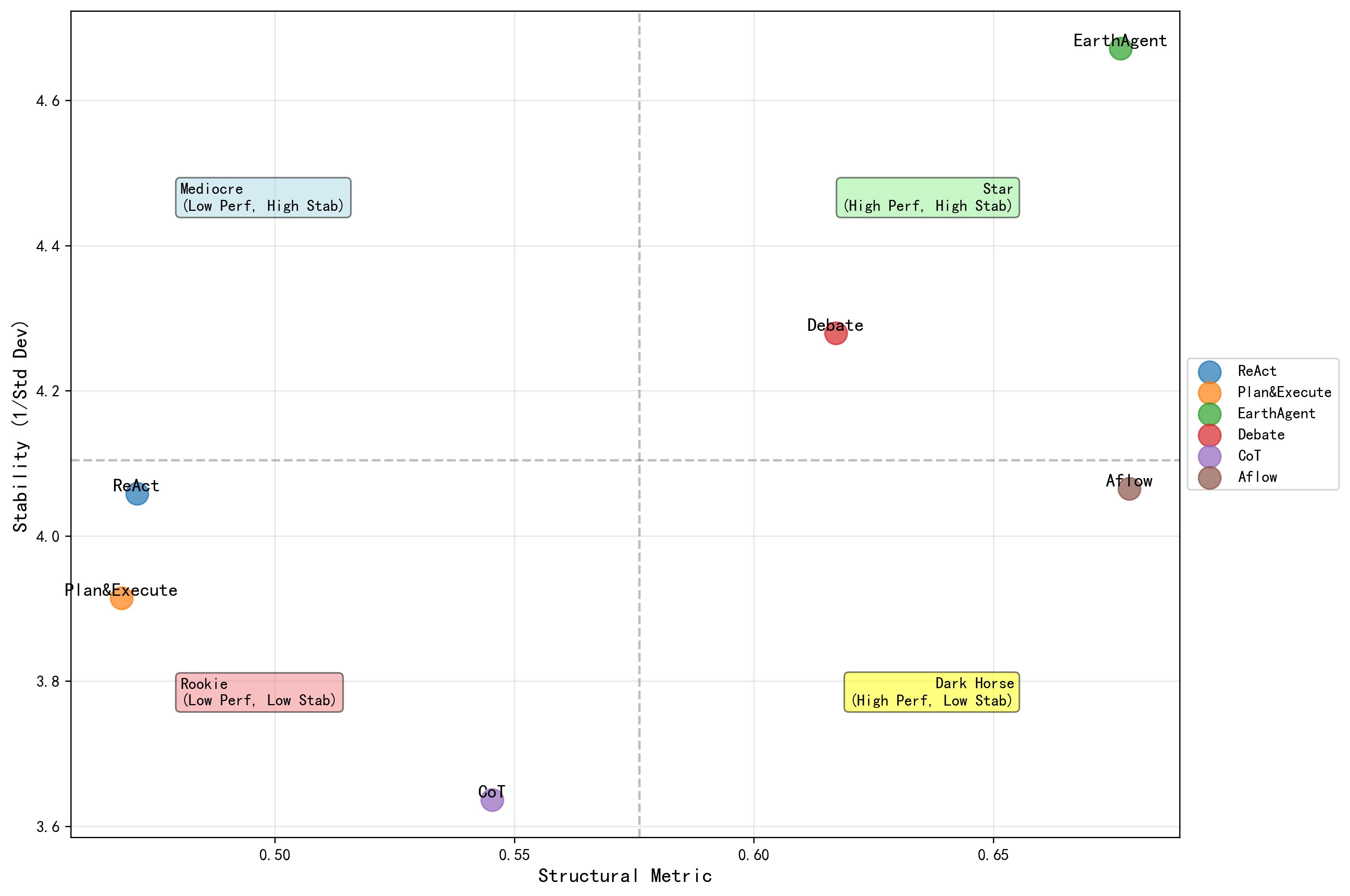}
        \caption{Quadrant Analysis based on Mean Structural Metrics Score.}
        \label{fig:quad_similarity}
\end{figure}
\begin{figure}[t]
    \centering
        \includegraphics[width=0.45\textwidth]{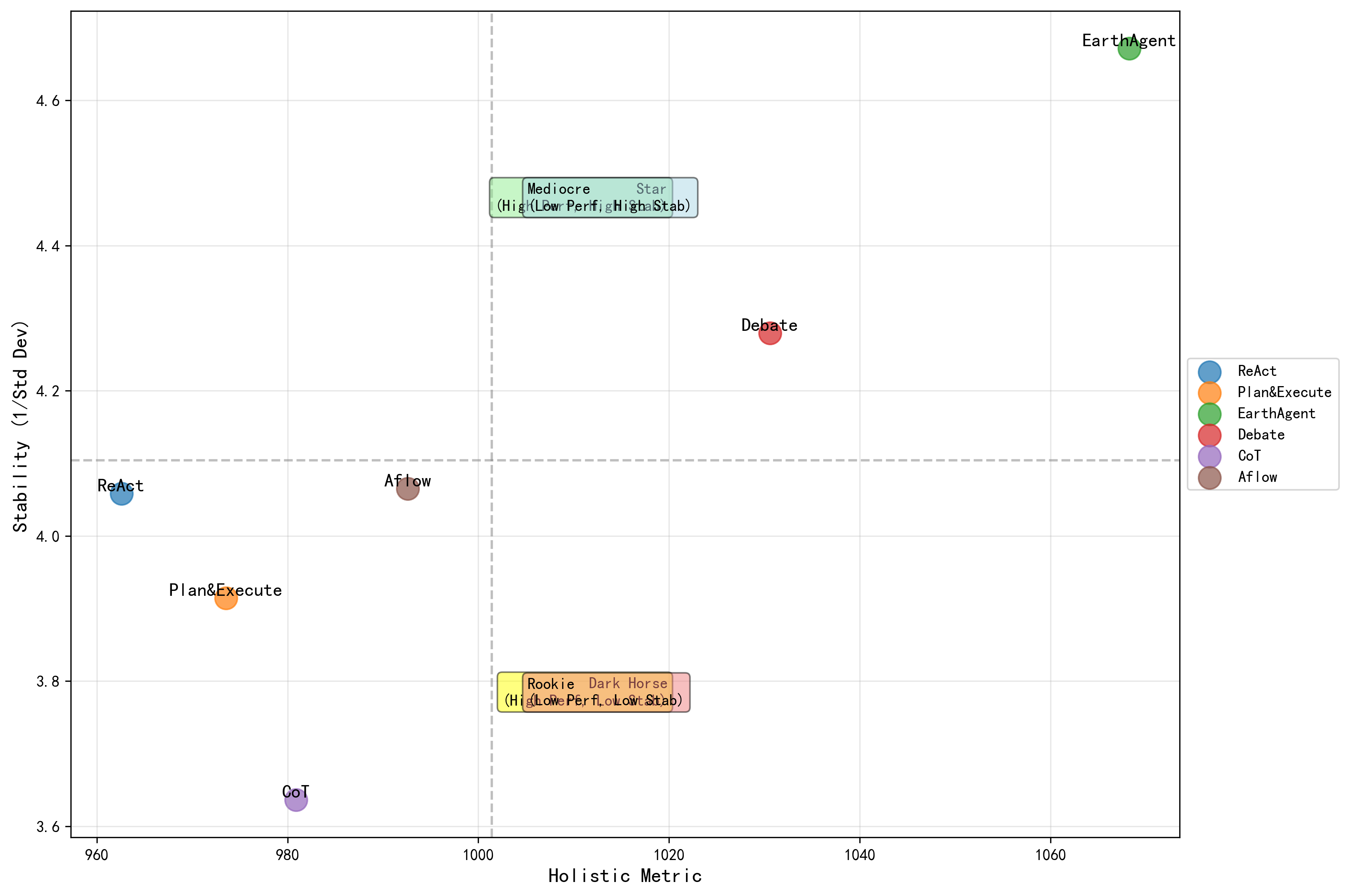}
        \caption{Quadrant Analysis based on Mean Completeness Score.}
        \label{fig:quad_completeness}
\end{figure}

\noindent
\textbf{Performance-Stability Quadrant Analysis.} We analyzed the agents by plotting their mean performance (X-axis) against their stability (Y-axis, defined as the inverse of the standard deviation). This creates four quadrants: High-Performance / High-Stability, High-Performance / Low-Stability (Inconsistent), Low-Performance / High-Stability (Reliably Poor), and Low-Performance / Low-Stability (Underperformer).
As shown in Figures \ref{fig:quad_f1}, \ref{fig:quad_similarity}, and \ref{fig:quad_completeness}, the analysis reveals different facets of each architecture's reliability. Based on F1 Score (Figure \ref{fig:quad_f1}), EarthAgent and Debate occupy the upper-right quadrant, delivering both high performance and high stability. Plan\&Execute is also a strong performer but exhibits slightly less consistency.

\noindent
\textbf{Impact of Rare Tools on Performance.} To understand how architectures handle tasks requiring specialized or domain-specific knowledge, we identified 20 ``rare tools'' which are the 20 least frequently occurring tools in the ground truth of all tasks (Table~\ref{tab:rare_tools}). We then compared all architectures' performance on tasks containing zero, one, or two or more of these rare tools.

\begin{table*}[htb]
\centering
\caption{The 20 tools identified as "rare" based on their low frequency in the ground truth dataset.}
\label{tab:rare_tools}
\scalebox{0.9}{
\begin{tabular}{lc|lc}
\toprule
\textbf{Tool Name} & \textbf{GT Frequency} & \textbf{Tool Name} & \textbf{GT Frequency} \\
\midrule
\texttt{analyze\_farming\_patterns} & 3 &  \texttt{monitor\_airspace\_violations} & 1 \\
\texttt{assess\_air\_quality\_patterns} & 1 & \texttt{monitor\_ice\_coverage} & 6 \\
\texttt{assess\_pasture\_quality} & 1 & \texttt{monitor\_traffic\_congestion} & 7 \\
\texttt{classify\_landscape\_type} & 2 & \texttt{predict\_harvest\_timing} & 4 \\
\texttt{download\_file} & 4 & \texttt{read\_database} & 5 \\
\texttt{evaluate\_fishing\_grounds} & 3 & \texttt{track\_biodiversity\_changes} & 8 \\
\texttt{evaluate\_green\_space\_distribution} & 3 & \texttt{track\_ocean\_currents} & 7 \\
\texttt{evaluate\_ore\_quality} & 8 & \texttt{track\_ship\_movements} & 1 \\
\texttt{extract\_keywords} & 4 & \texttt{track\_wildfire\_progression} & 1 \\
\texttt{monitor\_air\_pollution} & 2 & \texttt{translate\_text} & 1 \\
\bottomrule
\end{tabular}}
\end{table*}

\begin{figure}[t]
    \centering
    \includegraphics[width=0.5\textwidth]{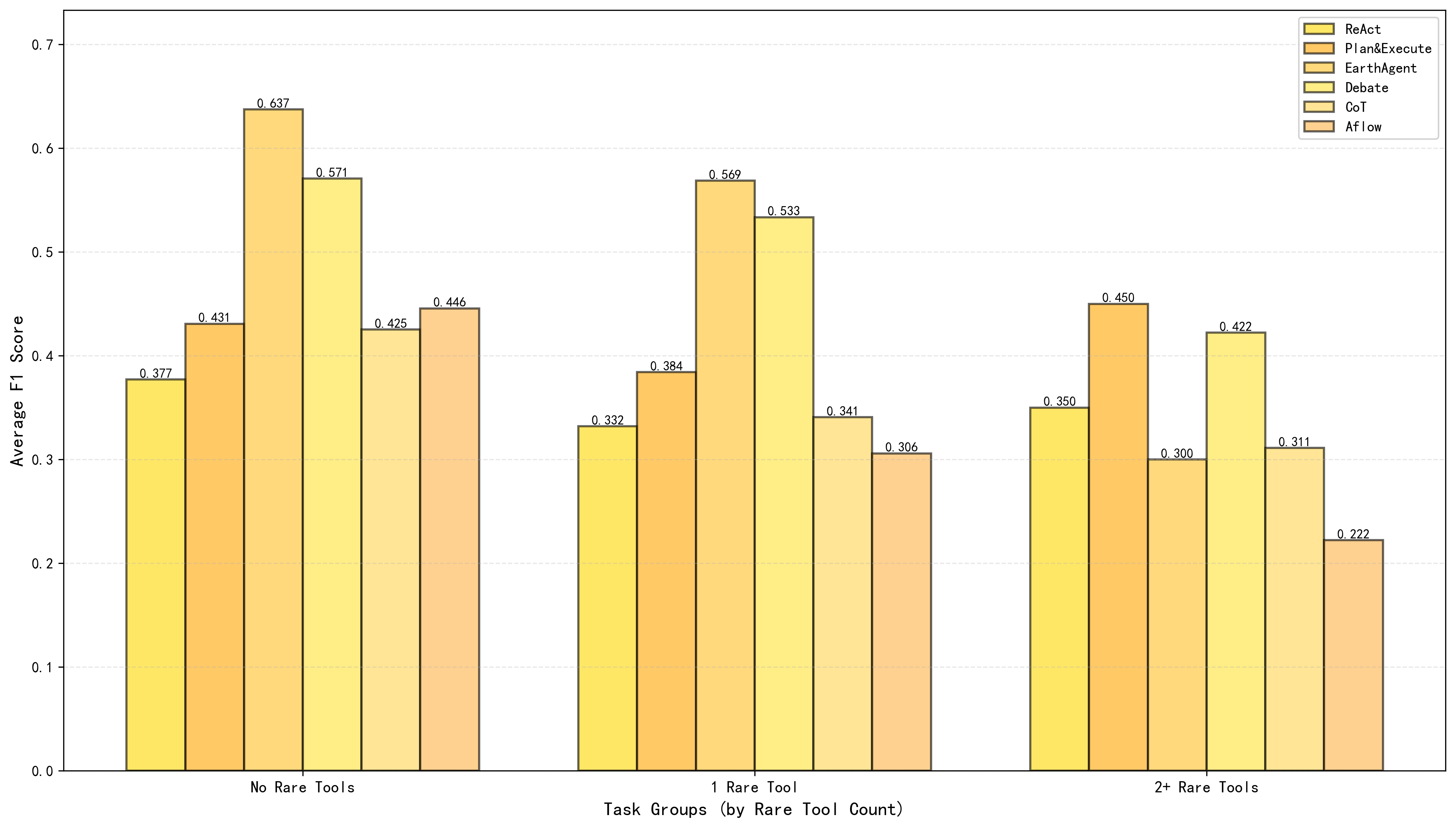}
    \caption{Impact of Rare Tools on Correctness Metric Score.}
    \label{fig:rare_tools_impact}
\end{figure}

The results, shown in Table \ref{tab:rare_performance} and visualized in Figure \ref{fig:rare_tools_impact}, demonstrate a significant performance drop for all architectures when faced with tasks requiring rare tools.

\begin{table}[t]
\centering
\caption{Correctness Metric Score on Tasks Grouped by the Number of Rare Tools in their Ground Truth. Sample size (n) is shown in parentheses.}
\label{tab:rare_performance}
    \resizebox{\linewidth}{!}{
\begin{tabular}{lccr}
\toprule
\textbf{Architecture} & \textbf{No Rare Tools} & \textbf{1 Rare Tool} & \textbf{2+ Rare Tools} \\
\midrule
ReAct & 0.3770 (n=926) & 0.3319 (n=68) & 0.3500 (n=2) \\
Plan\&Execute & 0.4308 (n=926) & 0.3841 (n=68) & \textbf{0.4500} (n=2) \\
EarthAgent & \textbf{0.6373} (n=926) & \textbf{0.5688} (n=68) & 0.3000 (n=2) \\
Debate & \underline{0.5708} (n=926) & \underline{0.5335} (n=68) & \underline{0.4222} (n=2) \\
CoT & 0.4253 (n=926) & 0.3407 (n=68) & 0.3111 (n=2) \\
AFlow & 0.4456 (n=926) & 0.3059 (n=68) & 0.2222 (n=2) \\
\bottomrule
\end{tabular}
}
\end{table}

This analysis strongly suggests that the architectures' ``knowledge'' is concentrated on common tools and workflows. When a task requires specialized, domain-specific tools (the ``long tail''), their ability to generate correct plans decreases significantly. Notably, EarthAgent experiences one of the steepest declines, dropping from 0.68 to 0.57, while AFlow also suffers considerable performance degradation. This underscores the challenge of generalization and the critical importance of domain-specific knowledge in complex task planning. Notice that the \textbf{2+ Rare Tools} category has a very small sample size, so we don't include it into our analysis.

\subsection{Case Study}
\label{sec:case_study}

In this part, we provide some representative example in evaluation process.
\subsubsection{Observation of EarthAgent in Structural Metric}
Here, we present the 5 worst-performing cases of EarthAgent method on structural metric. We find that these tasks basically are simple and have relatively short gournd truth. The low EarthAgent scores stem from its tendency to overcomplicate the questions.

\begin{lstlisting}[
    style=promptstyle, 
    label={tam_sim_case_1}
]
task_id: 28a25feb-131c-4fff-bf1e-0f82c4b849b9
question: Following the severe winter storms, determine the total land area lost from the coastal bluffs at Miramar Beach in Half Moon Bay from the beginning of the year until October 1st, 2023.
domain: Marine & Water Resources
complexity: Simple
ground_truth: ['get_current_time', 'calculate_time_difference', 'assess_coastal_erosion', 'generate_analysis_reports']
---------------------------------------
ReAct: ['download_satellite_imagery', 'geometric_correction', 'classify_land_cover', 'read_database']
Structural Metric: 0.7519181314855814
---------------------------------------
Plan&Execute: ['monitor_coastal_erosion', 'analyze_geological_structures', 'format_data']
Structural Metric: 0.5492723754137505
---------------------------------------
CoT: ['web_search', 'web_search', 'statistical_analysis', 'download_satellite_imagery', 'classify_land_cover', 'format_data']
Structural Metric: 0.6520651606196051
---------------------------------------
Debate: ['download_satellite_imagery', 'cloud_mask_removal', 'atmospheric_correction', 'geometric_correction', 'classify_land_cover', 'assess_coastal_erosion', 'statistical_analysis']   
Structural Metric: 0.5356477422539736
---------------------------------------
AFlow: ['download_satellite_imagery', 'geometric_correction', 'atmospheric_correction', 'cloud_mask_removal', 'classify_land_cover', 'detect_changes', 'format_data', 'generate_analysis_reports']
Structural Metric: 0.5211968036443684
---------------------------------------
EarthAgent: ['recommend_satellite_platforms', 'download_satellite_imagery', 'get_weather_data', 'atmospheric_correction', 'geometric_correction', 'cloud_mask_removal', 'assess_disaster_damage', 'detect_urban_expansion', 'classify_land_cover', 'segment_vegetation', 'classify_landscape_type', 'classify_terrain_type', 'monitor_flood_extent', 'evaluate_infrastructure_damage', 'generate_analysis_reports', 'statistical_analysis']
Structural Metric: 0.3833063380272993
============================================
task_id: 7f9c6c7a-df06-4955-b845-487dafc5079d
question: Assess the concentrations and distribution of heavy metals, microplastics, and nutrients in the San Francisco Bay following the first significant rainfall during the first week of October 2023.
domain: Marine & Water Resources
complexity: Simple
ground_truth: ['get_weather_data', 'assess_marine_pollution', 'format_data', 'generate_analysis_reports']
--------------------------------------- ['web_search', 'analyze_farming_patterns', 'analyze_farming_patterns', 'statistical_analysis', 'web_search', 'statistical_analysis']
Structural Metric: 0.6648747760559397
---------------------------------------
Plan&Execute: ['web_search', 'monitor_water_quality', 'format_data', 'statistical_analysis']
Structural Metric: 0.7290138204892476
---------------------------------------
CoT: ['get_weather_data', 'web_search', 'download_file', 'statistical_analysis', 'format_data']
Structural Metric: 0.7085224305041438
---------------------------------------
Debate: ['get_weather_data', 'web_search', 'read_database', 'statistical_analysis', 'summarize_text']
Structural Metric: 0.7237562433757787
---------------------------------------
AFlow: ['download_satellite_imagery', 'geometric_correction', 'atmospheric_correction', 'cloud_mask_removal', 'segment_water_bodies', 'analyze_pollutants', 'format_data', 'generate_analysis_reports']
Structural Metric: 0.5460570069461403
---------------------------------------
EarthAgent: ['recommend_satellite_platforms', 'download_satellite_imagery', 'get_weather_data', 'atmospheric_correction', 'geometric_correction', 'cloud_mask_removal', 'detect_urban_expansion', 'monitor_deforestation', 'classify_land_cover', 'segment_vegetation', 'classify_landscape_type', 'classify_terrain_type', 'assess_water_quality', 'track_biodiversity_changes', 'evaluate_ecosystem_services', 'monitor_forest_health', 'assess_wetland_conditions']
Structural Metric: 0.3826746628523331
============================================
task_id: 6d35f8c7-c6c9-451a-bfa3-5420aa5a920d
question: Considering the peak growing season in mid-June 2023, which residential neighborhoods in Manhattan exhibit the most significant deficit in public green space accessibility?    
domain: Urban & Regional Planning
complexity: Medium
ground_truth: ['web_search', 'recommend_satellite_platforms', 'download_satellite_imagery', 'extract_image_metadata']
--------------------------------------- ['evaluate_green_space_distribution', 'read_database', 'correlation_analysis']
Structural Metric: 0.6220541343342956
---------------------------------------
Plan&Execute: ['analyze_population_density', 'evaluate_green_space_distribution', 'assess_land_use_efficiency', 'assess_urbanization_level', 'summarize_text']
Structural Metric: 0.6318968263426029
---------------------------------------
CoT: ['read_database', 'evaluate_green_space_distribution', 'statistical_analysis', 'summarize_text']
Structural Metric: 0.7243667108317215
---------------------------------------
Debate: ['read_database', 'evaluate_green_space_distribution', 'analyze_population_density', 'assess_land_use_efficiency', 'correlation_analysis', 'summarize_text']
Structural Metric: 0.5707047106620089
---------------------------------------
AFlow: ['download_satellite_imagery', 'geometric_correction', 'atmospheric_correction', 'cloud_mask_removal', 'classify_land_cover', 'assess_green_space_accessibility', 'generate_analysis_reports', 'format_data']
Structural Metric: 0.5080498688191866
---------------------------------------
EarthAgent: ['recommend_satellite_platforms', 'download_satellite_imagery', 'get_weather_data', 'atmospheric_correction', 'geometric_correction', 'statistical_analysis', 'classify_landscape_type', 'assess_urbanization_level', 'classify_land_cover', 'segment_vegetation', 'detect_urban_expansion', 'monitor_deforestation', 'evaluate_green_space_distribution', 'analyze_population_density', 'assess_land_use_efficiency', 'assess_air_quality_patterns', 'correlation_analysis']
Structural Metric: 0.3813096139548148
============================================
task_id: d95da39a-9e37-428c-8ce2-0d7f636c9d4d
question: To what extent has Central Texas rebuilt its infrastructure and recovered its agricultural land following the severe flooding event in mid-September 2023?
domain: Disaster Emergency & Management
complexity: Medium
ground_truth: ['assess_flood_risk_zones', 'monitor_flood_extent', 'assess_disaster_damage', 'assess_recovery_progress']
--------------------------------------- ['monitor_flood_extent', 'assess_recovery_progress', 'monitor_crop_health', 'evaluate_infrastructure_damage', 'assess_environmental_impact', 'analyze_threat_patterns', 'assess_slope_stability', 'assess_urbanization_level']
Structural Metric: 0.5677680325256582
---------------------------------------
Plan&Execute: ['monitor_flood_extent', 'assess_infrastructure_damage', 'assess_recovery_progress', 'monitor_crop_health', 'assess_soil_moisture', 'predict_crop_yield']
Structural Metric: 0.629565405348949
---------------------------------------
CoT: ['web_search', 'web_search', 'web_search', 'web_search', 'assess_infrastructure_damage', 'monitor_flood_extent', 'assess_recovery_progress', 'summarize_text']
Structural Metric: 0.7181785252865864
---------------------------------------
Debate: ['monitor_flood_extent', 'evaluate_infrastructure_damage', 'assess_recovery_progress', 'monitor_crop_health', 'analyze_farming_patterns', 'classify_land_cover', 'statistical_analysis']
Structural Metric: 0.5856316607624632
---------------------------------------
AFlow: ['download_satellite_imagery', 'geometric_correction', 'atmospheric_correction', 'cloud_mask_removal', 'classify_land_cover', 'assess_disaster_damage', 'format_data', 'generate_analysis_reports']
Structural Metric: 0.4995047243114492
---------------------------------------
EarthAgent: ['recommend_satellite_platforms', 'download_satellite_imagery', 'get_weather_data', 'atmospheric_correction', 'geometric_correction', 'cloud_mask_removal', 'assess_disaster_damage', 'monitor_deforestation', 'classify_land_cover', 'segment_vegetation', 'evaluate_infrastructure_damage', 'assess_recovery_progress', 'statistical_analysis', 'monitor_drought_conditions', 'generate_analysis_reports']
Structural Metric: 0.37995968634487254
============================================
task_id: 12f3964d-c0a8-46fe-9070-8d6a48fb49a3
question: Considering the current environmental risks and population distribution in Mountain View, what is the safest vehicle evacuation route from the vulnerable residential area at 45.3839, -122.4836 to the designated emergency shelter at 45.3941, -122.4930?
domain: Disaster Emergency & Management
complexity: Simple
ground_truth: ['analyze_population_density', 'predict_landslide_risk', 'plan_evacuation_routes', 'generate_analysis_reports']
--------------------------------------- ['assess_flood_risk_zones', 'analyze_population_density', 'plan_evacuation_routes', 'monitor_traffic_congestion']
Structural Metric: 0.8181467528144518
---------------------------------------
Plan&Execute: ['evaluate_flood_risk_zones', 'monitor_traffic_congestion', 'plan_evacuation_routes']
Structural Metric: 0.5690546123325106
---------------------------------------
CoT: ['assess_environmental_impact', 'analyze_population_density', 'plan_evacuation_routes', 'get_weather_data', 'format_data']
Structural Metric: 0.7148938120542967
---------------------------------------
Debate: ['detect_roads', 'assess_flood_risk_zones', 'track_wildfire_progression', 'monitor_traffic_congestion', 'assess_road_conditions', 'predict_landslide_risk', 'assess_earthquake_damage', 'plan_evacuation_routes']
Structural Metric: 0.5495748485005734
---------------------------------------
AFlow: ['download_satellite_imagery', 'geometric_correction', 'atmospheric_correction', 'cloud_mask_removal', 'assess_disaster_risks', 'identify_evacuations_routes', 'generate_analysis_reports']
Structural Metric: 0.5696155175193354
---------------------------------------
EarthAgent: ['recommend_satellite_platforms', 'download_satellite_imagery', 'get_weather_data', 'atmospheric_correction', 'cloud_mask_removal', 'geometric_correction', 'monitor_deforestation', 'detect_urban_expansion', 'classify_landscape_type', 'assess_urbanization_level', 'detect_roads', 'detect_buildings', 'monitor_flood_extent', 'track_wildfire_progression', 'statistical_analysis', 'evaluate_infrastructure_damage']
Structural Metric: 0.37793323951852476

\end{lstlisting}

Meanwhile, there are tasks where EarthAgent are good at. We show 5 of them.

\begin{lstlisting}[
    style=promptstyle, 
    caption={},
    label={tam_sim_good_case_1}
]
task_id: fd5cd797-e422-41f4-a5a9-fa205472416b
question: Quantify the statistically significant trend of coastal erosion along the shoreline between latitudes 34N and 35N for the year 2023.
domain: Marine & Water Resources
complexity: Medium
ground_truth: ['recommend_satellite_platforms', 'download_satellite_imagery', 'geometric_correction', 'atmospheric_correction', 'cloud_mask_removal', 'crop_image', 'segment_water_bodies', 'calculate_time_difference', 'assess_coastal_erosion', 'statistical_analysis', 'format_data', 'generate_analysis_reports']
---------------------------------------
ReAct: ['download_satellite_imagery', 'statistical_analysis', 'statistical_analysis', 'generate_analysis_reports']
Structural Metric: 0.3803773766006371
---------------------------------------
Plan&Execute: ['download_satellite_imagery', 'atmospheric_correction', 'geometric_correction', 'detect_military_facilities', 'segment_coastal_erosion', 'statistical_analysis']
Structural Metric: 0.5234263483593165
---------------------------------------
CoT: ['web_search', 'web_search', 'I will download the dataset for analysis', 'format_data', 'statistical_analysis', 'linear_regression', 'summarize_text']
Structural Metric: 0.4564959376866067
---------------------------------------
Debate: ['download_satellite_imagery', 'atmospheric_correction', 'geometric_correction', 'cloud_mask_removal', 'classify_land_cover', 'assess_coastal_erosion', 'statistical_analysis', 'temporal_analysis']
Structural Metric: 0.6554164563067398
---------------------------------------
AFlow: ['download_satellite_imagery', 'geometric_correction', 'atmospheric_correction', 'cloud_mask_removal', 'segment_water_bodies', 'classify_land_cover', 'statistical_analysis', 'generate_analysis_reports']
Structural Metric: 0.7207758695382465
---------------------------------------
EarthAgent: ['recommend_satellite_platforms', 'download_satellite_imagery', 'get_weather_data', 'atmospheric_correction', 'cloud_mask_removal', 'geometric_correction', 'classify_landscape_type', 'classify_terrain_type', 'assess_coastal_erosion', 'statistical_analysis', 'linear_regression', 'generate_analysis_reports']
Structural Metric: 0.8962936784244246
\end{lstlisting}
\begin{lstlisting}[
    style=promptstyle, 
    caption={},
    label={tam_sim_good_case_2}
]
task_id: 14c50a12-f2cf-4181-b44f-f0935dd79ee3
question: Assess the health of corn, soybean, and wheat crops in the agricultural plots just east of downtown Los Angeles during January 2023, pinpointing any areas showing high-confidence indicators of specific plant diseases.
domain: Agriculture & Forestry
complexity: Complex
ground_truth: ['recommend_satellite_platforms', 'download_satellite_imagery', 'geometric_correction', 'atmospheric_correction', 'cloud_mask_removal', 'crop_image', 'enhance_image_resolution', 'segment_vegetation', 'classify_land_cover', 'monitor_crop_health', 'detect_plant_diseases']
---------------------------------------
ReAct: ['download_satellite_imagery', 'atmospheric_correction', 'detect_plant_diseases']
Structural Metric: 0.3809054440290979
---------------------------------------
Plan&Execute: ['monitor_crop_health', 'detect_plant_diseases']
Structural Metric: 0.2982246972559247
---------------------------------------
CoT: ['download_satellite_imagery', 'atmospheric_correction', 'geometric_correction', 'segment_agricultural_fields', 'monitor_crop_health', 'detect_plant_diseases', 'summarize_text']
Structural Metric: 0.6319789365923533
---------------------------------------
Debate: ['download_satellite_imagery', 'atmospheric_correction', 'geometric_correction', 'assess_soil_moisture', 'get_weather_data', 'segment_agricultural_fields', 'classify_land_cover', 'monitor_crop_health', 'detect_plant_diseases', 'summarize_text']
Structural Metric: 0.781518882364096
---------------------------------------
AFlow: ['download_satellite_imagery', 'geometric_correction', 'atmospheric_correction', 'cloud_mask_removal', 'monitor_crop_health', 'detect_plant_diseases', 'format_data', 'generate_analysis_reports']
Structural Metric: 0.7246626350376182
---------------------------------------
EarthAgent: ['recommend_satellite_platforms', 'download_satellite_imagery', 'suggest_processing_workflows', 'atmospheric_correction', 'geometric_correction', 'cloud_mask_removal', 'classify_land_cover', 'segment_vegetation', 'monitor_crop_health', 'detect_plant_diseases', 'monitor_crop_health', 'monitor_crop_health', 'detect_plant_diseases']
Structural Metric: 0.8638882881783954
\end{lstlisting}
\begin{lstlisting}[
    style=promptstyle, 
    caption={},
    label={tam_sim_good_case_3}
]
task_id: b591986a-38c5-4a1f-a775-2b38a48efa44
question: For the agricultural fields within the Los Angeles basin, quantify the relationship between individual field size, peak vegetation health during the summer 2023 growing season, and the resulting predicted crop yield.
domain: Agriculture & Forestry
complexity: Medium
ground_truth: ['web_search', 'recommend_satellite_platforms', 'download_satellite_imagery', 'cloud_mask_removal', 'geometric_correction', 'atmospheric_correction', 'crop_image', 'segment_vegetation', 'segment_agricultural_fields', 'statistical_analysis', 'predict_crop_yield', 'correlation_analysis', 'linear_regression']
---------------------------------------
ReAct: ['read_database', 'statistical_analysis', 'statistical_analysis', 'correlation_analysis', 'correlation_analysis', 'read_database']
Structural Metric: 0.3971436873952484
---------------------------------------
Plan&Execute: ['monitor_crop_health', 'predict_crop_yield', 'analyze_farming_patterns', 'correlation_analysis']
Structural Metric: 0.4188257513782573
---------------------------------------
CoT: ['read_database', 'statistical_analysis', 'correlation_analysis', 'linear_regression', 'summarize_text']
Structural Metric: 0.4319192084485427
---------------------------------------
Debate: ['download_satellite_imagery', 'atmospheric_correction', 'cloud_mask_removal', 'segment_agricultural_fields', 'monitor_crop_health', 'format_data', 'statistical_analysis', 'correlation_analysis', 'linear_regression', 'predict_crop_yield']
Structural Metric: 0.6972034412143604
---------------------------------------
AFlow: ['download_satellite_imagery', 'geometric_correction', 'atmospheric_correction', 'cloud_mask_removal', 'monitor_crop_health', 'statistical_analysis', 'generate_analysis_reports', 'format_data']
Structural Metric: 0.633400521092164
---------------------------------------
EarthAgent: ['recommend_satellite_platforms', 'download_satellite_imagery', 'get_weather_data', 'atmospheric_correction', 'geometric_correction', 'cloud_mask_removal', 'classify_land_cover', 'segment_vegetation', 'monitor_crop_health', 'predict_crop_yield', 'statistical_analysis', 'correlation_analysis', 'linear_regression']
Structural Metric: 0.8613930069483243
\end{lstlisting}
\begin{lstlisting}[
    style=promptstyle, 
    caption={},
    label={tam_sim_good_case_4}
]
task_id: 5757265e-e1b5-4cdb-bb41-a2f7c085d464
question: Following the recent seismic event in Humboldt County, California, produce a detailed building-level damage assessment and an evaluation of initial recovery activities observed during early August 2023.
domain: Disaster Emergency & Management
complexity: Medium
ground_truth: ['web_search', 'download_satellite_imagery', 'extract_image_metadata', 'atmospheric_correction', 'cloud_mask_removal', 'geometric_correction', 'crop_image', 'enhance_image_resolution', 'segment_individual_buildings', 'assess_earthquake_damage', 'assess_disaster_damage', 'assess_recovery_progress', 'generate_analysis_reports', 'format_data']
---------------------------------------
ReAct: ['assess_disaster_damage', 'assess_recovery_progress', 'analyze_population_density', 'assess_environmental_impact', 'assess_environmental_impact', 'monitor_air_pollution', 'assess_environmental_impact', 'assess_environmental_impact', 'assess_environmental_impact', 'assess_environmental_impact']
Structural Metric: 0.463475499035262
---------------------------------------
Plan&Execute: ['assess_earthquake_damage', 'assess_recovery_progress']
Structural Metric: 0.30108128142281765
---------------------------------------
CoT: ['web_search', 'web_search', 'assess_earthquake_damage', 'assess_recovery_progress', 'format_data']
Structural Metric: 0.4111871575296303
---------------------------------------
Debate: ['download_satellite_imagery', 'cloud_mask_removal', 'geometric_correction', 'atmospheric_correction', 'detect_individual_buildings', 'assess_earthquake_damage', 'assess_infrastructure_damage', 'assess_recovery_progress', 'monitor_traffic_congestion', 'analyze_urban_growth_patterns', 'classify_land_cover']
Structural Metric: 0.6371590403326893
---------------------------------------
AFlow: ['download_satellite_imagery', 'geometric_correction', 'atmospheric_correction', 'cloud_mask_removal', 'assess_disaster_damage', 'monitor_crop_health', 'generate_analysis_reports']
Structural Metric: 0.5766079862654101
---------------------------------------
EarthAgent: ['recommend_satellite_platforms', 'download_satellite_imagery', 'get_weather_data', 'atmospheric_correction', 'cloud_mask_removal', 'geometric_correction', 'detect_buildings', 'extract_image_metadata', 'segment_individual_buildings', 'segment_individual_buildings', 'assess_disaster_damage', 'monitor_deforestation', 'assess_earthquake_damage', 'assess_recovery_progress', 'statistical_analysis']
Structural Metric: 0.859065704757259
\end{lstlisting}
\begin{lstlisting}[
    style=promptstyle, 
    caption={},
    label={tam_sim_good_case_5}
]
task_id: e094fa9a-2e60-4bc4-aaed-ab15a70754c5
question: Quantify the average shoreline retreat along the urbanized coast of Mykonos, Greece, by comparing the waterfront position of January 2022 to that of January 2021.
domain: Marine & Water Resources
complexity: Simple
ground_truth: ['download_satellite_imagery', 'geometric_correction', 'atmospheric_correction', 'cloud_mask_removal', 'crop_image', 'segment_water_bodies', 'calculate_time_difference', 'assess_coastal_erosion', 'statistical_analysis', 'generate_analysis_reports']
---------------------------------------
ReAct: ['download_satellite_imagery', 'download_satellite_imagery', 'detect_ships', 'detect_ships', 'detect_ships', 'detect_ships', 'detect_buildings', 'detect_buildings', 'detect_buildings', 'detect_ships']
Structural Metric: 0.4431240695754549
---------------------------------------
Plan&Execute: ['download_satellite_imagery', 'download_satellite_imagery', 'geometric_correction', 'geometric_correction', 'extract_image_metadata', 'extract_image_metadata', 'detect_roads', 'detect_roads', 'correlation_analysis', 'format_data']
Structural Metric: 0.5899371128885651
---------------------------------------
CoT: ['download_satellite_imagery', 'atmospheric_correction', 'geometric_correction', 'segment_water_bodies', 'statistical_analysis', 'statistical_analysis', 'format_data']
Structural Metric: 0.6189795057360581
---------------------------------------
Debate: ['download_satellite_imagery', 'geometric_correction', 'atmospheric_correction', 'cloud_mask_removal', 'classify_land_cover', 'segment_water_bodies', 'statistical_analysis']
Structural Metric: 0.7455656458466684
---------------------------------------
AFlow: ['download_satellite_imagery', 'geometric_correction', 'atmospheric_correction', 'cloud_mask_removal', 'classify_land_cover', 'statistical_analysis', 'generate_analysis_reports']
Structural Metric: 0.7413091656675803
---------------------------------------
EarthAgent: ['recommend_satellite_platforms', 'download_satellite_imagery', 'download_satellite_imagery', 'atmospheric_correction', 'cloud_mask_removal', 'geometric_correction', 'classify_landscape_type', 'assess_urbanization_level', 'analyze_urban_growth_patterns', 'statistical_analysis', 'correlation_analysis']
Structural Metric: 0.8558551256855329

\end{lstlisting}

\subsubsection{Cases Debate Do Well in}
we present the 5 best-performing cases of Debate method on structural metric. Debate can leverage several LLM-driven debator agent to refine the answer over and over again. We find it does well on tasks that have medium length of ground truth and need little more reasoning and comprehensive skills.
\begin{lstlisting}[
    style=promptstyle, 
    caption={},
    label={debate_sim_case_1}
]
task_id: 7f86fe57-67e6-4f1d-979b-b5057ef3883e
question: Quantify the total carbon sequestration loss in metric tons from forest reduction within the area bounded by 45.0N, 44.5N, -73.0E, and -73.5W for the three-year period beginning January 1, 2020.
domain: Agriculture & Forestry
complexity: Simple
ground_truth: ['download_satellite_imagery', 'geometric_correction', 'atmospheric_correction', 'cloud_mask_removal', 'crop_image', 'classify_land_cover', 'monitor_deforestation', 'assess_carbon_sequestration', 'generate_analysis_reports']
---------------------------------------
ReAct: ['download_satellite_imagery', 'classify_land_cover', 'web_search', 'statistical_analysis']
Structural Metric: 0.4826636062742867
---------------------------------------
Plan&Execute: ['monitor_forest_health', 'assess_carbon_sequestration']
Structural Metric: 0.31060735321796074
---------------------------------------
CoT: ['download_satellite_imagery', 'classify_land_cover', 'monitor_deforestation', 'assess_carbon_sequestration', 'format_data']
Structural Metric: 0.5711478990381471
---------------------------------------
Debate: ['download_satellite_imagery', 'atmospheric_correction', 'geometric_correction', 'cloud_mask_removal', 'band_combination', 'segment_vegetation', 'monitor_deforestation', 'assess_carbon_sequestration', 'statistical_analysis']
Structural Metric: 0.9254394035648417
---------------------------------------
AFlow: ['download_satellite_imagery', 'geometric_correction', 'atmospheric_correction', 'cloud_mask_removal', 'classify_land_cover', 'format_data', 'generate_analysis_reports']
Structural Metric: 0.7841913067497179
---------------------------------------
EarthAgent: ['recommend_satellite_platforms', 'atmospheric_correction', 'geometric_correction', 'cloud_mask_removal', 'monitor_deforestation', 'assess_disaster_damage', 'classify_land_cover', 'segment_vegetation', 'classify_landscape_type', 'assess_urbanization_level', 'monitor_forest_health', 'assess_carbon_sequestration', 'track_biodiversity_changes', 'detect_illegal_logging', 'evaluate_ecosystem_services']
Structural Metric: 0.6513509848247034
\end{lstlisting}
\begin{lstlisting}[
    style=promptstyle, 
    caption={},
    label={debate_sim_case_2}
]
task_id: 385a42c1-6292-4a09-93b3-22bbec4a9ec7
question: For the Presidio Trust's 2023 land use assessment, provide a detailed architectural footprint analysis of structures between the Presidio Golf Course and eastern Golden Gate Park, including a statistical breakdown of building counts and average sizes categorized by type.
domain: Urban & Regional Planning
complexity: Simple
ground_truth: ['download_satellite_imagery', 'cloud_mask_removal', 'geometric_correction', 'crop_image', 'atmospheric_correction', 'detect_buildings', 'segment_individual_buildings', 'statistical_analysis', 'generate_analysis_reports']
---------------------------------------
ReAct: ['download_satellite_imagery', 'detect_buildings', 'statistical_analysis']
Structural Metric: 0.4258193318765149
---------------------------------------
Plan&Execute: ['download_satellite_imagery', 'geometric_correction', 'cloud_mask_removal', 'detect_buildings', 'classify_land_cover', 'extract_image_metadata', 'statistical_analysis']
Structural Metric: 0.7191934120716232
---------------------------------------
CoT: ['download_satellite_imagery', 'geometric_correction', 'detect_buildings', 'segment_individual_buildings', 'format_data', 'statistical_analysis', 'format_data']
Structural Metric: 0.701682978215084
---------------------------------------
Debate: ['download_satellite_imagery', 'atmospheric_correction', 'geometric_correction', 'cloud_mask_removal', 'detect_buildings', 'segment_urban_areas', 'segment_individual_buildings', 'statistical_analysis', 'format_data']
Structural Metric: 0.8970724741617838
---------------------------------------
AFlow: ['download_satellite_imagery', 'geometric_correction', 'atmospheric_correction', 'cloud_mask_removal', 'classify_land_cover', 'statistical_analysis', 'generate_analysis_reports'] 
Structural Metric: 0.759859128691575
---------------------------------------
EarthAgent: ['recommend_satellite_platforms', 'download_satellite_imagery', 'get_weather_data', 'atmospheric_correction', 'geometric_correction', 'statistical_analysis', 'detect_buildings', 'detect_roads', 'segment_individual_buildings', 'segment_agricultural_fields', 'classify_land_cover', 'segment_vegetation', 'assess_land_use_efficiency', 'statistical_analysis', 'evaluate_green_space_distribution', 'analyze_population_density', 'analyze_population_density']
Structural Metric: 0.6034446646374529
\end{lstlisting}
\begin{lstlisting}[
    style=promptstyle, 
    caption={},
    label={debate_sim_case_3}
]
task_id: 42a99dbb-7d54-44fe-bd88-54489937b8a9
question: Assess the 2022 carbon sequestration capacity of the varied ecosystems within the California region stretching from the Central Valley floor to the high elevations of the southern Sierra Nevada.
domain: Agriculture & Forestry
complexity: Simple
ground_truth: ['download_satellite_imagery', 'geometric_correction', 'atmospheric_correction', 'cloud_mask_removal', 'crop_image', 'segment_vegetation', 'classify_land_cover', 'assess_carbon_sequestration']
---------------------------------------
ReAct: ['assess_carbon_sequestration', 'assess_carbon_sequestration', 'analyze_farming_patterns', 'assess_land_use_efficiency', 'monitor_forest_health', 'evaluate_ecosystem_services', 'assess_carbon_sequestration', 'assess_soil_moisture', 'analyze_farming_patterns', 'track_biodiversity_changes']
Structural Metric: 0.693135573687668
---------------------------------------
Plan&Execute: ['assess_carbon_sequestration', 'summarize_text']
Structural Metric: 0.3044321291012029
---------------------------------------
CoT: ['web_search', 'web_search', 'summarize_text', 'statistical_analysis', 'correlation_analysis', 'format_data']
Structural Metric: 0.5203244053992784
---------------------------------------
Debate: ['download_satellite_imagery', 'atmospheric_correction', 'geometric_correction', 'cloud_mask_removal', 'classify_land_cover', 'segment_vegetation', 'assess_carbon_sequestration', 'statistical_analysis']
Structural Metric: 0.8968025321761767
---------------------------------------
AFlow: ['download_satellite_imagery', 'geometric_correction', 'atmospheric_correction', 'cloud_mask_removal', 'classify_land_cover', 'statistical_analysis', 'generate_analysis_reports'] 
Structural Metric: 0.8224986597176879
---------------------------------------
EarthAgent: ['recommend_satellite_platforms', 'download_satellite_imagery', 'get_weather_data', 'atmospheric_correction', 'geometric_correction', 'cloud_mask_removal', 'classify_landscape_type', 'classify_terrain_type', 'monitor_deforestation', 'detect_urban_expansion', 'classify_land_cover', 'segment_vegetation', 'assess_carbon_sequestration', 'evaluate_ecosystem_services', 'track_biodiversity_changes', 'monitor_forest_health', 'assess_water_quality']
Structural Metric: 0.5801916555758977
\end{lstlisting}
\begin{lstlisting}[
    style=promptstyle, 
    caption={},
    label={debate_sim_case_4}
]
task_id: 75f963f1-0746-42ca-bae3-e1a971efb4cd
question: Pinpoint the hotspots and quantify the spatial intensity of urban expansion in Lower Manhattan from 2010 to 2020.
domain: Urban & Regional Planning
complexity: Simple
ground_truth: ['download_satellite_imagery', 'geometric_correction', 'atmospheric_correction', 'cloud_mask_removal', 'crop_image', 'classify_land_cover', 'detect_urban_expansion', 'analyze_urban_growth_patterns', 'statistical_analysis', 'generate_analysis_reports']
---------------------------------------
ReAct: ['download_satellite_imagery', 'download_satellite_imagery', 'classify_land_cover', 'classify_land_cover', 'correlation_analysis']
Structural Metric: 0.3990976384140391
---------------------------------------
Plan&Execute: ['download_satellite_imagery', 'atmospheric_correction', 'geometric_correction', 'classify_land_cover', 'detect_urban_expansion', 'analyze_urban_growth_patterns', 'assess_land_use_efficiency']
Structural Metric: 0.7012794784075258
---------------------------------------
CoT: ['download_satellite_imagery', 'atmospheric_correction', 'geometric_correction', 'segment_urban_areas', 'detect_urban_expansion', 'assess_urbanization_level', 'analyze_urban_growth_patterns']
Structural Metric: 0.6824979742013693
---------------------------------------
Debate: ['download_satellite_imagery', 'atmospheric_correction', 'geometric_correction', 'cloud_mask_removal', 'classify_land_cover', 'segment_urban_areas', 'detect_urban_expansion', 'analyze_urban_growth_patterns', 'statistical_analysis', 'temporal_analysis']
Structural Metric: 0.8885315597057343
---------------------------------------
AFlow: ['download_satellite_imagery', 'geometric_correction', 'atmospheric_correction', 'cloud_mask_removal', 'classify_land_cover', 'quantify_changes', 'format_data', 'generate_analysis_reports']
Structural Metric: 0.7426862084609933
---------------------------------------
EarthAgent: ['recommend_satellite_platforms', 'download_satellite_imagery', 'download_satellite_imagery', 'atmospheric_correction', 'geometric_correction', 'statistical_analysis', 'detect_urban_expansion', 'monitor_deforestation', 'classify_land_cover', 'segment_urban_areas', 'analyze_urban_growth_patterns', 'statistical_analysis', 'assess_land_use_efficiency', 'evaluate_green_space_distribution', 'assess_air_quality_patterns']
Structural Metric: 0.7202834741050391
\end{lstlisting}
\begin{lstlisting}[
    style=promptstyle, 
    caption={},
    label={debate_sim_case_5}
]
task_id: 5ab85e87-cb9d-4e6f-be92-28adfb44acb4
question: To what extent did the environmental repercussions of vegetation loss during 2023, particularly on carbon emissions and biodiversity, vary between different districts of Los Angeles?
domain: Environmental Monitoring & Climate Change
complexity: Simple
ground_truth: ['download_satellite_imagery', 'geometric_correction', 'atmospheric_correction', 'cloud_mask_removal', 'classify_land_cover', 'monitor_deforestation', 'assess_environmental_impact', 'statistical_analysis', 'generate_analysis_reports']
---------------------------------------
ReAct: ['read_database', 'statistical_analysis', 'generate_analysis_reports']
Structural Metric: 0.42368791612803935
---------------------------------------
Plan&Execute: ['monitor_forest_health', 'assess_carbon_sequestration', 'track_biodiversity_changes', 'analyze_population_density', 'assess_environmental_impact', 'evaluate_ecosystem_services', 'summarize_text']
Structural Metric: 0.6476911148752748
---------------------------------------
CoT: ['web_search', 'web_search', 'read_database', 'statistical_analysis', 'correlation_analysis', 'summarize_text']
Structural Metric: 0.509136391560046
---------------------------------------
Debate: ['download_satellite_imagery', 'atmospheric_correction', 'geometric_correction', 'cloud_mask_removal', 'segment_vegetation', 'classify_land_cover', 'assess_carbon_sequestration', 'track_biodiversity_changes', 'correlation_analysis']
Structural Metric: 0.8743224684838895
---------------------------------------
AFlow: ['download_satellite_imagery', 'geometric_correction', 'atmospheric_correction', 'cloud_mask_removal', 'crop_image', 'classify_land_cover', 'assess_disaster_damage', 'format_data', 'generate_analysis_reports']
Structural Metric: 0.9129544281297259
---------------------------------------
EarthAgent: ['recommend_satellite_platforms', 'download_satellite_imagery', 'get_weather_data', 'atmospheric_correction', 'geometric_correction', 'statistical_analysis', 'monitor_deforestation', 'detect_urban_expansion', 'classify_land_cover', 'segment_vegetation', 'assess_carbon_sequestration', 'track_biodiversity_changes', 'evaluate_ecosystem_services', 'assess_water_quality', 'monitor_air_pollution']
Structural Metric: 0.6284940975578261

\end{lstlisting}
\subsubsection{Cases Where AFlow Behave Brilliant}
we present the 5 best-performing cases of AFlow method on structural metric. AFlow is a powerful workflow trained by 248 tasks in our GeoPlan-bench. Partly because of this, it achieves great results in our evaluation. One thing that is worth noticing is that in the cases below, AFlow gets much higher scores than other architectures.


\begin{lstlisting}[
    style=promptstyle, 
    caption={},
    label={aflow_sim_case_1}
]
task_id: 7e112f37-780a-44dc-9ce8-dd6c5051dc2e
question: Assess the early summer ecological health of California's major reservoirs by characterizing the prevalence of algal content and sediment levels throughout June 2023.
domain: Marine & Water Resources
complexity: Simple
ground_truth: ['download_satellite_imagery', 'geometric_correction', 'atmospheric_correction', 'cloud_mask_removal', 'segment_water_bodies', 'assess_water_quality', 'statistical_analysis', 'generate_analysis_reports']
---------------------------------------
ReAct: ['web_search', 'summarize_text']
Structural Metric: 0.31720570644298307
---------------------------------------
Plan&Execute: ['monitor_algae_blooms', 'assess_water_quality']
Structural Metric: 0.3478217087703318
---------------------------------------
CoT: ['web_search', 'web_search', 'download_file', 'format_data', 'statistical_analysis', 'summarize_text']
Structural Metric: 0.556981332423482
---------------------------------------
Debate: ['web_search', 'download_satellite_imagery', 'monitor_algae_blooms', 'assess_water_quality', 'read_database', 'statistical_analysis', 'correlation_analysis', 'summarize_text']   
Structural Metric: 0.7603195278594891
---------------------------------------
AFlow: ['download_satellite_imagery', 'geometric_correction', 'atmospheric_correction', 'cloud_mask_removal', 'segment_water_bodies', 'classify_land_cover', 'statistical_analysis', 'generate_analysis_reports']
Structural Metric: 0.9701370671391487
---------------------------------------
EarthAgent: ['recommend_satellite_platforms', 'download_satellite_imagery', 'get_weather_data', 'atmospheric_correction', 'geometric_correction', 'cloud_mask_removal', 'monitor_deforestation', 'detect_urban_expansion', 'classify_land_cover', 'segment_vegetation', 'assess_water_quality', 'assess_wetland_conditions', 'track_biodiversity_changes', 'evaluate_ecosystem_services', 'assess_carbon_sequestration']
Structural Metric: 0.5940126967061672
\end{lstlisting}
\begin{lstlisting}[
    style=promptstyle, 
    caption={},
    label={aflow_sim_case_2}
]
task_id: cb00ef56-5528-41a8-ab2c-11ff567dc835
question: Based on a full analysis of 2023, determine the level of urban dominance within a 10km radius of downtown Los Angeles.
domain: Urban & Regional Planning
complexity: Simple
ground_truth: ['download_satellite_imagery', 'geometric_correction', 'atmospheric_correction', 'cloud_mask_removal', 'classify_land_cover', 'assess_urbanization_level', 'generate_analysis_reports']
---------------------------------------
ReAct: ['analyze_urban_growth_patterns', 'assess_land_use_efficiency', 'analyze_population_density']
Structural Metric: 0.4351399434383397
---------------------------------------
Plan&Execute: ['segment_urban_areas', 'classify_land_cover', 'assess_urbanization_level']
Structural Metric: 0.506461112466987
---------------------------------------
CoT: ['download_satellite_imagery', 'atmospheric_correction', 'geometric_correction', 'segment_urban_areas', 'assess_urbanization_level', 'analyze_urban_growth_patterns', 'format_data'] 
Structural Metric: 0.8660623771803719
---------------------------------------
Debate: ['download_satellite_imagery', 'classify_land_cover', 'segment_urban_areas', 'analyze_population_density', 'evaluate_green_space_distribution', 'assess_land_use_efficiency', 'analyze_urban_growth_patterns', 'detect_urban_expansion', 'assess_air_quality_patterns', 'evaluate_infrastructure_damage']
Structural Metric: 0.6761787158402515
---------------------------------------
AFlow: ['download_satellite_imagery', 'geometric_correction', 'atmospheric_correction', 'cloud_mask_removal', 'classify_land_cover', 'statistical_analysis', 'generate_analysis_reports'] 
Structural Metric: 0.968482202007657
---------------------------------------
EarthAgent: ['recommend_satellite_platforms', 'download_satellite_imagery', 'get_weather_data', 'atmospheric_correction', 'cloud_mask_removal', 'geometric_correction', 'classify_land_cover', 'segment_urban_areas', 'assess_urbanization_level', 'classify_landscape_type', 'detect_urban_expansion', 'monitor_deforestation', 'assess_land_use_efficiency', 'analyze_population_density', 'analyze_urban_growth_patterns', 'evaluate_green_space_distribution', 'assess_air_quality_patterns']
Structural Metric: 0.5394293566352928
\end{lstlisting}
\begin{lstlisting}[
    style=promptstyle, 
    caption={},
    label={aflow_sim_case_3}
]
task_id: 5fbfd4c8-8d88-4b02-8f8f-a66e7d6e754e
question: To inform a new urban heat island mitigation strategy, determine the proportional land cover distribution of impervious surfaces, vegetation, and water bodies within the city's jurisdiction for January 2023.
domain: Agriculture & Forestry
complexity: Simple
ground_truth: ['download_satellite_imagery', 'geometric_correction', 'atmospheric_correction', 'cloud_mask_removal', 'classify_land_cover', 'statistical_analysis', 'generate_analysis_reports']
---------------------------------------
ReAct: ['download_satellite_imagery', 'classify_land_cover']
Structural Metric: 0.3996067886234683
---------------------------------------
Plan&Execute: ['download_satellite_imagery', 'perform_land_cover_classification', 'calculate_land_cover_distribution', 'format_data']
Structural Metric: 0.42131372720945826
---------------------------------------
CoT: ['download_satellite_imagery', 'atmospheric_correction', 'geometric_correction', 'classify_land_cover', 'segment_urban_areas', 'segment_vegetation', 'segment_water_bodies', 'statistical_analysis', 'format_data']
Structural Metric: 0.7455285880239504
---------------------------------------
Debate: ['download_satellite_imagery', 'atmospheric_correction', 'geometric_correction', 'cloud_mask_removal', 'band_combination', 'classify_land_cover', 'statistical_analysis', 'validation']
Structural Metric: 0.8162889877955118
---------------------------------------
AFlow: ['download_satellite_imagery', 'geometric_correction', 'atmospheric_correction', 'cloud_mask_removal', 'classify_land_cover', 'format_data', 'generate_analysis_reports']
Structural Metric: 0.9616162925958633
---------------------------------------
EarthAgent: ['recommend_satellite_platforms', 'download_satellite_imagery', 'suggest_processing_workflows', 'atmospheric_correction', 'geometric_correction', 'cloud_mask_removal', 'classify_land_cover', 'segment_vegetation', 'detect_urban_expansion', 'monitor_deforestation', 'assess_land_use_efficiency', 'generate_analysis_reports', 'correlation_analysis']
Structural Metric: 0.6346183194267676
\end{lstlisting}
\begin{lstlisting}[
    style=promptstyle, 
    caption={},
    label={aflow_sim_case_4}
]
task_id: d852e006-91ba-4ec1-8716-f8f670e2d313
question: To what extent did the health of the Ballona Wetlands' vegetation depend on the size of its open water areas during the 2023 growing season?
domain: Marine & Water Resources
complexity: Simple
ground_truth: ['download_satellite_imagery', 'geometric_correction', 'atmospheric_correction', 'cloud_mask_removal', 'segment_water_bodies', 'assess_wetland_conditions', 'statistical_analysis', 'generate_analysis_reports']
---------------------------------------
ReAct: ['monitor_crop_health', 'segment_water_bodies', 'correlation_analysis', 'assess_soil_moisture', 'get_weather_data']
Structural Metric: 0.5621206885732097
---------------------------------------
Plan&Execute: ['extract_keywords', 'web_search', 'statistical_analysis', 'correlation_analysis', 'summarize_text']
Structural Metric: 0.5357869020323358
---------------------------------------
CoT: ['web_search', 'web_search', 'web_search', 'download_file', 'correlation_analysis', 'linear_regression', 'summarize_text']
Structural Metric: 0.5402543879495068
---------------------------------------
Debate: ['download_satellite_imagery', 'atmospheric_correction', 'geometric_correction', 'classify_land_cover', 'segment_vegetation', 'segment_water_bodies', 'calculate_vegetation_indices', 'correlation_analysis', 'linear_regression']
Structural Metric: 0.8116500543223487
---------------------------------------
AFlow: ['download_satellite_imagery', 'geometric_correction', 'atmospheric_correction', 'cloud_mask_removal', 'segment_water_bodies', 'monitor_crop_health', 'format_data', 'generate_analysis_reports']
Structural Metric: 0.9380514131238064
---------------------------------------
EarthAgent: ['recommend_satellite_platforms', 'download_satellite_imagery', 'suggest_processing_workflows', 'atmospheric_correction', 'cloud_mask_removal', 'geometric_correction', 'classify_landscape_type', 'assess_urbanization_level', 'segment_vegetation', 'segment_water_bodies', 'monitor_deforestation', 'detect_urban_expansion', 'assess_wetland_conditions', 'track_biodiversity_changes', 'assess_carbon_sequestration', 'evaluate_ecosystem_services', 'monitor_forest_health']
Structural Metric: 0.5651216049354615
\end{lstlisting}
\begin{lstlisting}[
    style=promptstyle, 
    caption={},
    label={aflow_sim_case_5}
]
task_id: bc9bb181-88f2-4f3f-8491-c1b16ae19575
question: For May 2023, generate a health assessment for the vegetation in San Francisco's central community garden, distinguishing between general plant stress and specific infections from common fungal pathogens.
domain: Agriculture & Forestry
complexity: Medium
ground_truth: ['download_satellite_imagery', 'geometric_correction', 'atmospheric_correction', 'cloud_mask_removal', 'crop_image', 'monitor_crop_health', 'detect_plant_diseases']        
---------------------------------------
ReAct: ['monitor_crop_health', 'detect_plant_diseases', 'assess_soil_moisture', 'get_weather_data']
Structural Metric: 0.5226671627364606
---------------------------------------
Plan&Execute: ['monitor_crop_health', 'detect_plant_diseases']
Structural Metric: 0.37084967474084574
---------------------------------------
CoT: ['get_weather_data', 'statistical_analysis', 'download_satellite_imagery', 'atmospheric_correction', 'segment_vegetation', 'monitor_crop_health', 'detect_plant_diseases', 'format_data']
Structural Metric: 0.7985784836391487
---------------------------------------
Debate: ['get_weather_data', 'download_satellite_imagery', 'cloud_mask_removal', 'atmospheric_correction', 'geometric_correction', 'assess_soil_moisture', 'monitor_crop_health', 'detect_plant_diseases']
Structural Metric: 0.8157820831399438
---------------------------------------
AFlow: ['download_satellite_imagery', 'geometric_correction', 'atmospheric_correction', 'cloud_mask_removal', 'monitor_crop_health', 'detect_plant_diseases', 'generate_analysis_reports']
Structural Metric: 0.9277569069748833
---------------------------------------
EarthAgent: ['recommend_satellite_platforms', 'download_satellite_imagery', 'suggest_processing_workflows', 'atmospheric_correction', 'cloud_mask_removal', 'geometric_correction', 'segment_vegetation', 'classify_land_cover', 'monitor_deforestation', 'detect_urban_expansion', 'monitor_crop_health', 'detect_plant_diseases']
Structural Metric: 0.6436540071265643
\end{lstlisting}

\section{Tool Pool Details}
\label{sec:tool_pool}

\subsection{Overview}
We have developed a large set of tools that provide comprehensive coverage across the field of remote sensing and geographic information science. These tools span the full technological spectrum, covering data acquisition, data processing, spatial analysis, and diverse application areas. Our evaluation aims to assess the capabilities of different architectures in problem analysis and task planning. An overview of the tool set is presented in Table~\ref{tab:tools_appendix}.

\onecolumn
\newcolumntype{L}[1]{>{\raggedright\let\newline\\ \arraybackslash\hspace{0pt}}m{#1}}
\newcolumntype{C}[1]{>{\raggedright\let\newline\\
\arraybackslash\hspace{0pt}}m{#1}}
\newcolumntype{R}[1]{>{\raggedright\let\newline\\
\arraybackslash\hspace{0pt}}m{#1}}
\renewcommand{\arraystretch}{2.0}
\begin{longtable}{@{} L{7.0cm} C{3.0cm} R{5.0cm} @{}}
\caption{List of Available Tools}\\
\label{tab:tools_appendix}\\
\toprule
\textbf{Tool Name} & \textbf{Description} & \textbf{Parameters} \\
\midrule
\endfirsthead

\multicolumn{3}{c}%
{{\tablename\ \thetable{} -- continued from previous page}} \\
\toprule
\textbf{Tool Name} & \textbf{Description} & \textbf{Parameters} \\
\midrule
\endhead

\bottomrule
\multicolumn{3}{r}{{Continued on next page}} \\
\endfoot

\bottomrule
\endlastfoot

\texttt{download\_file} & Download file from URL to local storage & \texttt{url: str, save\_path: str, file\_type: str = "auto"} \\
\texttt{web\_search} & Search for relevant information on the web & \texttt{query: str, search\_engine: str = "google", max\_results: int = 10} \\
\texttt{get\_weather\_data} & Get weather data for specified location & \texttt{location: str, date\_range: str = "current"} \\
\texttt{statistical\_analysis} & Perform statistical analysis on data & \texttt{data: list, analysis\_type: str = "basic"} \\
\texttt{linear\_regression} & Perform linear regression analysis & \texttt{x\_data: list, y\_data: list} \\
\texttt{correlation\_analysis} & Calculate correlation between two datasets & \texttt{data1: list, data2: list, method: str = "pearson"} \\
\texttt{convert\_coordinates} & Convert coordinate systems & \texttt{coordinates: tuple, from\_system: str, to\_system: str} \\
\texttt{format\_data} & Format data output & \texttt{data: dict, output\_format: str = "json"} \\
\texttt{read\_database} & Execute database queries & \texttt{connection\_string: str, query: str, query\_type: str = "SELECT"} \\
\texttt{get\_current\_time} & Get current time & \texttt{timezone: str = "UTC", format: str = "\%Y-\%m-\%d \%H:\%M:\%S"} \\
\texttt{calculate\_time\_difference} & Calculate time difference & \texttt{start\_time: str, end\_time: str, unit: str = "days"} \\
\texttt{extract\_keywords} & Extract keywords from text & \texttt{text: str, max\_keywords: int = 10, method: str = "frequency"} \\
\texttt{translate\_text} & Translate text & \texttt{text: str, from\_lang: str, to\_lang: str} \\
\texttt{summarize\_text} & Text summarization & \texttt{text: str, max\_length: int = 200, method: str = "extractive"} \\
\texttt{resize\_image} & Resize image & \texttt{image\_path: str, new\_size: tuple, method: str = "bilinear"} \\
\texttt{crop\_image} & Crop image & \texttt{image\_path: str, crop\_box: tuple, save\_path: str} \\
\texttt{extract\_image\_metadata} & Extract image metadata & \texttt{image\_path: str} \\
\texttt{generate\_analysis\_reports} & Generate analysis report & \texttt{analysis\_results: dict, report\_format: str = "PDF"} \\
\texttt{download\_satellite\_imagery} & Download remote sensing imagery from specified satellite platform & \texttt{satellite: str, region: str, date\_range: str} \\
\texttt{atmospheric\_correction} & Perform atmospheric correction on remote sensing imagery & \texttt{image\_path: str, method: str = "DOS"} \\
\texttt{geometric\_correction} & Perform geometric correction on remote sensing imagery & \texttt{image\_path: str, reference\_image: str} \\
\texttt{cloud\_mask\_removal} & Remove cloud cover from remote sensing imagery & \texttt{image\_path: str, cloud\_threshold: float = 0.3} \\
\texttt{band\_combination} & Perform band combination to generate false/true color imagery & \texttt{image\_path: str, band\_indices: list, combination\_type: str = "RGB"} \\
\texttt{detect\_buildings} & Detect building targets in remote sensing imagery & \texttt{image\_path: str, min\_size: int = 50} \\
\texttt{detect\_roads} & Detect road networks in remote sensing imagery & \texttt{image\_path: str, road\_width\_threshold: float = 3.0} \\
\texttt{detect\_vehicles} & Detect vehicle targets in remote sensing imagery & \texttt{image\_path: str, vehicle\_type: str = "all"} \\
\texttt{detect\_ships} & Detect ship targets in ocean or water bodies & \texttt{image\_path: str, water\_mask: str} \\
\texttt{classify\_land\_cover} & Perform land cover classification on remote sensing imagery & \texttt{image\_path: str, classification\_scheme: str = "CORINE"} \\
\texttt{segment\_vegetation} & Segment vegetation areas in remote sensing imagery & \texttt{image\_path: str, vegetation\_index: str = "NDVI"} \\
\texttt{segment\_water\_bodies} & Segment water body areas in remote sensing imagery & \texttt{image\_path: str, water\_index: str = "NDWI"} \\
\texttt{segment\_urban\_areas} & Segment urban built-up areas in remote sensing imagery & \texttt{image\_path: str, urban\_index: str = "NDBI"} \\
\texttt{segment\_individual\_buildings} & Segment individual building instances & \texttt{image\_path: str, building\_mask: str} \\
\texttt{segment\_agricultural\_fields} & Segment agricultural field instances & \texttt{image\_path: str, field\_boundary\_method: str = "watershed"} \\
\texttt{classify\_landscape\_type} & Classify landscape types of remote sensing imagery & \texttt{image\_path: str, landscape\_categories: list} \\
\texttt{classify\_terrain\_type} & Classify terrain types based on terrain features & \texttt{image\_path: str, dem\_data: str} \\
\texttt{assess\_urbanization\_level} & Assess the urbanization level of the region & \texttt{image\_path: str, urban\_indicators: list} \\
\texttt{enhance\_image\_resolution} & Enhance the spatial resolution of remote sensing imagery & \texttt{image\_path: str, scale\_factor: int = 2} \\
\texttt{detect\_urban\_expansion} & Detect urban expansion changes & \texttt{image\_before: str, image\_after: str} \\
\texttt{monitor\_deforestation} & Monitor forest deforestation changes & \texttt{image\_before: str, image\_after: str, forest\_threshold: float = 0.5} \\
\texttt{assess\_disaster\_damage} & Assess disaster damage & \texttt{image\_before: str, image\_after: str, disaster\_type: str} \\
\texttt{explain\_remote\_sensing\_concepts} & Explain remote sensing concepts and principles & \texttt{concept: str, detail\_level: str = "intermediate"} \\
\texttt{recommend\_satellite\_platforms} & Recommend suitable satellite platforms & \texttt{application: str, requirements: dict} \\
\texttt{suggest\_processing\_workflows} & Suggest remote sensing data processing workflows & \texttt{data\_type: str, analysis\_goal: str} \\
\texttt{provide\_technical\_guidance} & Provide remote sensing technical guidance & \texttt{problem: str, context: str} \\
\texttt{monitor\_crop\_health} & Monitor crop health status & \texttt{image\_path: str, crop\_type: str, health\_indicators: list} \\
\texttt{predict\_crop\_yield} & Predict crop yield & \texttt{image\_path: str, crop\_type: str, growth\_stage: str} \\
\texttt{detect\_plant\_diseases} & Detect plant diseases & \texttt{image\_path: str, crop\_type: str, disease\_library: str} \\
\texttt{assess\_soil\_moisture} & Assess soil moisture & \texttt{image\_path: str, soil\_type: str, depth: str = "surface"} \\
\texttt{optimize\_irrigation\_schedule} & Optimize irrigation schedule & \texttt{field\_data: dict, weather\_forecast: str, crop\_requirements: dict} \\
\texttt{assess\_pasture\_quality} & Assess pasture quality & \texttt{image\_path: str, vegetation\_indices: list, season: str} \\
\texttt{predict\_harvest\_timing} & Predict harvest timing & \texttt{image\_path: str, crop\_type: str, maturity\_indicators: list} \\
\texttt{analyze\_farming\_patterns} & Analyze agricultural farming patterns & \texttt{image\_series: list, region: str, time\_span: str} \\
\texttt{monitor\_flood\_extent} & Monitor flood extent & \texttt{image\_path: str, water\_level\_threshold: float, affected\_areas: list} \\
\texttt{track\_wildfire\_progression} & Track wildfire progression & \texttt{image\_series: list, fire\_detection\_algorithm: str} \\
\texttt{assess\_earthquake\_damage} & Assess earthquake damage & \texttt{image\_before: str, image\_after: str, damage\_categories: list} \\
\texttt{predict\_landslide\_risk} & Predict landslide risk & \texttt{image\_path: str, slope\_data: str, precipitation\_data: str} \\
\texttt{monitor\_drought\_conditions} & Monitor drought conditions & \texttt{image\_series: list, drought\_indices: list, region: str} \\
\texttt{evaluate\_infrastructure\_damage} & Assess infrastructure damage & \texttt{image\_before: str, image\_after: str, infrastructure\_type: str} \\
\texttt{plan\_evacuation\_routes} & Plan evacuation routes & \texttt{road\_network: dict, safe\_zones: list, population\_density: str} \\
\texttt{assess\_recovery\_progress} & Assess recovery progress & \texttt{image\_series: list, recovery\_indicators: list, time\_period: str} \\
\texttt{analyze\_urban\_growth\_patterns} & Analyze urban growth patterns & \texttt{image\_series: list, city\_name: str, growth\_indicators: list} \\
\texttt{assess\_land\_use\_efficiency} & Assess land use efficiency & \texttt{image\_path: str, zoning\_data: str, efficiency\_metrics: list} \\
\texttt{monitor\_traffic\_congestion} & Monitor traffic congestion & \texttt{image\_path: str, road\_network: str, time\_period: str} \\
\texttt{evaluate\_green\_space\_distribution} & Assess green space distribution & \texttt{image\_path: str, green\_indices: list, accessibility\_threshold: float} \\
\texttt{assess\_air\_quality\_patterns} & Assess air quality patterns & \texttt{image\_path: str, pollutant\_types: list, meteorological\_data: str} \\
\texttt{analyze\_population\_density} & Analyze population density & \texttt{image\_path: str, census\_data: str, density\_estimation\_method: str} \\
\texttt{evaluate\_urban\_heat\_island} & Assess urban heat island effect & \texttt{image\_path: str, temperature\_data: str, urban\_morphology: dict} \\
\texttt{assess\_flood\_risk\_zones} & Assess flood risk zones & \texttt{image\_path: str, elevation\_data: str, drainage\_network: str} \\
\texttt{monitor\_air\_pollution} & Monitor air pollution & \texttt{image\_path: str, pollutant\_sensors: list, atmospheric\_conditions: str} \\
\texttt{assess\_water\_quality} & Assess water quality & \texttt{image\_path: str, water\_body\_type: str, quality\_parameters: list} \\
\texttt{track\_biodiversity\_changes} & Track biodiversity changes & \texttt{image\_series: list, species\_indicators: list, habitat\_types: list} \\
\texttt{monitor\_forest\_health} & Monitor forest health & \texttt{image\_path: str, health\_indicators: list, forest\_type: str} \\
\texttt{detect\_illegal\_logging} & Detect illegal logging & \texttt{image\_before: str, image\_after: str, forest\_boundary: str} \\
\texttt{assess\_wetland\_conditions} & Assess wetland conditions & \texttt{image\_path: str, wetland\_type: str, ecological\_indicators: list} \\
\texttt{evaluate\_ecosystem\_services} & Assess ecosystem services & \texttt{image\_path: str, service\_types: list, valuation\_method: str} \\
\texttt{assess\_carbon\_sequestration} & Assess carbon sequestration & \texttt{image\_path: str, vegetation\_type: str, biomass\_estimation\_method: str} \\
\texttt{analyze\_geological\_structures} & Analyze geological structures & \texttt{image\_path: str, geological\_features: list, analysis\_method: str} \\
\texttt{identify\_mineral\_deposits} & Identify mineral deposits & \texttt{image\_path: str, mineral\_types: list, spectral\_signatures: dict} \\
\texttt{assess\_slope\_stability} & Assess slope stability & \texttt{image\_path: str, slope\_angle\_threshold: float, stability\_factors: list} \\
\texttt{map\_fault\_systems} & Map fault systems & \texttt{image\_path: str, fault\_detection\_method: str, tectonic\_context: str} \\
\texttt{evaluate\_groundwater\_resources} & Assess groundwater resources & \texttt{image\_path: str, hydrogeological\_data: str, aquifer\_characteristics: dict} \\
\texttt{assess\_volcanic\_activity} & Assess volcanic activity & \texttt{image\_path: str, thermal\_data: str, gas\_emission\_data: str} \\
\texttt{explore\_mineral\_resources} & Explore mineral resources & \texttt{image\_path: str, target\_minerals: list, exploration\_method: str} \\
\texttt{monitor\_mining\_operations} & Monitor mining operations & \texttt{image\_path: str, mine\_location: str, operation\_type: str} \\
\texttt{assess\_environmental\_impact} & Assess environmental impact & \texttt{image\_before: str, image\_after: str, impact\_indicators: list} \\
\texttt{track\_land\_reclamation} & Track land reclamation & \texttt{image\_series: list, reclamation\_standards: dict, progress\_indicators: list} \\
\texttt{evaluate\_ore\_quality} & Assess ore quality & \texttt{image\_path: str, ore\_type: str, quality\_parameters: list} \\
\texttt{assess\_mining\_safety} & Assess mining safety & \texttt{image\_path: str, safety\_indicators: list, risk\_factors: list} \\
\texttt{monitor\_sea\_surface\_temperature} & Monitor sea surface temperature & \texttt{image\_path: str, temperature\_range: tuple, seasonal\_variation: bool} \\
\texttt{track\_ocean\_currents} & Track ocean currents & \texttt{image\_series: list, current\_vectors: dict, tracking\_method: str} \\
\texttt{assess\_marine\_pollution} & Assess marine pollution & \texttt{image\_path: str, pollution\_types: list, contamination\_levels: dict} \\
\texttt{monitor\_algae\_blooms} & Monitor algae blooms & \texttt{image\_path: str, bloom\_indicators: list, water\_quality\_threshold: float} \\
\texttt{track\_ship\_movements} & Track ship movements & \texttt{image\_path: str, vessel\_types: list, tracking\_duration: str} \\
\texttt{assess\_coastal\_erosion} & Assess coastal erosion & \texttt{image\_before: str, image\_after: str, erosion\_rate\_threshold: float} \\
\texttt{monitor\_ice\_coverage} & Monitor ice coverage & \texttt{image\_path: str, ice\_type: str, coverage\_percentage: float} \\
\texttt{evaluate\_fishing\_grounds} & Assess fishing grounds & \texttt{image\_path: str, fish\_species: list, productivity\_indicators: list} \\
\texttt{detect\_military\_facilities} & Detect military facilities & \texttt{image\_path: str, facility\_types: list, classification\_level: str = "standard"} \\
\texttt{monitor\_border\_security} & Monitor border security & \texttt{image\_path: str, border\_segment: str, detection\_sensitivity: float = 0.8} \\
\texttt{assess\_strategic\_infrastructure} & Assess strategic infrastructure & \texttt{image\_path: str, infrastructure\_categories: list, vulnerability\_analysis: bool = True} \\
\texttt{track\_vessel\_activities} & Track suspicious vessel activities & \texttt{image\_path: str, vessel\_types: list, suspicious\_behavior\_threshold: float = 0.7} \\
\texttt{monitor\_airspace\_violations} & Monitor airspace violations & \texttt{image\_path: str, restricted\_zones: list, aircraft\_detection\_method: str = "radar\_fusion"} \\
\texttt{analyze\_threat\_patterns} & Analyze threat patterns & \texttt{image\_series: list, threat\_indicators: list, risk\_assessment\_model: str} \\
\texttt{assess\_force\_deployment} & Assess force deployment & \texttt{image\_path: str, deployment\_areas: list, asset\_types: list} \\
\texttt{evaluate\_operational\_readiness} & Assess operational readiness & \texttt{image\_path: str, readiness\_indicators: list, assessment\_criteria: dict} \\
\end{longtable}
\twocolumn

\subsection{Shared Tool Set}

In EarthAgent's implementation, a certain set of tools are frequently shared across different specialized sub-agents. These common tools typically provide foundational, general-purpose, or cross-domain functionalities, which are crucial for building a collaborative and efficient multi-agent system. Table \ref{shared-tools} highlights the most frequently shared tools based on the provided setup. The rationale for their shared nature is analyzed below.

\noindent
\textbf{\texttt{format\_data}: The universal need for data formatting.} This tool is responsible for processing and converting data formats. In a complex workflow, the output from one agent (e.g., a JSON file with detection results) might need to be converted into an input format acceptable to the next agent (e.g., a CSV file for statistical analysis). Data format alignment is a fundamental requirement for nearly every agent involved in data processing, analysis, or reporting. It ensures interoperability and allows data to flow smoothly through multi-step tasks. Its widespread sharing is a logical consequence of building a modular system where different agents must communicate and exchange data effectively.

\noindent
\textbf{\texttt{statistical\_analysis}: A core competency for data interpretation.} Remote sensing is inherently data-driven. Statistical analysis is essential for extracting meaningful information from raw data or model outputs. Whether it involves analyzing pixel value distributions (\texttt{PreprocessingAgent}), counting detected objects (\texttt{ObjectDetectorAgent}), or quantifying metrics for specific applications like agriculture or urban planning (\texttt{AgriScoutAgent}, \texttt{UrbanistAIAgent}), this tool is a cornerstone of quantitative analysis. Sharing it across agents ensures that all analytical agents have access to this core capability.

\noindent
\textbf{\texttt{generate\_analysis\_reports}: The standardized endpoint for deliverables.} The ultimate goal for most application-oriented agents is to deliver a clear and comprehensive report to the user. Instead of each agent implementing its own reporting logic, a shared tool ensures consistency in output format and quality. It serves as a standardized ``exit point'' for presenting findings from diverse tasks like disaster assessment or environmental monitoring, reflecting a sound design principle of reusability.

\noindent
\textbf{\texttt{get\_current\_time}: Providing temporal context.} Time is a critical dimension in remote sensing. A timestamp is necessary when acquiring data (\texttt{DataFetcherAgent}), establishing a baseline for time-series analysis (\texttt{AgriScoutAgent}), or logging events for traceability. This simple yet vital utility provides essential temporal context to data and operations across different agents.

\noindent
\textbf{\texttt{get\_weather\_data}: A key ancillary data source.} Weather is a critical external variable that influences both the quality of remote sensing imagery (\eg, cloud cover) and the state of ground features (\eg, soil moisture). For domains highly sensitive to atmospheric conditions—such as data acquisition, agriculture, and oceanography—access to weather data is a prerequisite for accurate analysis. Therefore, it is logically shared among agents operating in these fields.

\noindent
\textbf{\texttt{read\_database}: A universal interface to structured data.} Remote sensing analysis often needs to be augmented with auxiliary information stored in structured databases, such as GIS vector data, historical records, or demographic statistics. For example, an agent might need to query mission parameters (\texttt{DataFetcherAgent}), population data (\texttt{UrbanistAIAgent}), or intelligence databases (\texttt{DefenseSecurityAgent}). Providing a common tool for database access enables various agents to seamlessly integrate with external structured data sources.

\onecolumn
\begin{table}[t]
\centering
\caption{Statistics and Distribution of Frequently Shared Tools}
\label{shared-tools}
\begin{tabular}{p{5cm} >{\centering\arraybackslash}p{4cm} p{5cm}}
\toprule
\textbf{Tool Name} & \textbf{Usage Freq.} & \textbf{Shared by Agents } \\
\midrule
\texttt{format\_data} & 11 & \texttt{PreprocessingAgent}, \texttt{ObjectDetectorAgent}, \texttt{AgriScoutAgent}, \texttt{CrisisCommanderAgent}, \texttt{UrbanistAIAgent}, \etc \\
\addlinespace
\texttt{statistical\_analysis} & 10 & \texttt{PreprocessingAgent}, \texttt{ObjectDetectorAgent}, \texttt{GeneralChatBotAgent}, \texttt{AgriScoutAgent}, \texttt{UrbanistAIAgent}, \etc \\
\addlinespace
\texttt{generate\_analysis\_reports} & 9 & \texttt{GeneralChatBotAgent}, \texttt{AgriScoutAgent}, \texttt{CrisisCommanderAgent}, \texttt{UrbanistAIAgent}, \texttt{EnvironmentalistAgent}, \etc \\
\addlinespace
\texttt{get\_current\_time} & 4 & \texttt{DataFetcherAgent}, \texttt{AgriScoutAgent}, \texttt{OceanographerAgent}, \texttt{DefenseSecurityAgent} \\
\addlinespace
\texttt{get\_weather\_data} & 3 & \texttt{DataFetcherAgent}, \texttt{AgriScoutAgent}, \texttt{OceanographerAgent} \\
\addlinespace
\texttt{read\_database} & 3 & \texttt{DataFetcherAgent}, \texttt{UrbanistAIAgent}, \texttt{DefenseSecurityAgent} \\
\bottomrule
\end{tabular}
\end{table}
\twocolumn

\subsection{Importance Analysis}
\label{Importance Analysis}

Here we clarify how we compute Out-degree Centrality (ODC) and PageRank Centrality (PRC), which are the foundation of Combined Importance (CI) and Combined Cost (CC). These transform standard Levenshtein distance into a weighted one, which makes our \textbf{Structural Metrics} (Section~\ref{Structural Metrics: Evaluating the Overall Plan}) more intelligent and adaptable. We design a process as below to determine the ultimate CI and CC.

\noindent
\textbf{Out-degree Centrality (ODC).}
\label{Out-degree Centrality (ODC)}
This metric quantifies a tool's direct, immediate influence. It operates on the principle that a tool which serves as a direct prerequisite for many other tools is a fundamental building block of the system. Its importance is measured by the breadth of its direct utility.
The calculation first involves counting the number of outgoing connections for each tool in the dependency graph. To ensure comparability across all tools, this raw count is then normalized by dividing it by the highest out-degree found within the entire network. This process yields a score between 0 and 1.

\begin{align}
ODC = \frac{Deg_{out}}{MaxDeg_{out}}
\end{align}

A score of 1 signifies that the tool has the most direct outgoing connections, marking it as a primary foundational component.

\noindent
\textbf{PageRank Centrality (PRC).}
\label{PageRank Centrality (PRC)}
This metric assesses a tool's holistic, or indirect, influence throughout the network. It is based on the idea of transitive importance: a tool is considered significant if other important tools rely on it. This functions as a ``vote of confidence,'' where endorsements from highly influential peers carry more weight than those from peripheral ones.
PRC is determined through an iterative algorithm that simulates the flow of influence through the network. A tool's score is a sum of the scores of all tools that connect to it, weighted by the importance of those linking tools. The process is repeated until the scores stabilize. In more advanced versions, the frequency or strength of a connection can be used as a weight, meaning a more frequently used dependency path contributes more to the target tool's score.

\noindent
\textbf{Out-degree Cost (ODCt).}
This metric translates the direct influence of a tool into an associated cost. The underlying assumption is that a tool with a higher number of direct dependents will incur a greater cost to modify, deprecate, or replace, due to the larger number of immediate downstream components that would be affected.
The cost is modeled as a linear function of the tool's ODC score. It begins with a universal base cost, which represents the intrinsic cost of any tool, and adds to it an amount proportional to its ODC score.

\begin{align}
ODCt = Cost_{base} \times (1 + ODC{score})
\end{align}

\noindent
\textbf{PageRank Cost (PRCt).}
This metric models the cost associated with a tool's systemic importance. The principle is that altering a tool with high global influence (a high PRC score) carries a greater risk and potential cost, as the changes could have far-reaching and unpredictable effects across the entire network ecosystem.
Similar to ODCt, this cost is modeled as a linear function of the tool's PRC score. It also starts with a base cost, to which a value proportional to the PRC score is added. A weighting coefficient can be introduced to adjust how strongly the PRC score impacts the final cost.

\begin{align}
\text{PRCt} = Cost_{base} \times (1 + \alpha \times PRC_{Score})
\end{align}

Here, $\alpha$ is a coefficient that can tune the sensitivity of the cost to the PageRank score.

\noindent
\textbf{Combined Importance (CI).}
To simplify analysis, this metric provides a single, unified score for a tool's overall significance. It synthesizes the two different facets of influence—direct ODC and indirect PRC—into one holistic measure.
The combined importance is typically calculated as the arithmetic mean of the normalized ODC and PRC scores. This gives equal consideration to both the tool's role as a fundamental building block and its role as a central hub within the network.



\noindent
\textbf{Combined Cost (CC).}
This metric provides a corresponding aggregated cost that reflects a tool's overall operational and systemic liabilities. It merges the costs associated with both direct and indirect influence into a single, comprehensive figure.
The combined cost is calculated as the arithmetic mean of ODC and PRC. This approach provides a balanced estimate of the total cost associated with maintaining or altering a tool, taking into account both its immediate and network-wide dependencies.

\begin{align}
CC = \frac{ODCt + PRCt}{2}
\end{align}

Together, CI and CC allow for a direct assessment of a tool's overall value proposition by comparing its comprehensive importance against its comprehensive cost.

\newcolumntype{H}{>{\centering\arraybackslash}p{1.5cm}}

\onecolumn

\begin{longtable}{@{} l *{6}{S[table-format=1.3, round-mode=places, round-precision=3]} @{}}

\caption{Tool Importance Score Evaluation}

\label{tab:tool_scores}\\

\toprule 
\multicolumn{1}{c}{\textbf{Tool Name}} & 
{\parbox{1.0cm}{\centering \textbf{ODC.}}} &
{\parbox{1.0cm}{\centering \textbf{PRC.}}} &
{\parbox{1.0cm}{\centering \textbf{ODCt.}}} &
{\parbox{1.0cm}{\centering \textbf{PRCt.}}} &
{\parbox{1.0cm}{\centering \textbf{CI.}}} &
{\parbox{1.0cm}{\centering \textbf{CC.}}} \\
\midrule 
\endfirsthead 

\caption[]{Tool Importance Score Evaluation} \\
\toprule 
\multicolumn{1}{c}{\textbf{Tool Name}} & 
{\parbox{1.0cm}{\centering \textbf{ODC.}}} &
{\parbox{1.0cm}{\centering \textbf{PRC.}}} &
{\parbox{1.0cm}{\centering \textbf{ODCt.}}} &
{\parbox{1.0cm}{\centering \textbf{PRCt.}}} &
{\parbox{1.0cm}{\centering \textbf{CI.}}} &
{\parbox{1.0cm}{\centering \textbf{CC.}}} \\
\midrule 
\endhead 

\bottomrule
\endfoot 

\bottomrule
\endlastfoot 

\texttt{Assess\_Environmental\_Impact} & 0.17460 & 0.01571 & 1.17460 & 1.01571 & 0.09516 & 1.09516 \\
\texttt{Monitor\_Forest\_Health} & 0.15873 & 0.00202 & 1.15873 & 1.00202 & 0.08037 & 1.08037 \\
\texttt{Optimize\_Irrigation\_Schedule} & 0.03174 & 0.00243 & 1.03174 & 1.00243 & 0.01709 & 1.01709 \\
\texttt{Track\_Biodiversity\_Changes} & 0.11111 & 0.00297 & 1.11111 & 1.00297 & 0.05704 & 1.05704 \\
\texttt{Classify\_Terrain\_Type} & 0.23809 & 0.00277 & 1.23809 & 1.00277 & 0.12043 & 1.12043 \\
\texttt{Summarize\_Text} & 0.20634 & 0.04750 & 1.20634 & 1.04750 & 0.12692 & 1.12692 \\
\texttt{Segment\_Individual\_Buildings}  & 0.28571 & 0.00262 & 1.28571 & 1.00262 & 0.14417 & 1.14417 \\
\texttt{Detect\_Roads} & 0.26984 & 0.00247 & 1.26984 & 1.00247 & 0.13615 & 1.13615 \\
\texttt{Detect\_Vehicles} & 0.19047 & 0.00186 & 1.19047 & 1.00186 & 0.09617 & 1.09617 \\
\texttt{Assess\_Force\_Deployment} & 0.15873 & 0.00325 & 1.15873 & 1.00325 & 0.08099 & 1.08099 \\
\texttt{Classify\_Landscape\_Type} & 0.07936 & 0.00152 & 1.07936 & 1.00152 & 0.04044 & 1.04044 \\
\texttt{Assess\_Water\_Quality} & 0.20634 & 0.00365 & 1.20634 & 1.00365 & 0.10500 & 1.10500 \\
\texttt{Monitor\_Traffic\_Congestion} & 0.12698 & 0.00199 & 1.12698 & 1.00199 & 0.06448 & 1.06448 \\
\texttt{Predict\_Crop\_Yield} & 0.14285 & 0.00454 & 1.14285 & 1.00454 & 0.07370 & 1.07370 \\
\texttt{Track\_Land\_Reclamation} & 0.06349 & 0.00265 & 1.06349 & 1.00265 & 0.03307 & 1.03307 \\
\texttt{Assess\_Operational\_Readiness} & 0.01587 & 0.00153 & 1.01587 & 1.00153 & 0.00870 & 1.00870 \\
\texttt{Evaluate\_Urban\_Heat\_Island} & 0.11111 & 0.00254 & 1.11111 & 1.00254 & 0.05682 & 1.05682 \\
\texttt{Analyze\_Population\_Density} & 0.28571 & 0.00304 & 1.28571 & 1.00304 & 0.14437 & 1.14437 \\
\texttt{Assess\_Pasture\_Quality} & 0.04761 & 0.00160 & 1.04761 & 1.00160 & 0.02461 & 1.02461 \\
\texttt{Evaluate\_Groundwater\_Resources} & 0.12698 & 0.00444 & 1.12698 & 1.00444 & 0.06571 & 1.06571 \\
\texttt{Monitor\_Ice\_Coverage} & 0.14285 & 0.00209 & 1.14285 & 1.00209 & 0.07247 & 1.07247 \\
\texttt{Web\_Search} & 0.36507 & 0.00140 & 1.36507 & 1.00140 & 0.18323 & 1.18323 \\
\texttt{Download\_Satellite\_Imagery} & 0.36507 & 0.01647 & 1.36507 & 1.01647 & 0.19077 & 1.19077 \\
\texttt{Download\_File} & 0.42857 & 0.00202 & 1.42857 & 1.00202 & 0.21529 & 1.21529 \\
\texttt{Get\_Weather\_Data} & 0.63492 & 0.00171 & 1.63492 & 1.00171 & 0.31831 & 1.31831 \\
\texttt{Monitor\_Airspace\_Violations} & 0.06349 & 0.00176 & 1.06349 & 1.00176 & 0.03262 & 1.03262 \\
\texttt{Extract\_Keywords} & 0.12698 & 0.00963 & 1.12698 & 1.00963 & 0.06830 & 1.06830 \\
\texttt{Assess\_Recovery\_Progress} & 0.11111 & 0.00458 & 1.11111 & 1.00458 & 0.05784 & 1.05784 \\
\texttt{Assess\_Flood\_Risk\_Zones} & 0.12698 & 0.00246 & 1.12698 & 1.00246 & 0.06472 & 1.06472 \\
\texttt{Assess\_Earthquake\_Damage} & 0.09523 & 0.00262 & 1.09523 & 1.00262 & 0.04893 & 1.04893 \\
\texttt{Assess\_Slope\_Stability} & 0.11111 & 0.00352 & 1.11111 & 1.00352 & 0.05731 & 1.05731 \\
\texttt{Provide\_Technical\_Guidance} & 0.09523 & 0.02020 & 1.09523 & 1.02020 & 0.05772 & 1.05772 \\
\texttt{Linear\_Regression} & 0.09523 & 0.00699 & 1.09523 & 1.00699 & 0.05111 & 1.05111 \\
\texttt{Detect\_Plant\_Diseases} & 0.09523 & 0.00210 & 1.09523 & 1.00210 & 0.04867 & 1.04867 \\
\texttt{Monitor\_Flood\_Extent} & 0.25396 & 0.00255 & 1.25396 & 1.00255 & 0.12826 & 1.12826 \\
\texttt{Assess\_Disaster\_Damage} & 0.22222 & 0.01050 & 1.22222 & 1.01050 & 0.11636 & 1.11636 \\
\texttt{Segment\_Water\_Bodies} & 0.50793 & 0.00369 & 1.50793 & 1.00369 & 0.25581 & 1.25581 \\
\texttt{Segment\_Vegetation} & 0.50793 & 0.00342 & 1.50793 & 1.00342 & 0.25568 & 1.25568 \\
\texttt{Assess\_Mining\_Safety} & 0.03174 & 0.00216 & 1.03174 & 1.00216 & 0.01695 & 1.01695 \\
\texttt{Track\_Ship\_Movments} & 0.01587 & 0.00144 & 1.01587 & 1.00144 & 0.00865 & 1.00865 \\
\texttt{Segment\_Urban\_Areas} & 0.34920 & 0.00208 & 1.34920 & 1.00208 & 0.17564 & 1.17564 \\
\texttt{Assess\_Soil\_Moisture} & 0.22222 & 0.00301 & 1.22222 & 1.00301 & 0.11262 & 1.11262 \\
\texttt{Assess\_Marine\_Pollution} & 0.11111 & 0.00312 & 1.11111 & 1.00312 & 0.05711 & 1.05711 \\
\texttt{Classify\_Land\_Cover} & 0.92063 & 0.00438 & 1.92063 & 1.00438 & 0.46251 & 1.46251 \\
\texttt{Segment\_Agricultural\_Fields} & 0.19047 & 0.00179 & 1.19047 & 1.00179 & 0.09613 & 1.09613 \\
\texttt{Correlation\_Analysis} & 0.23809 & 0.01667 & 1.23809 & 1.01667 & 0.12738 & 1.12738 \\
\texttt{Suggest\_Processing\_Workflows} & 0.04761 & 0.00708 & 1.04761 & 1.00708 & 0.02735 & 1.02735 \\
\texttt{Analyze\_Urban\_Growth\_Patterns} & 0.19047 & 0.00324 & 1.19047 & 1.00324 & 0.09686 & 1.09686 \\
\texttt{Translate\_Text} & 0.06349 & 0.00301 & 1.06349 & 1.00301 & 0.03325 & 1.03325 \\
\texttt{Predict\_Landslide\_Risk} & 0.00000 & 0.00141 & 1.00000 & 1.00141 & 0.00070 & 1.00070 \\
\texttt{Monitor\_Algae\_Blooms} & 0.12698 & 0.00281 & 1.12698 & 1.00281 & 0.06489 & 1.06489 \\
\texttt{Evaluate\_Infrastructure\_Damage} & 0.17460 & 0.00451 & 1.17460 & 1.00451 & 0.08955 & 1.08955 \\
\texttt{Assess\_Infrastructure\_Damage} & 0.04761 & 0.00144 & 1.04761 & 1.00144 & 0.02453 & 1.02453 \\
\texttt{Monitor\_Drought\_Conditions} & 0.31746 & 0.00365 & 1.31746 & 1.00365 & 0.16055 & 1.16055 \\
\texttt{Band\_Combination} & 0.55555 & 0.00374 & 1.55555 & 1.00374 & 0.27964 & 1.27964 \\
\texttt{Identify\_Mineral\_Deposits} & 0.17460 & 0.00403 & 1.17460 & 1.00403 & 0.08931 & 1.08931 \\
\texttt{Assess\_Wetland\_Conditions} & 0.14285 & 0.00189 & 1.14285 & 1.00189 & 0.07237 & 1.07237 \\
\texttt{Evaluate\_Fishing\_Grounds} & 0.04761 & 0.00274 & 1.04761 & 1.00274 & 0.02518 & 1.02518 \\
\texttt{Assess\_Strategic\_Infrastructure} & 0.22222 & 0.00407 & 1.22222 & 1.00407 & 0.11314 & 1.11314 \\
\texttt{Evaluate\_Ore\_Quality} & 0.06349 & 0.00206 & 1.06349 & 1.00206 & 0.03277 & 1.03277 \\
\texttt{Evaluate\_Green\_Space\_Distribution} & 0.12698 & 0.00182 & 1.12698 & 1.00182 & 0.06440 & 1.06440 \\
\texttt{Resize\_Image} & 0.50793 & 0.00171 & 1.50793 & 1.00171 & 0.25482 & 1.25482 \\
\texttt{Assess\_Urbanization\_Level} & 0.23809 & 0.00259 & 1.23809 & 1.00259 & 0.12034 & 1.12034 \\
\texttt{Calculate\_Time\_Difference} & 0.53968 & 0.00580 & 1.53968 & 1.00580 & 0.27274 & 1.27274 \\
\texttt{Track\_Vessel\_Activities} & 0.17460 & 0.00408 & 1.17460 & 1.00408 & 0.08934 & 1.08934 \\
\texttt{Analyze\_Farming\_Patterns} & 0.09523 & 0.00190 & 1.09523 & 1.00190 & 0.04856 & 1.04856 \\
\texttt{Track\_Ocean\_Currents} & 0.17460 & 0.00179 & 1.17460 & 1.00179 & 0.08819 & 1.08819 \\
\texttt{Track\_Wildfire\_Progression} & 0.12698 & 0.00182 & 1.12698 & 1.00182 & 0.06440 & 1.06440 \\
\texttt{Evaluate\_Operational\_Readiness} & 0.09523 & 0.00407 & 1.09523 & 1.00407 & 0.04965 & 1.04965 \\
\texttt{Assess\_Air\_Quality\_Patterns} & 0.11111 & 0.00321 & 1.11111 & 1.00321 & 0.05716 & 1.05716 \\
\texttt{Monitor\_Sea\_Surface\_Temperature} & 0.26984 & 0.00216 & 1.26984 & 1.00216 & 0.13600 & 1.13600 \\
\texttt{Evaluate\_Ecosystem\_Services} & 0.11111 & 0.00562 & 1.11111 & 1.00562 & 0.05837 & 1.05837 \\
\texttt{Assess\_Carbon\_Sequestration} & 0.11111 & 0.00366 & 1.11111 & 1.00366 & 0.05738 & 1.05738 \\
\texttt{Monitor\_Crop\_Health} & 0.22222 & 0.00265 & 1.22222 & 1.00265 & 0.11243 & 1.11243 \\
\texttt{Explore\_Mineral\_Resources} & 0.11111 & 0.00360 & 1.11111 & 1.00360 & 0.05735 & 1.05735 \\
\texttt{Extract\_Image\_Metadata} & 0.12698 & 0.00714 & 1.12698 & 1.00714 & 0.06706 & 1.06706 \\
\texttt{Cloud\_Mask\_Removal} & 0.66666 & 0.00864 & 1.66666 & 1.00864 & 0.33765 & 1.33765 \\
\texttt{Recommend\_Satellite\_Platforms} & 0.01587 & 0.01531 & 1.01587 & 1.01531 & 0.01559 & 1.01559 \\
\texttt{Atmospheric\_Correction} & 0.50793 & 0.00965 & 1.50793 & 1.00965 & 0.25879 & 1.25879 \\
\texttt{Detect\_Military\_Facilities} & 0.15873 & 0.00219 & 1.15873 & 1.00219 & 0.08046 & 1.08046 \\
\texttt{Monitor\_Deforestation} & 0.28571 & 0.00266 & 1.28571 & 1.00266 & 0.14418 & 1.14418 \\
\texttt{Plan\_Evacuation\_Routes} & 0.07936 & 0.00575 & 1.07936 & 1.00575 & 0.04255 & 1.04255 \\
\texttt{Monitor\_Border\_Security} & 0.12698 & 0.00364 & 1.12698 & 1.00364 & 0.06531 & 1.06531 \\
\texttt{Geometric\_Correction} & 0.61904 & 0.01110 & 1.61904 & 1.01110 & 0.31507 & 1.31507 \\
\texttt{Track\_Ship\_Movements} & 0.20634 & 0.00325 & 1.20634 & 1.00325 & 0.10480 & 1.10480 \\
\texttt{Crop\_Image} & 1.00000 & 0.00755 & 2.00000 & 1.00755 & 0.50377 & 1.50377 \\
\texttt{Detect\_Urban\_Expansion} & 0.14285 & 0.00270 & 1.14285 & 1.00270 & 0.07278 & 1.07278 \\
\texttt{Map\_Fault\_Systems} & 0.19047 & 0.00262 & 1.19047 & 1.00262 & 0.09654 & 1.09654 \\
\texttt{Convert\_Coordinates} & 0.77777 & 0.01255 & 1.77777 & 1.01255 & 0.39516 & 1.39516 \\
\texttt{Analyze\_Geological\_Structures} & 0.19047 & 0.00296 & 1.19047 & 1.00296 & 0.09671 & 1.09671 \\
\texttt{Detect\_Illegal\_Logging} & 0.04761 & 0.00181 & 1.04761 & 1.00181 & 0.02471 & 1.02471 \\
\texttt{Predict\_Harvest\_Timing} & 0.04761 & 0.00235 & 1.04761 & 1.00235 & 0.02498 & 1.02498 \\
\texttt{Read\_Database} & 0.52380 & 0.00145 & 1.52380 & 1.00145 & 0.26263 & 1.26263 \\
\texttt{Assess\_Coastal\_Erosion} & 0.12698 & 0.00255 & 1.12698 & 1.00255 & 0.06477 & 1.06477 \\
\texttt{Predict\_Landslide\_Risk} & 0.15873 & 0.00519 & 1.15873 & 1.00519 & 0.08196 & 1.08196 \\
\texttt{Analyze\_Threat\_Patterns} & 0.12698 & 0.01422 & 1.12698 & 1.01422 & 0.07060 & 1.07060 \\
\texttt{Generate\_Analysis\_Reports} & 0.11111 & 0.27282 & 1.11111 & 1.27282 & 0.19196 & 1.19196 \\
\texttt{Detect\_Ships} & 0.17460 & 0.00222 & 1.17460 & 1.00222 & 0.08841 & 1.08841 \\
\texttt{Format\_Data} & 0.04761 & 0.21034 & 1.04761 & 1.21034 & 0.12898 & 1.12898 \\
\texttt{Explain\_Remote\_Sensing\_Concepts} & 0.06349 & 0.00140 & 1.06349 & 1.00140 & 0.03244 & 1.03244 \\
\texttt{Statistical\_Analysis} & 0.36507 & 0.03619 & 1.36507 & 1.03619 & 0.20063 & 1.20063 \\
\texttt{Monitor\_Air\_Pollution} & 0.07936 & 0.00233 & 1.07936 & 1.00233 & 0.04084 & 1.04084 \\
\texttt{Monitor\_Mining\_Operations} & 0.14285 & 0.00227 & 1.14285 & 1.00227 & 0.07256 & 1.07256 \\
\texttt{Download\_Satellite\_Imagery} & 0.01587 & 0.00140 & 1.01587 & 1.00140 & 0.00863 & 1.00863 \\
\texttt{Enhance\_Image\_Resolution} & 0.47619 & 0.00265 & 1.47619 & 1.00265 & 0.23942 & 1.23942 \\
\texttt{Get\_Current\_Time} & 0.11111 & 0.00141 & 1.11111 & 1.00141 & 0.05626 & 1.05626 \\
\texttt{Detect\_Buildings} & 0.25396 & 0.00230 & 1.25396 & 1.00230 & 0.12813 & 1.12813 \\
\texttt{Assess\_Land\_Use\_Efficiency} & 0.15873 & 0.00329 & 1.15873 & 1.00329 & 0.08101 & 1.08101 \\

\end{longtable}
\twocolumn

\section{A Basic Release Version of EarthAgent}
\label{sec:demo}

\begin{figure*}
  \centering
    \includegraphics[width=\textwidth]{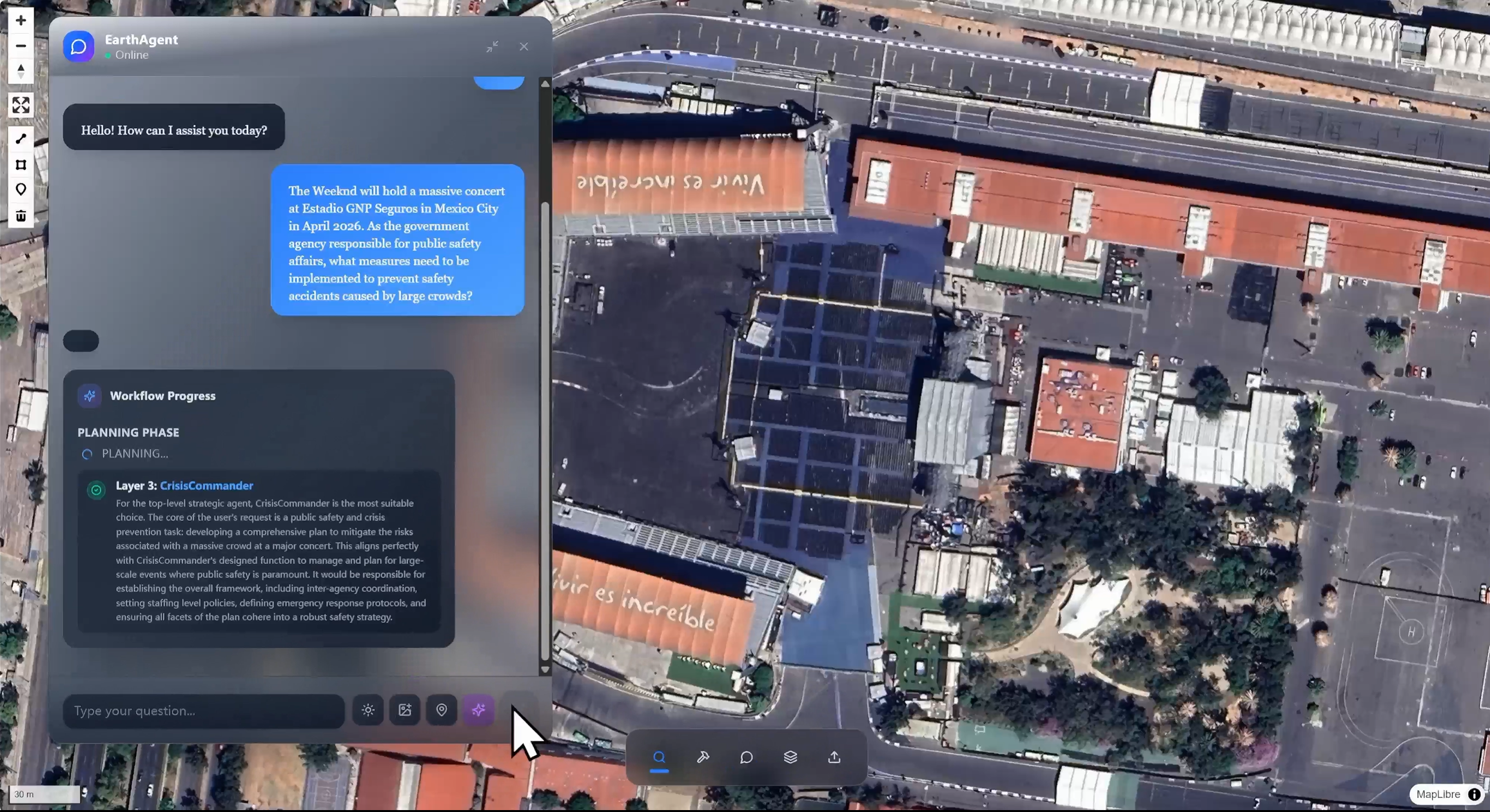}
    \caption{\textbf{A snapshot of EarthAgent web-application.} This image is taken from a comprehensive task on risk prediction for a large-scale concert. EarthAgent first analyzes the whole question and select sub-agents in each layer.}
    \label{demo-1}
\end{figure*}
\begin{figure*}
  \centering
    \includegraphics[width=\textwidth]{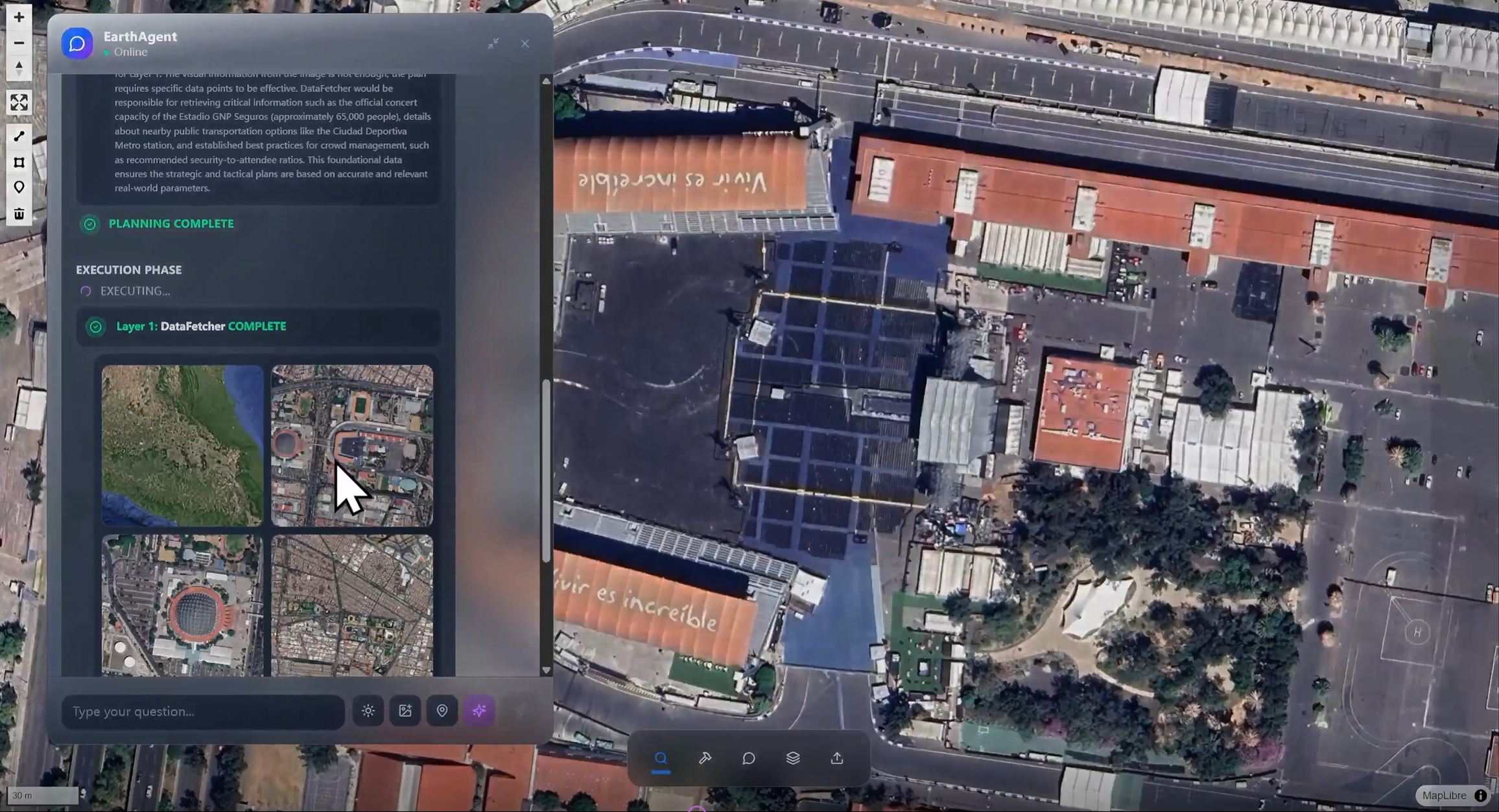}
    \caption{\textbf{A snapshot of EarthAgent web-application.} This image is taken from the same task scenario as Figure~\ref{demo-1}. EarthAgent calls agent from the first layer (Data Acquisition and Preprocessing Layer), who has acquired relevant ROIs successfully.}
    \label{demo-2}
\end{figure*}
\begin{figure*}
  \centering
    \includegraphics[width=\textwidth]{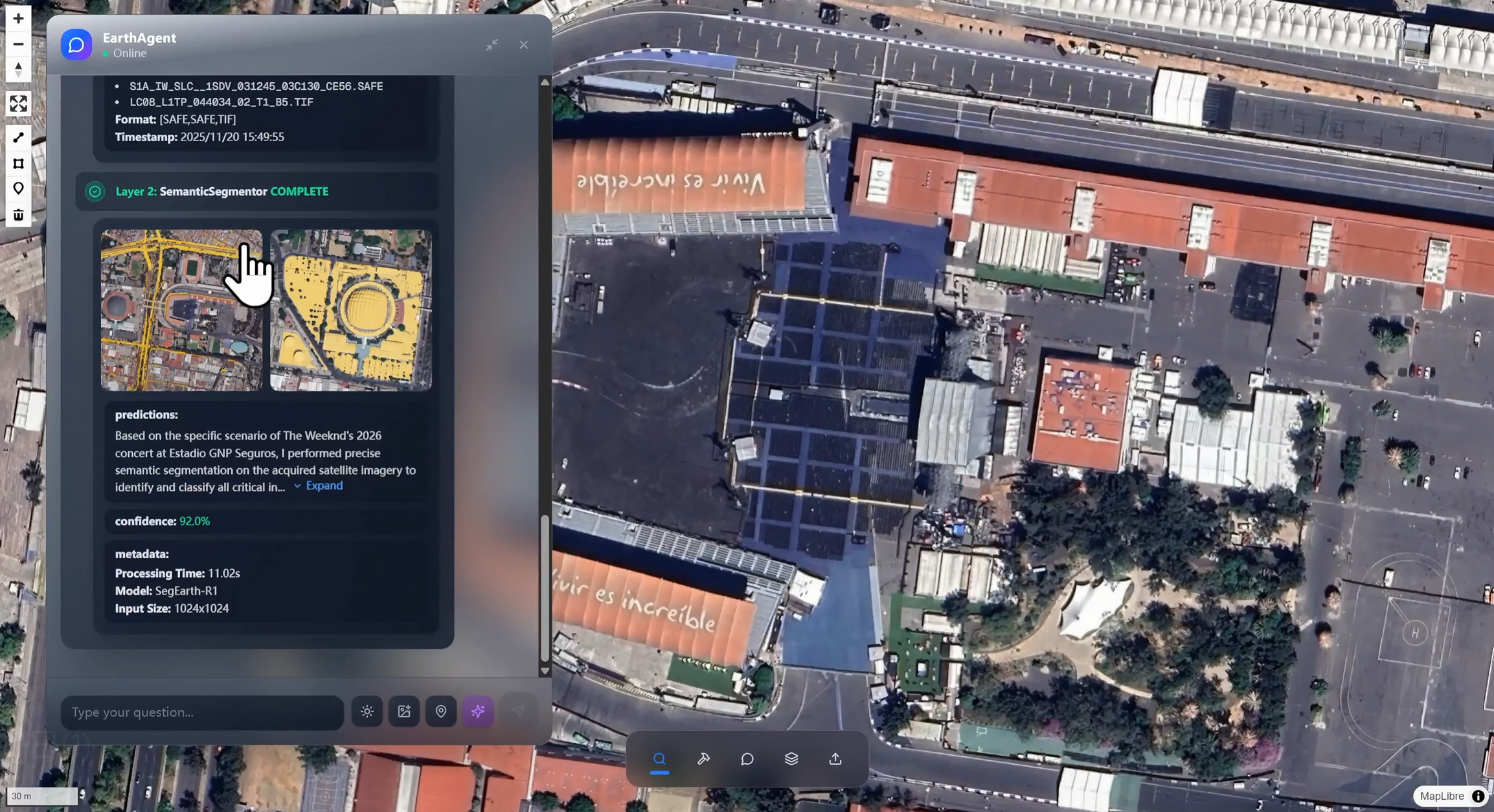}
    \caption{\textbf{A snapshot of EarthAgent web-application.} This image is also taken from the same task scenario as Figure~\ref{demo-1}. EarthAgent has currently progressed to layer 2 (Data Processing and Analysis Layer), where semantic segmentation of the regions of interest has been achieved.}
    \label{demo-3}
\end{figure*}
\begin{figure*}
  \centering
    \includegraphics[width=\textwidth]{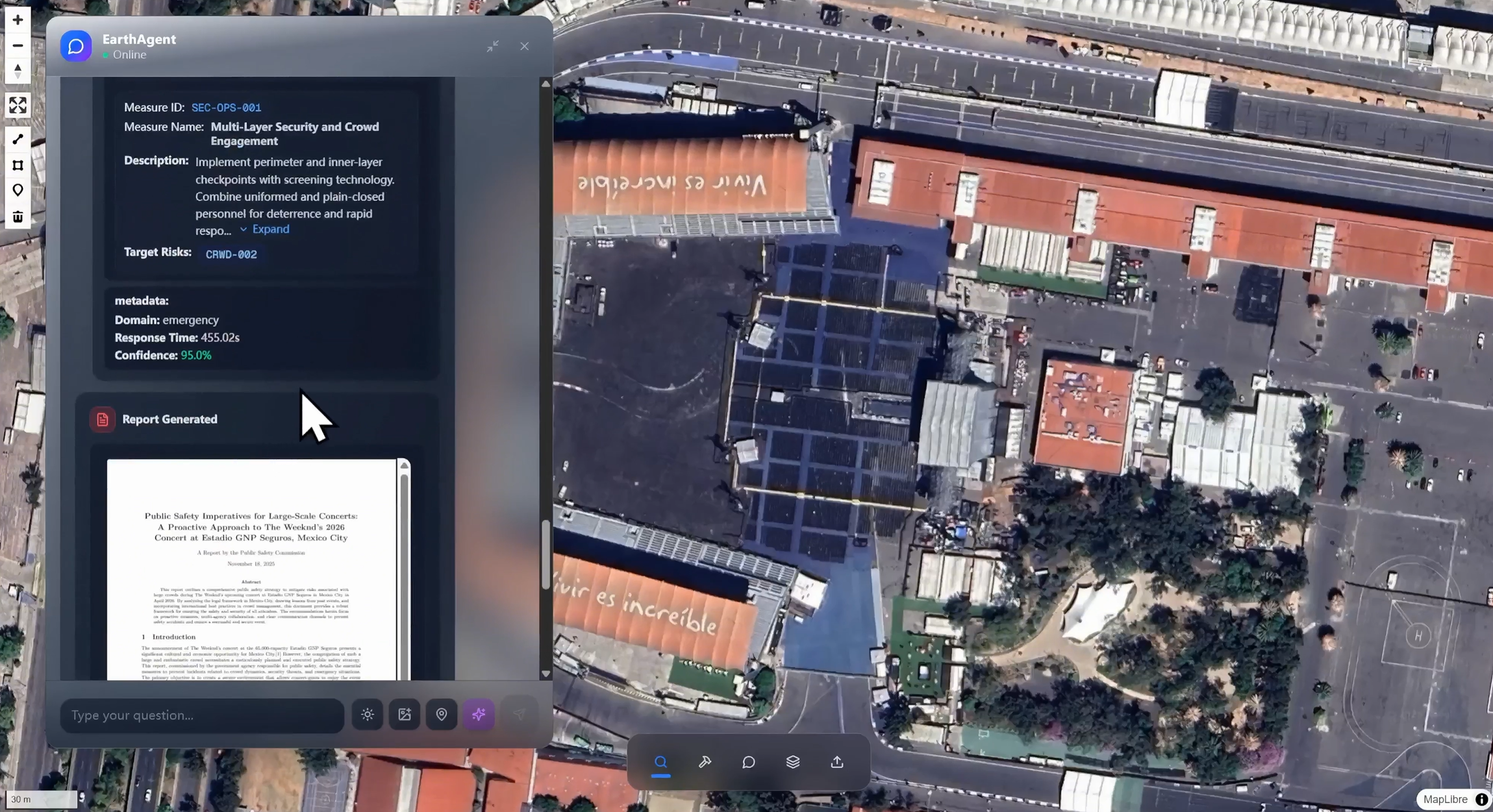}
    \caption{\textbf{A Snapshot of EarthAgent Web-Application.} Taken from the same task scenario as Figure~\ref{demo-1}, this image showcases agent from layer 3 (Synthesis and Application Layer) executes rigorous reasoning and synthesize to output a integrated technical report.}
    \label{demo-4}
\end{figure*}
EarthAgent is built upon an HTAM-style multi-agent architecture and has demonstrated significant robustness and efficacy. To facilitate practical deployment, we have developed EarthAgent into a web-based application capable of addressing complex real-world challenges. A video demonstration of the inference process is available on our project page and code repository. A snapshot of the system interface is illustrated in Figure~\ref{demo-1}, Figure~\ref{demo-2}, Figure~\ref{demo-3}, and Figure~\ref{demo-4}.

\section{A Note on Concurrent Work with a Similar Title}
\label{A Note on Concurrent Work with a Similar Title}

During the final preparation of this paper, we became aware of a concurrently released work, ``Earth-Agent: Unlocking the Full Landscape of Earth Observation with Agents''~\cite{feng2025earth}. We acknowledge the similarity in our titles, which may potentially lead to confusion. We assure that the similarity in the titles of the two papers is purely coincidental. This section aims to clarify the distinct focuses and contributions of the two works.

The ``Earth-Agent''~\cite{feng2025earth} introduces an agentic framework using a ReAct-style reasoning loop and leveraging Model Context Protocol (MCP). It is designed specifically for Earth Observation (EO), addressing key limitations of existing MLLMs and early EO agents. Earth-Agent's core focus is supporting RGB and multi-spectral data, enabling it to perform complex scientific tasks such as geophysical parameter retrieval, trend analysis, and change detection.

In contrast, our work, ``Designing Domain-Specific Agents via Hierarchical Task Abstraction Mechanism'', investigates a new framework for designing domain-specific agents, \ie, HTAM. Briefly speaking, it moves beyond social role assignments and instead organizes agents into a logical hierarchy derived from the intrinsic dependencies of a domain's tasks. As a new agent design paradigm, HTAM stands alongside other famous agent architectures like ReAct, Plan\&Execute, AFlow, \etc. Based on that, we present its implementation called EarthAgent. Experiments show that it substantially outperforms a range of established single- and multi-agent systems.
To further elucidate the differences, we provide a concise comparison below:

\noindent
\textbf{Differences of Agentic System.}
EarthAgent is an implementation for HTAM for remote sensing tasks. In other words, EarthAgent is an HTAM-style multi-agent system, while the concurrent ``Earth-Agent'' is a ReAct-style single-agent framework.

\noindent
\textbf{Differences of Comparison Items.}
We conducted numerous experiments and compared our EarthAgent with instantiations of other agent design paradigms like ReAct, Plan\&Execute and Debate \etc. While ``Earth-Agent'' was evaluated with general agents such as Operator by OpenAI and Manus~\cite{shen2025mind} on Earth-specific tasks and other remote sensing MLLMs which are basically ReAct-style agent or agentic workflow.

Finally, we would like to explicitly acknowledge that the concurrent work ``Earth-Agent: Unlocking the Full Landscape of Earth Observation with Agents'' represents a valuable contribution to the field, presenting a robust and thoughtful methodology. We wish to state unequivocally that the striking similarity in the nomenclature between their ``Earth-Agent'' and our ``EarthAgent'' is purely a coincidence in naming conventions rather than any substantive overlap. Fundamentally, the two works diverge in their core motivation and methodology. By providing this part, we sincerely hope to preempt any potential for misinterpretation within the community and ensure that both bodies of work are evaluated fairly.

\end{document}